\documentclass{article} 
\usepackage{iclr2027_conference,times}

\usepackage{amsmath,amsfonts,bm}

\def\eqref#1{equation~\ref{#1}}

\def\1{\bm{1}}

\DeclareMathAlphabet{\mathsfit}{\encodingdefault}{\sfdefault}{m}{sl}
\SetMathAlphabet{\mathsfit}{bold}{\encodingdefault}{\sfdefault}{bx}{n}

\usepackage{amsthm}
\usepackage{algpseudocode}
\usepackage{algorithm}
\usepackage{hyperref}
\usepackage{url}
\usepackage{bm}
\usepackage{amssymb}

\usepackage[dvipsnames]{xcolor}

\newtheorem{theorem}{Theorem}
\newtheorem{lemma}{Lemma}
\newtheorem{assumption}{Assumption}
\newtheorem{remark}{Remark}
\newtheorem{definition}{Definition}

\newcommand{\norm}[1]{\left\|#1\right\|}

\newcommand{\abs}[1]{\left|#1\right|}

\newcommand{\indep}{\!\perp \!\!\! \perp}

\title{Generative Sequence Modeling for Infinite Memory Processes via Predictive States}

\author{Michael Wieck-Sosa \& Cosma Rohilla Shalizi \\
Department of Statistics \& Data Science\\
Carnegie Mellon University\\
Pittsburgh, PA 15213, USA \\
\texttt{\{mwiecksosa, cshalizi\}@cmu.edu}
}

\iclrfinalcopy 
\usepackage{tikz}
\usetikzlibrary{positioning,arrows.meta}
\begin{document}

\maketitle

\begin{abstract}
We consider estimating the one-step-ahead conditional distribution of a multivariate stochastic process. Many existing approaches rely on structural assumptions such as finite-range memory, sparsity, or additivity, which can be poorly suited to processes with long-range nonlinear interactions. However, without such structural assumptions, nonparametric estimation is known to be challenging due to the curse of dimensionality. To address this challenge, we introduce a new estimator based on the idea of learning representations of the \emph{predictive states} of a stochastic process. We theoretically analyze several aspects of this problem. As a bonus, we establish convergence rate guarantees for an instantiation of our method based on deep neural network estimators. Our results suggest that faster convergence rates are achievable when the predictive state space is lower-dimensional. That is, when the past history can be compressed into a lower-dimensional statistic that is sufficient for predicting the future. We implement our method in PyTorch and evaluate its performance on synthetic data and real \emph{C. elegans} data.
\end{abstract}


\section{Introduction}\label{section:intro}
 
Let $X_t$, $t\in\mathbb{Z}$, be a stationary process. We aim to learn the one-step-ahead conditional distribution \begin{equation}\label{eqn:one_step_ahead_cond_law}\mathcal{L}(X_{t+1} \mid \mathcal{F}_t), \quad t\in\mathbb{Z},\end{equation} where $\mathcal{F}_t = \sigma(X_t,X_{t-1},\ldots)$, $t\in\mathbb{Z}$, is an increasing sequence of $\sigma$-algebras known as the \textit{filtration} which can be understood as the information available up to time $t$. This is a fundamental problem in multi-step forecasting, optimization, and control, because, as we explain below, success at this task lets us learn the predictive distribution for the {\em whole future}, \begin{equation}\label{eqn:infinite_step_ahead_cond_law}\mathcal{L}(X_{t+1:\infty} \mid \mathcal{F}_t), \quad t\in\mathbb{Z}.\end{equation}

We consider the class of stochastic processes that admit finite-dimensional \emph{predictive states}, which formalize the notion that the information up to time $t$ (i.e., the filtration $\mathcal{F}_t$) can be compressed into a finite-dimensional statistic, $S_t$, that is \textit{sufficient for predicting the future} values of a stochastic process. In this sense, finite-dimensional predictive states can be viewed as a dynamic version of sufficient dimension reduction \citep{Li_1991,cook_2007} for stochastic processes.

Our main goal is to study a recursive estimator of the one-step-ahead conditional distribution that can achieve faster convergence rates when the predictive state space is lower-dimensional. This structural property of a ``low-dimensional predictive state space'' offers an alternative to other conditions used to mitigate the curse of dimensionality, such as additivity \citep{Stone_1985}, sparsity \citep{Tibshirani_1996}, low intrinsic dimensionality \citep{Bickel_2007}, and hierarchical compositionality \citep{schmidt_hieber_2020}. As far as we know, we are the first to theoretically analyze how conditional distribution estimators can take advantage of low-dimensional predictive states.

\subsection{Related Work}\label{subsection:related_work}

We provide an overview of relevant strands of literature.

\subsubsection{Predictive States}\label{subsubsection:predictive_states_literature}

Predictive states have a long history, with many independent reinventions of (essentially) the same ideas in different formalisms and different jargons \citep{Salmon-1971,Salmon-1984, knight_prediction_process,Knight-essays-on-prediction, Knight-foundations-of-prediction, aldous_prediction_process,Crutchfield_Young_1989,cosma_computational_mechanics,Grassberger_1986,jaeger_2000,littman_et_al_2001,langford_et_al_2009}.

\paragraph*{Computational Mechanics.} In computational mechanics \citep{Crutchfield_Young_1989,cosma_computational_mechanics}, methods for reconstructing (a version of) predictive states have been used in various fields, such as ecology \citep{BOSCHETTI200837}, chemistry \citep{nerukh_et_al_2002}, neuroscience \citep{spike-train-complexity,nehaniv_antonova_2017,Munoz_et_al_2020}, finance \citep{PARK2007179}, and the Earth Sciences \citep{Palmer_et_al_2000,Clark_et_al_2003}. Our focus is related to, but distinct from, computational mechanics \citep{Crutchfield_Young_1989,cosma_computational_mechanics}. Originally, \citet{Crutchfield_Young_1989} focused on equivalence classes of histories that induce the same conditional distribution over future events, referred to as the ``causal'' states. We instead focus on the process, $S_t$, $t\in\mathbb{Z}$, which we refer to as the predictive states, to avoid overloading terms and conflating predictive states with causality in the sense of \cite{pearl_causality, SGS_CPS} (though see \citet{What-is-a-macrostate}).

\paragraph*{Deep Learning.} Function learning approaches for predictive state estimation have also been proposed \citep{langford_et_al_2009}. Building on these ideas, \cite{boots_et_al_2013} and \cite{hefy_et_al_2015} proposed learning predictive states by regressing ``sufficient features'' of future observations on the past observations. For example, the sufficient features for a Gaussian conditional distribution are based on the first two conditional moments. More recently, \cite{downey_et_al_2017} and \cite{Venkatraman_et_al_2017} considered using the hidden states of recurrent neural networks (RNNs) \citep{hopfield,elman,vanishing_grad,JORDAN1997471,ICML2011Sutskever_524,graves2014generatingsequencesrecurrentneural,bahdanau2016neuralmachinetranslationjointly} to encode the predictive states.

We adapt predictive states for the purposes of \textit{generative sequence modeling}, which plays a central role in modern machine learning largely due to the success of the transformer architecture \citep{Vaswani_et_al_2017}. Ideally, we would like to characterize each $\mathcal{L}(X_{t+1} \mid \mathcal{F}_t)$ by identifying ``sufficient features'', as in the aforementioned works. Unfortunately, outside of simple cases, it is challenging to determine what these sufficient features should be. As a result, this approach does not readily extend to richer families of distributions with nonstandard behavior.

\paragraph*{Random Fourier Features.} We show that user-chosen ``sufficient features'' can be replaced by a small number of \textit{random} features \citep{Rahimi_Recht_2007_random_features}. We emphasize that our use of random features is fundamentally different from their typical use for approximating kernel methods. In contrast, we use a small number of random features as low-dimensional witness functions (or probes) to distinguish different predictive states, as we will discuss.

\subsubsection{Comparisons}\label{subsubsection:comparison_with_existing}

\paragraph*{Predictive State Recurrent Neural Networks (PSRNNs).} The closest work to ours is \cite{downey_et_al_2017}. They introduce PSRNNs, which combine recurrent neural networks (RNNs) and predictive state representations (PSRs) based on random Fourier features. Their main contributions are the PSRNN architecture and an efficient training procedure based on Two-Stage Regression (2SR) \citep{hefy_et_al_2015} and Backpropagation Through Time (BPTT) \citep{bptt}.

\paragraph*{Modern State Space Models (SSMs) and Recurrent Models.} 

Recently, there has been renewed interest in SSMs due to the quadratic complexity of self-attention in the
input sequence length \citep{Vaswani_et_al_2017}. There have been many advancements in \textit{structured} SSMs, which are closely related to classical SSMs and RNNs. Notable  developments related to structured SSMs include high-order polynomial projection operators (HiPPO)
\citep{gu2022trainhippostatespace,gu2020hipporecurrentmemoryoptimal}, H3 \citep{fu2023hungryhungryhipposlanguage}, the Structured State Space sequence (S4) model \citep{gu2022efficientlymodelinglongsequences}, Mamba 1-3 \citep{gu2024mambalineartimesequencemodeling,dao2024transformersssmsgeneralizedmodels,ICLR2026_8abd2043}. In closely related lines of work on recurrent modeling, we also mention linear attention \citep{pmlr-v119-katharopoulos20a}, DeltaNet \citep{NEURIPS2024_d13a3eae}, and Gated DeltaNet \citep{gated_deltanet}. These works introduce different types of recurrence to efficiently model sequences using latent states.

\paragraph*{Identifiability of Latent States.} SSMs model sequences as being driven by latent dynamics. However, these latent states generally cannot be identified from observational data alone. In contrast, we show that \textit{predictive} states can be identified from estimable representations with well-defined probabilistic interpretations in terms of the conditional distribution of future observations given the past observations. Crucially, our approach does not require a user-imposed parametric model. This identifiability is important if one desires a fixed coordinate system for interpreting the hidden states, e.g., for comparing the predictive states across models or the same model across training runs.

\subsubsection{Purpose of Our Work}\label{subsubsection:purpose_of_work}


\paragraph*{Foundations of Recurrent Modeling.}

The purpose of our paper is to provide a \textit{theoretical understanding} of generative sequence modeling based on predictive state representations, rather than to introduce a new architecture or training procedure. Historically, progress in machine learning has mostly been driven by architectural advancements, as judged by performance improvements on benchmarks. Nevertheless, research that formalizes ideas and asks fundamental questions is valuable because it can help clarify why certain ideas work (e.g., predictive state representations, in the case of this paper) and can inform the development of new architectures (e.g., for sequence modeling).

Specifically, we revisit ideas from the literature on predictive state representations and formalize the underlying probabilistic sequence model (Section~\ref{subsection:model}) and study identification of predictive states (Section~\ref{subsection:predictive_state_representations}). As a bonus, we use these results to derive convergence rates (Appendix~\ref{section:theoretical_results_on_convergence_rates}). In view of the societal impacts of generative artificial intelligence and the growing demand for computationally efficient alternatives to transformers \citep{Vaswani_et_al_2017} based on recurrent architectures \citep{gu2024mambalineartimesequencemodeling,dao2024transformersssmsgeneralizedmodels,ICLR2026_8abd2043,pmlr-v119-katharopoulos20a,peng-etal-2023-rwkv,gu2022efficientlymodelinglongsequences,fu2023hungryhungryhipposlanguage,gu2022trainhippostatespace,gu2020hipporecurrentmemoryoptimal,NEURIPS2025_c4025018,pmlr-v139-schlag21a,NEURIPS2024_d13a3eae,gated_deltanet,nvidia2025nemotronhfamilyaccurateefficient,kimiteam2025kimilinearexpressiveefficient,tencenthunyuanteam2025hunyuanturbosadvancinglargelanguage,ICLR2026_57d8ebf4}, it is becoming increasingly important to understand the statistical principles underlying recurrent sequence modeling. This paper aims to advance our understanding of this important topic.

\paragraph*{Purpose of Experiments.} The goal of our experiments is to help us understand the problem of learning representations of predictive states in a controlled setting, rather than to establish state-of-the-art performance on benchmarks. In general, our ideas do not require using an RNN (or any particular recurrent architecture, for that matter), though we study an instantiation of our method based on RNNs because many readers are already familiar with them. Hence, while we \textit{do} compare with a reasonable baseline model, the objectives of this paper (stated above) do not necessitate an exhaustive comparison with existing sequence modeling methods.

\subsection{Our Contributions}

The main contributions of this paper are:
\begin{enumerate}
    \item In Theorem~\ref{thm:one_step_implies_pred_suff_entire_future} (and Theorem~\ref{thm:one_step_implies_pred_suff_entire_future_standard_Borel_space} in the appendix), we show that any process $S_t$, $t\in\mathbb{Z}$, which is one-step-ahead \textit{predictively sufficient}, $X_{t+1} \indep X_{-\infty:t} \mid S_t$ for all $t \in \mathbb{Z}$, and which can be updated recursively, $S_{t+1} = \tau(S_t, X_{t+1})$ for some (measurable) function $\tau$, is predictively sufficient for the \textit{whole future}, $X_{t+1:\infty} \indep X_{-\infty:t} \mid S_t$ for all $t \in \mathbb{Z}$. This 
generalizes Corollary 2 of Theorem 1 in \cite{cosma_computational_mechanics} from discrete measurable spaces to standard Borel spaces.
    \item  In Theorem~\ref{thm:Phi_is_one_to_one}, we show that $p$-dimensional predictive states $S_t$ can be identified from the conditional expectations of $2p+1$ random Fourier features \citep{Rahimi_Recht_2007_random_features} of a finite block of future observations. In Theorem~\ref{thm:equality_of_cond_distr_St_Zt}, we show that these conditional expectations also identify the one-step-ahead conditional distribution of the process.
\end{enumerate}

\paragraph*{Paper Outline.} The main text is structured as follows. Section~\ref{section:methods} presents the main ideas. Specifically, Section~\ref{subsection:model} introduces the probabilistic sequence model, Section~\ref{subsection:predictive_state_representations} introduces the predictive state representations, Section~\ref{subsection:learning_conditional_distributions} presents the algorithms, and Section~\ref{subsection:implmenetation_neural_networks} implements our two-stage estimator using deep neural networks. Section~\ref{subsubsection:rnn} contains the implementation for the first stage, in which we learn the predictive state representation using a recurrent neural network (RNN). Section~\ref{subsubsection:mdn} contains the details for the second stage, in which we learn the conditional distribution using a mixture density network (MDN). Section~\ref{section:experiments} reports the results from the experiments. Section~\ref{section:conclusion_and_future_work} concludes and discusses future work.

The appendix contains additional results. Appendix~\ref{section:theoretical_results_on_convergence_rates} contains bonus results on the convergence rate of our method based on an RNN and MDN. Appendix~\ref{section:additional_theoretical_results} contains additional results about the MDN estimator and the ``predictive sufficiency'' of predictive states. Appendix~\ref{section:preliminaries} contains results from the literature used in our proofs. The proofs are given in Appendix~\ref{section:main_results} and Appendix~\ref{section:proof_of_additional_results}.

\section{Methods and Main Ideas}
\label{section:methods}

\subsection{Probabilistic Sequence Model}\label{subsection:model}

We observe an $\mathcal{X}$-valued sequence $X_t$, $t=1,\ldots,n$, for some sample size $n\in\mathbb{N}$, where the unknown observation space $\mathcal{X}$ is a Borel subset of $\mathbb{R}^d$ for some fixed $d\in\mathbb{N}$. We consider the model  \begin{equation}\label{eqn:data_generating_process_conditional_distributions} X_{t+1} = G(\varepsilon_{t+1},S_t),\quad S_t = \tau(S_{t-1},X_t)=R(X_{-\infty:t}),\quad t\in\mathbb{Z},\end{equation} where $\varepsilon_{t}$, $t\in\mathbb{Z}$, is an iid sequence of latent noise inputs satisfying $\varepsilon_{t+1} \indep \mathcal{F}_t$ for all $t\in \mathbb{Z}$.  $S_t$, $t\in\mathbb{Z}$, is an $\mathcal{S}$-valued sequence of latent predictive states, where the unknown predictive state space $\mathcal{S}$ is a Borel subset of $\mathbb{R}^r$ for some ambient dimension $r\in\mathbb{N}$. $G$ is a generation function, $\tau$ is a transition function, and $R$ is a reduction function, all of which are unknown and Borel measurable.

\begin{remark}\label{rmk:future_work_exogenous_covariates}
The model from~\eqref{eqn:data_generating_process_conditional_distributions} includes many nonlinear processes with infinite-range memory, as well as finite-order Markov processes. It is straightforward to extend this model to include exogenous covariates, but we leave this extension for future work. 
\end{remark}


The conditional distribution of $X_{t+1}$ given $S_t$ is given by \begin{equation}\label{eqn:one_step_ahead_given_S_t} \mathcal{L}(X_{t+1} \mid S_t)=\mathcal{L}(X_{t+1} \mid \mathcal{F}_t),\quad t\in\mathbb{Z},\end{equation} almost surely, because $S_t=R(X_{-\infty:t})$ is $\mathcal{F}_t$-measurable and because $\varepsilon_{t+1} \indep \mathcal{F}_t$. By Doob's conditional
independence property (Theorem~8.9 in~\cite{Kallenberg-mod-prob}),~\eqref{eqn:one_step_ahead_given_S_t} is equivalent to the one-step-ahead predictive sufficiency property, \begin{equation}\label{eqn:pred_suff_property_CI}X_{t+1} \indep X_{-\infty:t} \mid S_t,\quad t\in\mathbb{Z}.\end{equation} In fact, $S_t$ has the strongest possible form of predictive sufficiency, as shown by Theorem~\ref{thm:one_step_implies_pred_suff_entire_future}. We give a more general result in Theorem~\ref{thm:one_step_implies_pred_suff_entire_future_standard_Borel_space} in the appendix for standard Borel spaces.

\begin{theorem}
\label{thm:one_step_implies_pred_suff_entire_future} The predictive states from~\eqref{eqn:data_generating_process_conditional_distributions} have the strong predictive sufficiency property,  
\begin{equation}
\label{eqn:entire_future_predictive_sufficiency_CI}
X_{t+1:\infty}
\indep
X_{-\infty:t}
\mid S_t,\quad t\in\mathbb{Z}.
\end{equation}
\end{theorem}

The \emph{prediction process} \citep{knight_prediction_process,aldous_prediction_process,lohr2010},  
\begin{equation}\label{eqn:prediction_process}\Pi_t=\mathcal{L}(X_{t+1:\infty} \mid \mathcal{F}_t), \quad t\in\mathbb{Z},\end{equation} characterizes the future values of a stochastic process given its past. By Theorem~\ref{thm:one_step_implies_pred_suff_entire_future} and the aforementioned Doob's conditional independence property, 
\begin{equation}\label{eqn:prediction_process_given_predictive_state}\Pi_t=\mathcal{L}(X_{t+1:\infty} \mid S_t),\quad t\in\mathbb{Z},\end{equation} almost surely, so by the chain rule, predictive sufficiency, and the recursive update in~\eqref{eqn:data_generating_process_conditional_distributions}, 
\begin{equation}\label{eqn:predictive_distribution_finite_L}\Pi_{t,L}=\mathcal{L}(X_{t+1:t+L} \mid \mathcal{F}_t)=\mathcal{L}(X_{t+1:t+L} \mid S_t)=\prod_{j=1}^L
\mathcal{L}(X_{t+j}\mid S_{t+j-1}), \quad t\in\mathbb{Z},\end{equation} almost surely, for arbitrary horizon $L\in\mathbb{N}$, where the product denotes iterated composition of conditional probability kernels. Hence, for multi-step generative sequence modeling, it suffices to characterize the one-step-ahead conditional distributions $\mathcal{L}(X_{t+1} \mid S_t)$; see Section~\ref{subsection:learning_conditional_distributions}.


Unfortunately, we do not observe $S_t$. We therefore aim to find an estimable, injective map $\Phi$ that sends the predictive state values $s\in\mathcal{S}$ to elements in some feature space $\Phi(s)\in\Phi(\mathcal{S})$. As we will show, this estimate of $\Phi(S_t)$ can be used to estimate the conditional distribution of $X_{t+1}$ given $S_t$.

\subsection{Predictive State Representations}\label{subsection:predictive_state_representations}

For some $m\in\mathbb{N}_0$, denote the conditional joint distribution of the next $m+1$ observations by \begin{equation}\label{eqn:cond_distr_m+1_steps_ahead} P_{S_t} = \mathcal{L}(X_{t+1:t+m+1} \mid \mathcal{F}_t)=\mathcal{L}(X_{t+1:t+m+1} \mid S_{t}),\quad t\in\mathbb{Z}.
\end{equation} 
The family of distributions $\{P_s:s\in\mathcal{S}\}$ must satisfy some identifiability conditions, so that every value $s\in\mathcal{S}$ corresponds to a unique distribution $P_s$. Specifically, we need $m$ to be large enough that every predictive state leads to a distinct $(m+1)$-step-ahead distribution.

We now present the ingredients used to injectively map $\mathcal{S}$ into $[-1,1]^k$ for some $k\geq 2p+1$. We randomly draw $k\geq 2p+1$ real-valued test functions $\varphi_1,\ldots,\varphi_k$ independently of the data from a distribution over a function class. Let $x_1,\ldots,x_{m+1}\in\mathcal{X}$ and define $x=(x_1,\ldots,x_{m+1})^{\top}\in\mathcal{X}^{m+1}$. In particular, we use random Fourier features \citep{Rahimi_Recht_2007_random_features} of the form  
\begin{equation}\label{eqn:random_Fourier_features}\varphi_{i}(x)=\cos\left(\sum_{j=1}^{m+1}\Omega_{i,j} \cdot x_j + \alpha_i\right), \end{equation} where $\Omega_{i,j} \overset{\mathrm{iid}}{\sim} N(0,I_d)$ and $\alpha_i \overset{\mathrm{iid}}{\sim} U(-\pi,\pi)$ for $i=1,\ldots,k$ and $j=1,\ldots,m+1$. Denote $\varphi=(\varphi_1,\ldots,\varphi_k):\mathcal{X}^{m+1} \xrightarrow[]{} [-1,1]^k$ and let \begin{equation}\label{eqn:target_Yt+1_varphi_Xt+1}
Y_{t+1}=\varphi(X_{t+1:t+m+1}),\quad t\in\mathbb{Z}.
\end{equation}

Consider the mapping $\Phi:\mathcal{S}\xrightarrow[]{}[-1,1]^k$ defined as \begin{equation}\label{eqn:Phi_expectation_random_features} \Phi(s)=\int_{\mathcal{X}^{m+1}}\varphi(x) dP_s(x).\end{equation} 
For each $t\in\mathbb{Z}$, the predictive states $S_t$ are mapped into feature space $\Phi(S_t)$ via the quantity 
\begin{equation}\label{eqn:Phi_S_t_conditional_expectation_random_features} Z_t=\Phi(S_t)=\int_{\mathcal{X}^{m+1}}\varphi(x) d\mathcal{L}(X_{t+1:t+m+1} \mid \mathcal{F}_t)(x)=\mathbb{E}[Y_{t+1}\mid \mathcal{F}_t], 
\end{equation} where the first and third equalities are by definition, and the second is by~\eqref{eqn:Phi_expectation_random_features} and Theorem~\ref{thm:one_step_implies_pred_suff_entire_future}. 

To guarantee that the mapping $\Phi$ is (almost surely) one-to-one, we must limit the geometric complexity of the predictive state space $\mathcal{S}$. In the simplest case, $\mathcal{S}$ can be a Cartesian product of closed, bounded intervals, i.e., for real numbers $a_j < b_j$, $j=1,\ldots,r\equiv p$, we can take $$\mathcal{S}=\prod_{j=1}^r [a_j,b_j] \subset \mathbb{R}^r.$$  More generally, we only need to be able to characterize $\mathcal{S}$ by a finite number of polynomial equalities and inequalities. A more precise statement uses the notion of a ``semialgebraic set''; we simplify the definition from \cite{bochnak_coste_roy_real_algebraic_geometry} to Euclidean spaces.

\begin{definition}[Definition 2.1.4 in \cite{bochnak_coste_roy_real_algebraic_geometry}]\label{def:semialgebraic_set}
Let $r,a,b_1,\ldots,b_a \in \mathbb{N}$. A \underline{semialgebraic} subset of $\mathbb{R}^r$
is a subset of the form
$\bigcup_{i=1}^{a}\bigcap_{j=1}^{b_i}
\left\{x \in \mathbb{R}^r
\mid f_{i,j}(x) \mathbin{*_{i,j}} 0\right\}$, where $f_{i,j} \in \mathbb{R}[x_1,\ldots,x_r]$, $*_{i,j}$ is either
$<$ or $=$, for $i=1,\ldots,a$ and $j=1,\ldots,b_i$, and $\mathbb{R}[x_1,\ldots,x_r]$ is the ring of polynomials in
the variables $x_1,\ldots,x_r$ with coefficients in $\mathbb{R}$.
\end{definition}

We now have the materials needed to state our first assumption. Denote by $\psi_{s}(\omega)$ the characteristic function of $P_{s}$ evaluated at $\omega\in\mathbb{R}^{(m+1)\times d}$. The second condition is that $m\in\mathbb{N}_0$ from~\eqref{eqn:cond_distr_m+1_steps_ahead} is large enough to identify each $s\in \mathcal{S}$, so for two different predictive state values $s\neq s'$ we have $P_s\neq P_{s'}$. Note that we require the sample size to satisfy $n > 2m+1$.

\begin{assumption}\label{asmpt:identifiability_regularity_model}
The following conditions hold:
\begin{enumerate} 
    \item $\mathcal{S}$ is a compact, semialgebraic subset of $\mathbb{R}^{r}$ with Hausdorff dimension $\dim_H(\mathcal{S})=p$. 
    \item The map $s\mapsto P_{s}$ is one-to-one.  
    \item The real and imaginary parts of the map $(s,\omega)\mapsto \psi_{s}(\omega)$ each admit a jointly real analytic extension to $O \times \mathbb{R}^{(m+1)\times d}$, where $O$ is open and $\mathcal{S}\subset O\subset\mathbb{R}^r$. 
\end{enumerate}    
\end{assumption}

The following result establishes that $\Phi$ is one-to-one almost surely. The key ingredient is the finite witness theorem from \cite{amir_et_al_finite_witness_thm}. Note that $k\geq 2p+1$, not necessarily $k=2p+1$.

\begin{theorem}\label{thm:Phi_is_one_to_one}
Suppose Assumption~\ref{asmpt:identifiability_regularity_model} holds. Then, with probability one over the draw of $k\geq 2p+1$ random Fourier features $\varphi=(\varphi_1,\ldots,\varphi_k)$, the mapping $s\mapsto\Phi(s)$ is one-to-one.  
\end{theorem}

Going forward, we condition on a fixed realization of the random Fourier features. We make the following weak assumption, which holds with probability one by Theorem~\ref{thm:Phi_is_one_to_one}.

\begin{assumption}\label{asmpt:injective_Phi}
    Assume that the mapping $s\mapsto\Phi(s)$ is one-to-one.
\end{assumption}

The following result establishes the equality of the conditional distributions given $S_t$ and $Z_t$.
\begin{theorem}\label{thm:equality_of_cond_distr_St_Zt} Suppose Assumptions~\ref{asmpt:identifiability_regularity_model} and~\ref{asmpt:injective_Phi} hold. Then, almost surely, \begin{equation}\label{eqn:Q_Zt} Q_{Z_t}=\mathcal{L}(X_{t+1} \mid Z_t)=\mathcal{L}(X_{t+1} \mid S_t),\quad t\in\mathbb{Z}.\end{equation}
\end{theorem}

If the predictive state representations $(Z_t)_t$ from~\eqref{eqn:Phi_S_t_conditional_expectation_random_features} can be estimated well, then we can estimate the conditional distribution $Q_{z}=\mathcal{L}(X_{t+1} \mid Z_t=z)$ from~\eqref{eqn:Q_Zt} with a conditional distribution estimator using the estimates $(\hat{Z}_t)_t$ of $(Z_t)_t$. We explain how to learn the predictive state representations $(Z_t)_t$ using a recurrent neural network (RNN) in Section~\ref{subsubsection:rnn}, and we explain how to learn the conditional distribution $Q_{z}$ using a mixture density network (MDN) in Section~\ref{subsubsection:mdn}.


\subsection{Algorithms}\label{subsection:learning_conditional_distributions}

Algorithm~\ref{algo:cond_distr_estim_pred_states} presents our two-stage estimation method. In practice, the predictive state space's dimension $\dim_H(\mathcal{S})=p$ is unknown, so we treat the number $k$ of random Fourier features $\varphi$ as a hyperparameter to be selected from some candidate set $\mathcal{K}\subset\mathbb{N}$, $\min(\mathcal{K})\geq 3$. The value of $k\in\mathcal{K}$, $m\in\mathbb{N}_0$, and the estimator hyperparameters are selected jointly by optimizing a criterion. We are, here, agnostic about {\em which} criterion, though a form of cross-validated log-likelihood suggests itself.

\begin{algorithm}[H] 
\caption{Learning the one-step-ahead conditional distribution via predictive states} \label{algo:cond_distr_estim_pred_states}
\begin{algorithmic}[1]
    \State \textbf{Input:} Observations $(X_t)_{t=1}^n$, random Fourier feature generator, candidate set $\mathcal{K}$, conditional expectation estimator, and conditional distribution estimator 
    \State Select value of $k\in\mathcal{K}$, $m\in\mathbb{N}_0$, and estimator hyperparameters by optimizing a criterion
    \State $(\hat{Z}_t)_t \gets$ Fit and apply conditional expectation estimator of $(Z_t)_t$ from~\eqref{eqn:Phi_S_t_conditional_expectation_random_features}   
    \State  $\hat{Q}_{z}\gets$ Fit and apply conditional distribution estimator of $Q_z$ from~\eqref{eqn:Q_Zt} using $(\hat{Z}_t,X_{t+1})_t$    
    \State \textbf{Output:} Fitted estimators and estimates $\hat{Q}_{z}$, $(\hat{Z}_t)_t$ 
\end{algorithmic}
\end{algorithm}

Section~\ref{subsection:implmenetation_neural_networks} explains how to implement Algorithm~\ref{algo:cond_distr_estim_pred_states} using deep neural networks. Specifically, Section~\ref{subsubsection:rnn} implements Step 3 using an RNN, and Section~\ref{subsubsection:mdn} implements Step 4 using an MDN.

Algorithm~\ref{algo:gen_seq_modeling_via_pred_states} explains how to estimate functionals of the conditional distribution from~\eqref{eqn:predictive_distribution_finite_L} for arbitrary horizons, e.g., conditional quantiles in multi-step forecasting and conditional risk functionals in optimization. We will report the theory and experiments for conditional functional estimation in a separate manuscript. Note that $\delta_x$ is the Dirac measure at $x$.


\begin{algorithm}[H] 
\caption{Conditional functional estimation via predictive states} 
\label{algo:gen_seq_modeling_via_pred_states}
\begin{algorithmic}[1]
    \State \textbf{Input:} Inputs of Algorithm~\ref{algo:cond_distr_estim_pred_states}, number of simulations $N^{\mathrm{sim}}$, horizon $L$, and functional $\Gamma(\cdot)$
    \State Fitted estimators and estimates $\hat{Q}_{z},(\hat{Z}_t)_t \gets$ Run Algorithm~\ref{algo:cond_distr_estim_pred_states}
    \For{$j=1,\ldots,N^{\mathrm{sim}}$}
         \State Sample $\hat{X}_{n+1}^{(j)} \sim \hat{Q}_{\hat{Z}_{n}}$
        \For{$\ell=1,\ldots,L-1$}
            \State Predict $\hat{Z}_{n+\ell}^{(j)}$ based on $(X_{1:n},\hat{X}_{n+1:n+\ell}^{(j)})$ and sample $\hat{X}_{n+\ell+1}^{(j)} \sim \hat{Q}_{\hat{Z}_{n+\ell}^{(j)}}$
        \EndFor
    \EndFor
    \State Obtain estimate of conditional functional $\Gamma(\hat{\Pi}_{n,L})$, where $\hat{\Pi}_{n,L}=\frac{1}{N^{\mathrm{sim}}}\sum_{j=1}^{N^{\mathrm{sim}}} \delta_{\hat{X}_{n+1:n+L}^{(j)}}$ 
    \State \textbf{Output:} Estimate of conditional functional $\Gamma(\hat{\Pi}_{n,L})$
\end{algorithmic}
\end{algorithm}

\subsection{Implementation with Deep Neural Networks}\label{subsection:implmenetation_neural_networks}

We explain how to implement Algorithm~\ref{algo:cond_distr_estim_pred_states} using a recurrent neural network (RNN) as the conditional expectation estimator and a mixture density network (MDN) as the conditional distribution estimator. In Section~\ref{subsubsection:rnn}, we present the details for learning the predictive state representations $(Z_t)_t$ from~\eqref{eqn:Phi_S_t_conditional_expectation_random_features} using an RNN. In Section~\ref{subsubsection:mdn}, we show how to learn the conditional distributions $(Q_{Z_t})_t$ from~\eqref{eqn:Q_Zt}.

We consider an RNN and MDN as concrete examples because many readers are already familiar with these architectures. However, in general, other estimators can be used with the method from Algorithm~\ref{algo:cond_distr_estim_pred_states}. For example, other recurrent networks can be used in place of the RNN, and for high-dimensional sequences diffusion or flow matching models can be used in place of the MDN.

\subsubsection{Recurrent Neural Network}\label{subsubsection:rnn}

We obtain the RNN parameter estimates by minimizing the sum of squared prediction errors,
\begin{equation}\label{eqn:rnn_estimates}\begin{aligned}
    (\hat{\rho}(\cdot),\hat{W}_h, \hat{v}_h,\hat{\lambda}(\cdot))
    \in
    &\underset{(\rho(\cdot),W_h, v_h,\lambda(\cdot))\in\mathcal{M}_n^{\mathrm{RNN}}}{\arg\min}
    \sum_{t=m+1}^{n-m-1}
    \left\|
        Y_{t+1}-\lambda(H_{t+1})
    \right\|_2^2,\\
    & \text{subject to } H_{t+1} = \sigma_{v_h}(W_h H_t + \rho(I_{t+1})) \text{ and } H_{m+1} = 0,
\end{aligned}
\end{equation} where $I_{t+1}=(X_{t-m:t},Y_{t-m})$ and $\mathcal{M}^{\mathrm{RNN}}_n$ is the RNN function class (for the details, see Assumption~\ref{asmpt:rnn_function_class} in Appendix~\ref{section:theoretical_results_on_convergence_rates}). The learned predictive state representation, $\hat{Z}_t$ (i.e., the estimated counterpart of $Z_t$ from~\eqref{eqn:Phi_S_t_conditional_expectation_random_features}), is given by \begin{equation}\label{eqn:Zhat}\hat{Z}_t=\hat{\lambda}(\hat{H}_{t+1}),\end{equation} where $\hat{H}_{t+1}$ is the RNN hidden state estimate and $\hat{\lambda}(\cdot)$ is the RNN output map from~\eqref{eqn:rnn_estimates}.

We provide an explanation of how the RNN architecture works. At each step $t$, the input $I_{t+1}=(X_{t-m:t},Y_{t-m})$ is transformed by a deep neural network (DNN) $\rho(\cdot)$. This transformed input is then linearly combined with the linearly transformed current hidden state $W_h H_t$, which is passed through a shifted ReLU activation, $\sigma_{v_h}$, defined as $\sigma_{v_h}(x)=[\max(x_i-v_{h,i},0)]_{1\leq i \leq \mathrm{dim}(H_t)}$, to produce the next hidden state $H_{t+1}$. A second DNN, $\lambda(\cdot)$, maps $H_{t+1}$ to the output $O_{t+1}$. Figure~\ref{fig:rnn_architecture} (based on Figure~1 in \citet{xiu_rnn_ts}) illustrates the RNN architecture. 

\begin{figure}[!ht]
    \centering
    \begin{tikzpicture}[
    scale=0.7,
    transform shape,
    >=Stealth,
        line/.style={->, thick, draw=black!55},
        hidden/.style={
            draw=violet!70!black,
            fill=violet!15,
            rounded corners=8pt,
            minimum width=2.25cm,
            minimum height=0.72cm,
            thick
        },
        input/.style={
            circle,
            draw=orange!75!black,
            fill=orange!18,
            minimum size=0.95cm,
            thick
        },
        output/.style={
            circle,
            draw=teal!70!black,
            fill=teal!15,
            minimum size=0.95cm,
            thick
        }
    ]

    \node[hidden] (Hm1) at (0,0) {$H_{t-1}$};
    \node[hidden] (Ht)  at (3.5,0) {$H_t$};
    \node[hidden] (Hp1) at (7,0) {$H_{t+1}$};

    \node[input] (Im1) at (0,-1.85) {$I_{t-1}$};
    \node[input] (It)  at (3.5,-1.85) {$I_t$};
    \node[input] (Ip1) at (7,-1.85) {$I_{t+1}$};

    \node[output] (Om1) at (0,1.85) {$O_{t-1}$};
    \node[output] (Ot)  at (3.5,1.85) {$O_t$};
    \node[output] (Op1) at (7,1.85) {$O_{t+1}$};

    \node at (-3.0,0) {$\cdots$};
    \node at (10.0,0) {$\cdots$};

    \draw[line] (-2.35,0) -- (Hm1.west);
    \draw[line] (Hm1.east) -- node[below=4pt] {$W_h$} (Ht.west);
    \draw[line] (Ht.east)  -- node[below=4pt] {$W_h$} (Hp1.west);
    \draw[line] (Hp1.east) -- node[below=4pt] {$W_h$} (9.35,0);

    \draw[line] (Im1.north) -- node[right=7pt] {$\rho(\cdot)$} (Hm1.south);
    \draw[line] (It.north)  -- node[right=7pt] {$\rho(\cdot)$} (Ht.south);
    \draw[line] (Ip1.north) -- node[right=7pt] {$\rho(\cdot)$} (Hp1.south);

    \draw[line] (Hm1.north) -- node[right=7pt] {$\lambda(\cdot)$} (Om1.south);
    \draw[line] (Ht.north)  -- node[right=7pt] {$\lambda(\cdot)$} (Ot.south);
    \draw[line] (Hp1.north) -- node[right=7pt] {$\lambda(\cdot)$} (Op1.south);

    \end{tikzpicture}

    \caption{Illustration of the RNN architecture.}
    \label{fig:rnn_architecture}
\end{figure}

\subsubsection{Mixture Density Network}\label{subsubsection:mdn}

Recall $Q_z=\mathcal{L}(X_{t+1} \mid Z_t=z)$,  $z\in\mathcal{Z}=\Phi(\mathcal{S})$ from~\eqref{eqn:Q_Zt}. We assume, for all $z\in\mathcal{Z}$, $Q_{z}$ is absolutely continuous w.r.t.\ Lebesgue measure, and a jointly measurable version of its density exists \begin{equation}\label{eqn:conditional_density_Xt+1_given_Zt}
f_{X_{t+1}|Z_t}(x|z), \quad z\in\mathcal{Z}.\end{equation}

The following notation will be useful. For each $j=1,\ldots,d$, define $\tilde{\mathcal{Z}}_j=\mathcal{Z} \times \mathcal{X}_{1:j-1}$ and let \begin{equation}\label{eqn:notation_W} \tilde{Z}_{t,j} = (Z_t^{\top},X_{t+1,1},\ldots,X_{t+1,j-1})^{\top}, \quad \tilde{z}_j = (z^{\top},x_1,\ldots,x_{j-1})^{\top}\in \tilde{\mathcal{Z}}_j,\end{equation} 
where we define $\tilde{Z}_{t,1}=Z_t$ and $\tilde{z}_1=z\in \mathcal{Z}$. Denote the factorization of $f_{X_{t+1}|Z_t}(x|z)$ by
\begin{equation}\label{eqn:factorization_conditional_density_Xt+1_given_Zt}
f_{X_{t+1}|Z_t}(x|z)=
    \prod_{j=1}^d
    f_{X_{t+1,j}\mid \tilde{Z}_{t,j}}(x_j\mid \tilde{z}_j),
\end{equation}  where we assume there exist jointly measurable versions of $f_{X_{t+1,j}\mid \tilde{Z}_{t,j}}(x_j\mid \tilde{z}_j)$, $j=1,\ldots,d$. Note that, for any Borel set $B\in\mathcal{B}(\mathcal{X})$, we have, almost surely, that \begin{equation}\label{eqn:equality_PSt_integrated_density} Q_{Z_t}(B)=\mathcal{L}(X_{t+1}\mid S_t)(B)
=\mathcal{L}(X_{t+1}\mid Z_t)(B)
= \int_B f_{X_{t+1}\mid Z_t}(x|Z_t) dx.\end{equation}

Denote the learned predictive state representations concatenated with the first $j$ dimensions of the observation, $\hat{\tilde{Z}}_{t,j}$ (i.e., the estimated counterpart of $\tilde{Z}_{t,j}$ from~\eqref{eqn:notation_W}), by  \begin{equation}\label{eqn:notation_What} \hat{\tilde{Z}}_{t,j} = (\hat{Z}_t^{\top},X_{t+1,1},\ldots,X_{t+1,j-1})^{\top}, \quad j = 1,\ldots,d,\end{equation} 
where $\hat{Z}_t=\hat{\tilde{Z}}_{t,1}$ is the learned predictive state representation from~\eqref{eqn:Zhat} based on the RNN from Section~\ref{subsubsection:rnn}. The MDN estimator is obtained by maximizing the conditional log-likelihood
\begin{equation}\label{eqn:mdn_estimator}\begin{aligned}
\hat{f}(x|z)=
    \prod_{j=1}^d
    \hat{f}_{j}(x_j\mid \tilde{z}_j),\quad \hat{f}_{j} \in \underset{f_{j}\in\mathcal{M}_{n,j}^{\mathrm{MDN}}}{\arg\max}
     \sum_{t=m+1}^{n-m-1} \log
    f_{j}(X_{t+1,j}|\hat{\tilde{Z}}_{t,j}), \quad j=1,\ldots,d,
\end{aligned}\end{equation} where the argmax is taken over the MDN function class $\mathcal{M}_{n,j}^{\mathrm{MDN}}$ from Assumption~\ref{asmpt:mdn_model}. In practice, an optimization procedure is used.

The Monte Carlo estimator of the conditional distribution $Q_{Z_t}(B)$ from~\eqref{eqn:equality_PSt_integrated_density} based on the Gaussian MDN is given below. Denote the number of Monte Carlo simulations by $N^{\mathrm{sim}}$ and let $\tilde{X}_1,\ldots,\tilde{X}_{N^{\mathrm{sim}}}\overset{\mathrm{iid}}{\sim} \hat{f}(\cdot \mid \hat{Z}_t)$. Note that it is straightforward to sample from the estimated Gaussian MDN $\hat{f}(\cdot \mid \hat{Z}_t)$ by selecting a component according to the mixture weights (i.e., using a categorical distribution), then drawing a sample from the selected component's Gaussian distribution. The final conditional distribution estimator is given by 
\begin{equation}\label{eqn:final_cond_distr_estimate} \hat{Q}_{\hat{Z}_t}(B)=
 \frac{1}{N^{\mathrm{sim}}}\sum_{i=1}^{N^{\mathrm{sim}}} 1(\tilde{X}_i \in B) \xrightarrow[]{\mathrm{a.s.}} \int_B \hat{f}(x|\hat{Z}_t) dx,\quad N^{\mathrm{sim}}\to\infty.\end{equation}

\begin{remark}\label{remark:discussion_of_main_theoretical_results} As a bonus, we derive the convergence rate for an instantiation of our method based on the RNN estimator from \cite{xiu_rnn_ts} and the MDN estimator from \cite{Zhou_et_al_2023}. Due to the technical nature and length of this analysis, we defer these results to Appendix~\ref{section:theoretical_results_on_convergence_rates}. 
\end{remark}

\section{Experiments}
\label{section:experiments}

We implement Algorithm~\ref{algo:cond_distr_estim_pred_states} in PyTorch based on an RNN (Section~\ref{subsubsection:rnn}) and MDN (Section~\ref{subsubsection:mdn}) as described above. As a baseline, we compare with an MDN whose lag window is a hyperparameter, where the largest possible lag grows with $n$. As discussed in Section~\ref{subsubsection:purpose_of_work}, the purpose of these experiments is to compare with a reasonable baseline in controlled settings, rather than an exhaustive comparison with existing sequence models on benchmarks. This is a reasonable baseline because an estimator with a growing lag window can, under certain conditions, asymptotically approximate the one-step-ahead conditional distribution of a process with infinite-range memory.

\subsection{Synthetic Experiments}
\label{subsection:synthetic_experiments}

It is of course impossible to provide an exhaustive set of synthetic experiments that cover all situations. We consider the data generating processes (DGPs) below, where we vary transition function and dimensions. The matrices $A_j^S\in\mathbb{R}^{r\times r}$, $A_j^X\in\mathbb{R}^{r\times d}$, $j\in\{1,2,3,4\}$, and $A_j^O\in\mathbb{R}^{d\times r}$, $j\in\{3,4\}$, are chosen so that the process in each DGP has a unique stationary distribution. We use truncated normal conditional distributions with $B_{\mathrm{trunc}}=[-2.5,2.5]^d$. Note that the transition functions, $\tanh(\cdot)$ and $\cos(\cdot)$, are applied coordinate-wise.

\begin{figure}[h!]
    \centering
    \includegraphics[width=0.45\linewidth]
    {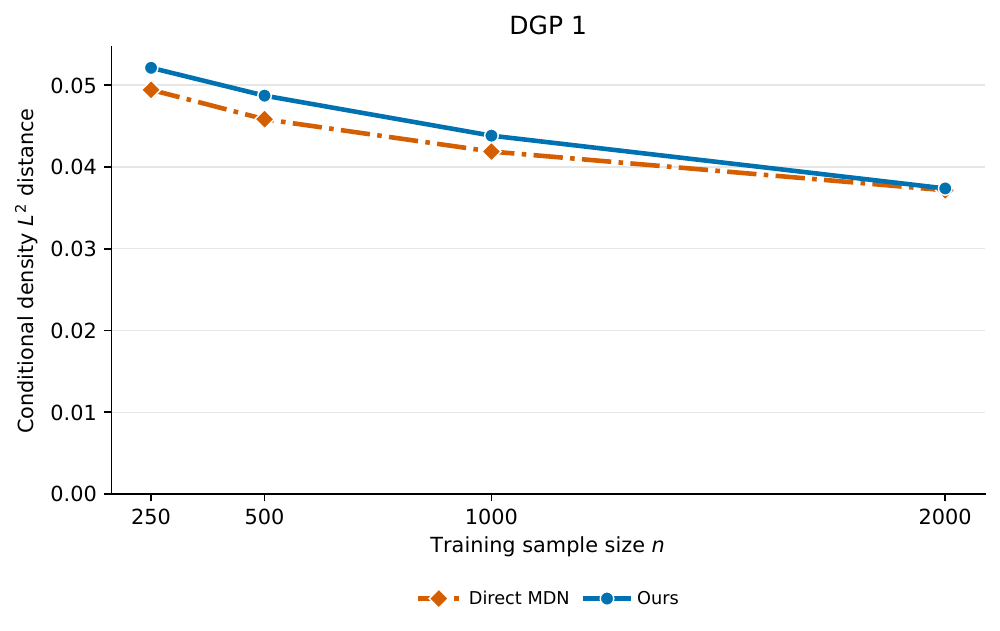}%
    \hfill
    \includegraphics[width=0.45\linewidth]
    {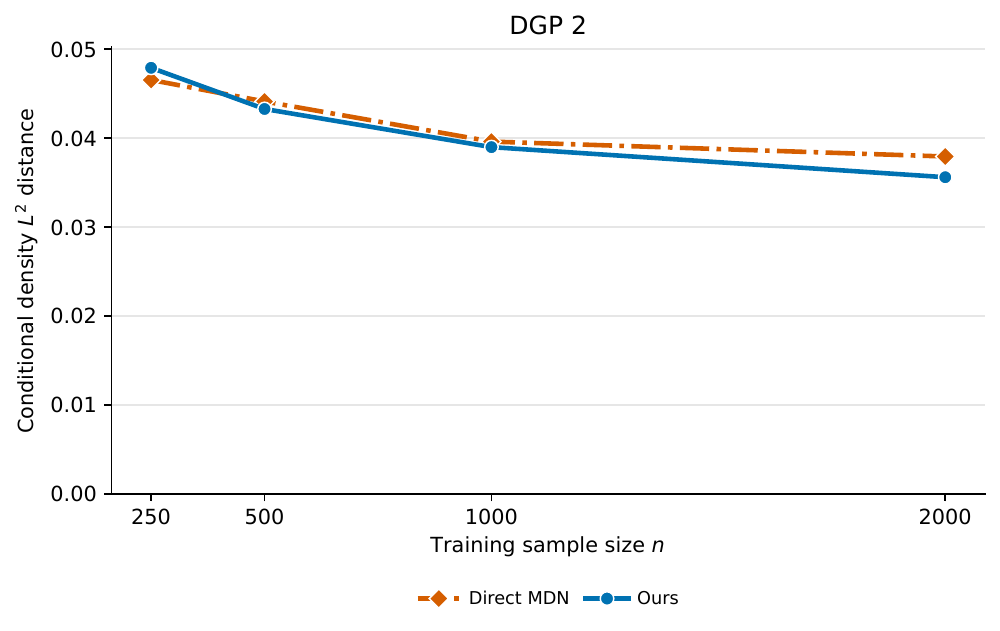}
    \caption{Average out-of-sample conditional density $L^2$ distance by training sample size $n$.}
    \label{fig:l2_distance_dgps_12}
\end{figure}

\begin{figure}[h!]
    \centering
    \includegraphics[width=0.45\linewidth]
    {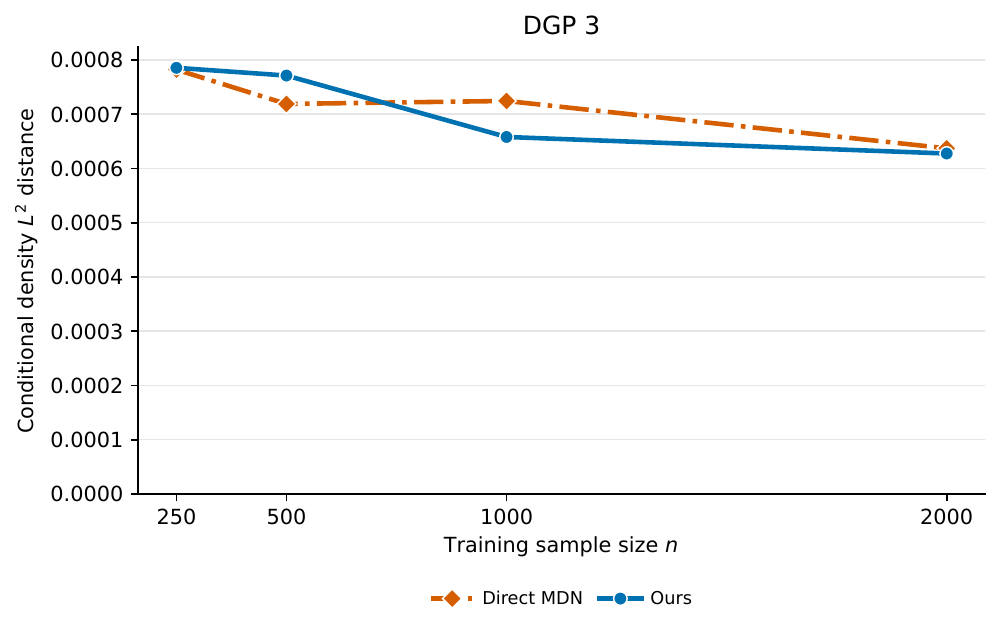}%
    \hfill
    \includegraphics[width=0.45\linewidth]
    {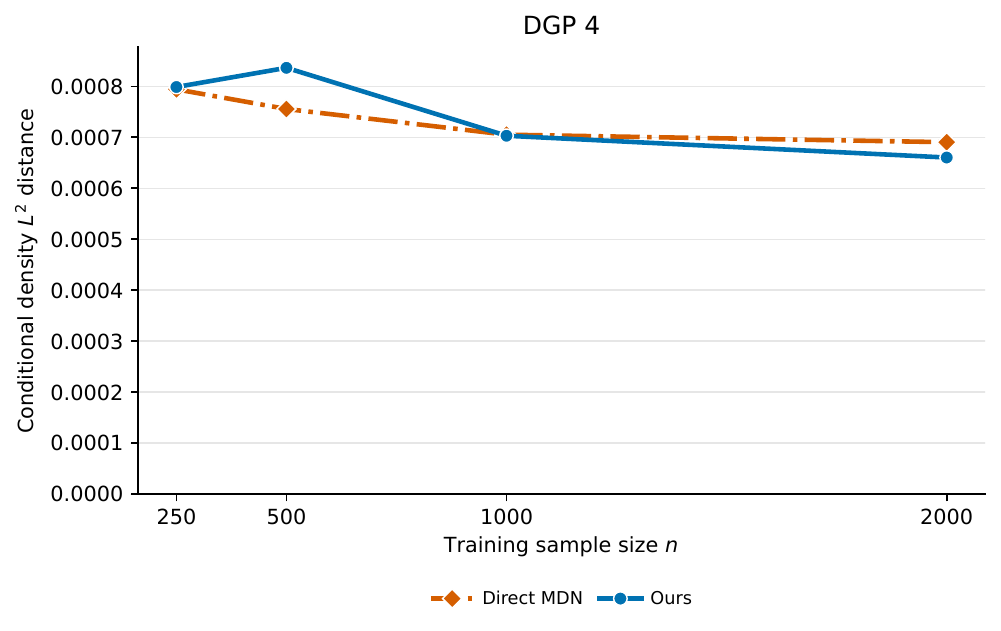}
    \caption{Average out-of-sample conditional density $L^2$ distance by training sample size $n$.}
    \label{fig:l2_distance_dgps_34}
\end{figure}

We fit each estimator (i.e., the ``Direct MDN'' baseline and ``Ours'') using times $t=1,\ldots,n$. Then, we calculate an averaged $L^2$ distance of the estimated conditional density and the true conditional density over times $t=n+1,\ldots,n+500$. With a slight abuse of notation, denote the true conditional density by $f_t$ and the estimated conditional density by $\hat f_t\in\{\hat f_t^{\mathrm{Ours}},\hat f_t^{\mathrm{MDN}}\}$. We calculate \begin{equation}\label{eqn:averaged_L2}\left[\frac{1}{500} \sum_{t=n}^{n+500-1}\int_{B_{\mathrm{trunc}}}(\hat f_t(x)-f_t(x))^2 dx\right]^{1/2}.\end{equation} Figures~\ref{fig:l2_distance_dgps_12} and~\ref{fig:l2_distance_dgps_34} show the value from~\eqref{eqn:averaged_L2} averaged over $100$ independent simulated trajectories at different training sample sizes $n$. We find that our method performs comparably to the MDN with a growing lag window.

First, we consider two DGPs with relatively low-dimensional observations.
$$\begin{alignedat}{3}
\textbf{DGP 1:}\quad X_{t+1} &\sim \mathrm{TN}(S_t,I_d; B_{\mathrm{trunc}}), &\quad S_t &= \mathrm{tanh}(A_1^S S_{t-1} + A_1^X X_t), &\quad r=p=d=3.\\
\textbf{DGP 2:}\quad X_{t+1} &\sim \mathrm{TN}(S_t,I_d; B_{\mathrm{trunc}}), &\quad S_t &=  \cos(A_2^S S_{t-1} + A_2^X X_t), &\quad r=p=d=3.
\end{alignedat}$$

Second, we consider two DGPs with moderate-dimensional observations.
$$\begin{alignedat}{3}
\textbf{DGP 3:}\quad X_{t+1} &\sim \mathrm{TN}(A_3^O S_t,I_d; B_{\mathrm{trunc}}), &\quad S_t &= \mathrm{tanh}(A_3^S S_{t-1} + A_3^X X_t), &\quad r=p=1, \  d=10.\\
\textbf{DGP 4:}\quad X_{t+1} &\sim \mathrm{TN}(A_4^O S_t,I_d; B_{\mathrm{trunc}}), &\quad S_t &=  \cos(\pi/2 + A_4^S S_{t-1} + A_4^X X_t), &\quad r=p=1, \  d=10.
\end{alignedat}$$

\subsection{Real Data Application: \textit{C. elegans} Movement Dynamics}
\label{subsection:real_data_application}

We consider a high-dimensional ($d=41$) time series of measurements from a single \textit{C. elegans} worm \citep{openworm_movement_data}. Each dimension captures a different aspect of the worm's body shape, posture, and movement over time. See \citet{openworm} for more details about OpenWorm.

We preprocess the raw data in the following ways. First, to handle missing values, we group the raw data into non-overlapping bins of three time-steps, and then we take the median of the available values in each bin. Second, we use the longest consecutive sequence of the binned data. Third, we standardize the binned data by dividing by the sample standard deviation of each dimension (based on the first $1000$ bins) and then by taking the first difference of the standardized binned data. After preprocessing, we obtain a time series of length $n=3395$.

Next, we describe how we apply and evaluate our method. We fit each conditional density estimator (``Direct MDN'' and ``Ours'') using the training window which consists of the observations at time-steps $t=1,\ldots,n$. For our estimator, we continue the fitted recurrence from the end of this training window through a fixed testing window which consists of the last $250$ observations of the time series.

\begin{figure}[h!]
    \centering
    \includegraphics[width=0.45\linewidth]{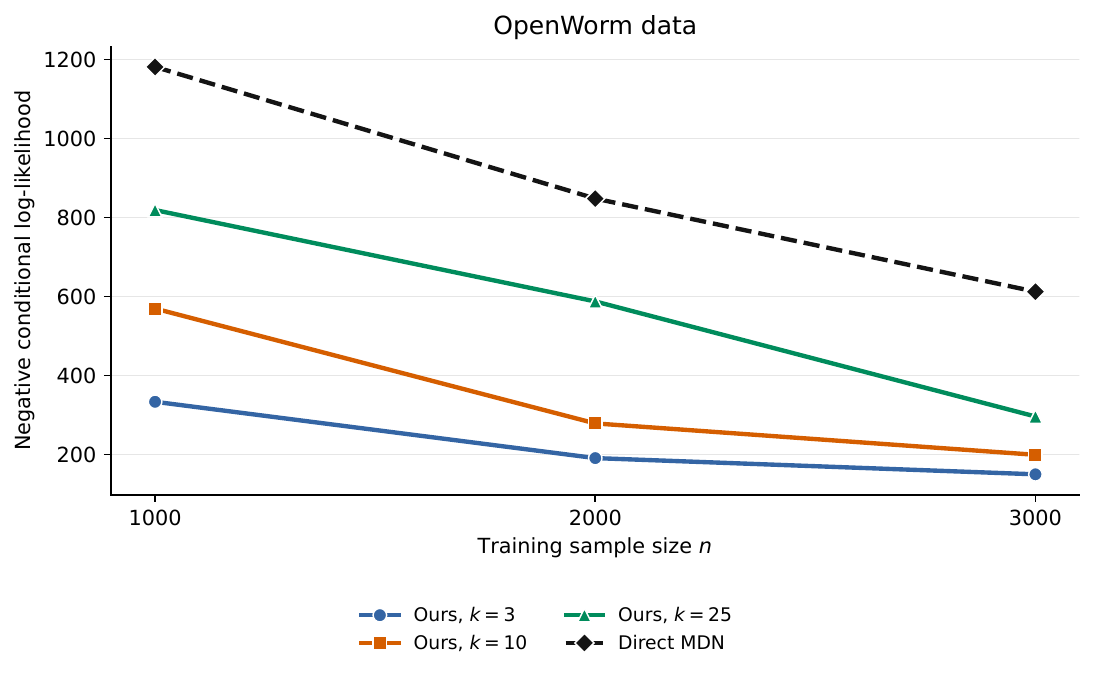}%
    \caption{Average out-of-sample negative conditional log-likelihood by training sample size $n$.}
    \label{fig:l2_distance_openworm}
\end{figure}

Since this is a real data example, we do not know the ``true'' conditional density function. Hence, we evaluate our method using the negative conditional log-likelihood of the next observation, and then we average these values over the testing window. Figure~\ref{fig:l2_distance_openworm} shows this value averaged over $5$ random seeds (i.e., for training the neural networks) at different training sample sizes $n$. 

\paragraph*{Our interpretation.} Interestingly, we find that the smallest number of random features ($k=3$) that we considered performs best. The results shown in Figure~\ref{fig:l2_distance_openworm} suggest that our estimator can take advantage of structure in the \textit{C. elegans} data that the baseline MDN estimator cannot. Since this is real data, we do not know the ground truth (so all interpretations should be taken with a grain of salt). Nevertheless, Figure~\ref{fig:l2_distance_openworm} is consistent with the explanation that the past information from the higher-dimensional observations ($d=41$) can be compressed into much lower-dimensional spaces, and that this has practical implications for estimating the conditional density. 

The \textit{C. elegans} movement dynamics are \textit{much} more complicated than the data generating processes from Section~\ref{subsection:synthetic_experiments}. This might be another reason why our method performs so much better than the baseline MDN estimator that only uses finitely many lags. However, this reason alone would not explain why the performance of our method improves as \textit{less} features are used, i.e., \textit{more} compression. Our interpretation is that our method takes advantage of intrinsic structure in the data which allows the past information to be compressed into a useful lower-dimensional representation.


\section{Conclusion and Future Work}
\label{section:conclusion_and_future_work}

We introduce a generative sequence modeling framework based on learning representations of the predictive states of a stochastic process. Despite the long history of predictive state representations, this paper is the first to theoretically study them in the context of generative sequence modeling. We implement our method in PyTorch using an RNN and MDN, as described in Sections~\ref{subsubsection:rnn} and~\ref{subsubsection:mdn}, respectively, and we conduct experiments with both real and synthetic data.

There are several promising directions for future work. First, it would be worthwhile to investigate applications of our predictive state representations in view of the discussion about identifiability from Section~\ref{subsubsection:comparison_with_existing}. Second, an important next step is to characterize how the performance of the generative sequence modeling procedure from Algorithm~\ref{algo:gen_seq_modeling_via_pred_states} scales with larger sample sizes and neural network architectures. Third, a fascinating direction would be to explore parallelization for predictive state space models by building on ideas from \citet{NEURIPS2024_0b2b199f,NEURIPS2025_1b7a2397,gonzalez2026unifyingframeworkparallelizingsequential,Kozachkov_2026}. Fourth, it would be interesting to consider language modeling with large predictive state space models.

\subsection*{AI use statement}


Starting from core ideas of the authors, we acknowledge that generative AI was used to: identify and search for relevant literature, brainstorm ideas involving the conceptual frameworks, suggest edits to the paper to improve prose, suggest edits to the proofs to check correctness, shorten length, and improve readability, suggest edits to software code for the method, experiments, and figures, suggest experimental parameters for experiments, identify possible gaps in the paper, provide feedback on the experiments, and generate the TikZ code for Figure~\ref{fig:rnn_architecture}. 

Generative AI was not used to: draft parts of the paper, format references, formulate the presented mathematical claims, provide the critical ingredients for proving mathematical claims,  suggest a structure for the paper, assist with translation, interpret results, generate synthetic data sets, or propose a title or keywords for the paper. 

Research subtasks not applicable to this work: propose or refine hypotheses, clean and reformat dataset, support qualitative and thematic data analysis, formulate questions for surveys or interviews, create artifacts,  transcribe recordings of research material.

The authors have reviewed all content suggested by generative AI and produced with the aid of generative AI. For example, the authors have reviewed all literature found by generative AI and checked the correctness of any suggestions made by generative AI. We take responsibility for the final content of this work, including text, claims or artifacts produced with the aid of generative AI.

\subsection*{Ethics statement}

All authors of this paper have read and adhere to the ICLR Code of Ethics. This work focuses on generative sequence modeling, which is used in a wide range of tasks, such as multi-step forecasting, optimization, and control. Our research does not involve human subjects or the collection of sensitive data.  The authors declare that there are no conflicts of interest.

\subsection*{Reproducibility statement}

Proofs of the theoretical results are in the appendix. Additional results and their proofs are also in the appendix. The code will be made publicly available upon acceptance to facilitate reproducibility and will be made available to reviewers during the review process.



\bibliography{iclr2027_conference}
\bibliographystyle{iclr2027_conference}

\appendix

\tableofcontents

\section{Theoretical Results on Convergence Rates}\label{section:theoretical_results_on_convergence_rates}

To help readers navigate this section and locate the important information, we provide this list here:
\begin{itemize}
    \item Section~\ref{subsection:assumptions_for_recurrent_neural_network_rnn_estimator} has results and assumptions related to the recurrent neural network (RNN).
    \item Section~\ref{subsection:assumptions_for_mixture_density_network_mdn_estimator} contains the assumptions related to the mixture density network (MDN).
    \item Section~\ref{subsection:stationarity_mixing_regularity_conditions} includes the assumptions involving the stationarity and decay of temporal dependence of the stochastic process, which are required by both the RNN and MDN. 
    \item Section~\ref{subsection:theoretical_results_on_convergence_rates} presents the theoretical convergence rate results.
\end{itemize}

\subsection{Assumptions for Recurrent Neural Network Estimator}\label{subsection:assumptions_for_recurrent_neural_network_rnn_estimator}

Define the noise process, \begin{equation}\label{eqn:xi_t+1}\xi_{t+1}=Y_{t+1} - Z_t,\quad t\in\mathbb{Z},\end{equation} which satisfies $\mathbb{E}[\xi_{t+1}\mid \mathcal{F}_t] =0$, so that the features $Y_{t+1}$ from~\eqref{eqn:target_Yt+1_varphi_Xt+1} can be written as
\begin{equation}\label{eqn:Yt+1_Zt_Xt+1}
    Y_{t+1} = Z_t +\xi_{t+1},\quad t\in\mathbb{Z}.
\end{equation} 
Note that $Y_t\in [-1,1]^k$ and $Z_t\in \mathcal{Z}=\Phi(\mathcal{S})\subset[-1,1]^k$, so $\xi_t\in [-2,2]^k$ by~\eqref{eqn:xi_t+1}.

Define $\tau^{(m+1)}$ as the recursive application of $\tau$ from~\eqref{eqn:data_generating_process_conditional_distributions} a total of $(m+1)$ times, so that
\begin{equation}\label{eqn:tau_m+1}
S_{t+m+1}
=
\tau^{(m+1)}(S_t,X_{t+1:t+m+1})=
\tau(\cdots
\tau(
\tau(S_t, X_{t+1}),
X_{t+2}
)
\cdots,
X_{t+m+1}
).
\end{equation}

The following result is used to establish the convergence rate guarantees of the RNN estimator.

\begin{lemma}\label{lma:conditional_expectation_as_function_of_noise_process} Suppose Assumptions~\ref{asmpt:identifiability_regularity_model} and~\ref{asmpt:injective_Phi} hold, and that the predictive states $S_t$, $t\in\mathbb{Z}$, satisfy the recursive update rule from~\eqref{eqn:data_generating_process_conditional_distributions}. Then, $Y_{t+1}$ from~\eqref{eqn:target_Yt+1_varphi_Xt+1} admits the representation
\begin{equation*}
    Y_{t+1}  = g^{\ast}(Y_{t-m},\xi_{t-m}, X_{t-m:t}) +\xi_{t+1},\quad t\in\mathbb{Z},
\end{equation*} where the noise process $\xi_{t+1}$ is from~\eqref{eqn:xi_t+1} and the function $g^{\ast}(\cdot):\mathcal{D}\xrightarrow[]{}\Phi(\mathcal{S})$ with domain $$\mathcal{D}=\{(y,e,x)\in [-1,1]^k\times [-2,2]^k \times \mathcal{X}^{m+1}:y-e\in \Phi(\mathcal{S}),\tau^{(m+1)}(\Phi^{-1}(y-e),x)\in\mathcal{S}\},$$ where $\tau^{(m+1)}$ is from~\eqref{eqn:tau_m+1}, is defined as \begin{equation}\label{eqn:definition_of_m} g^{\ast}(y,e,x)=\Phi(\tau^{(m+1)}(\Phi^{-1}(y-e),x)).
\end{equation} 
\end{lemma}

We require the following compactness condition on the unknown set $\mathcal{X}$ from~\eqref{eqn:data_generating_process_conditional_distributions}.

\begin{assumption}\label{asmpt:compact_X}
     Assume $X_t\in\mathcal{X}$ almost surely for every $t\in\mathbb{Z}$ for some compact set $\mathcal{X}\subset\mathbb{R}^d$.
     
 \end{assumption}

Let $\mathcal{X}_j$ be the $j$-th dimension of $\mathcal{X}$ for each $j\in [d]$, and denote $\mathcal{X}=\mathcal{X}_{1:d}$.


We require the definition of a H\"{o}lder ball of functions.

\begin{definition}
\label{defn:holder_ball}
Let $\eta, c>0$ and let $A$ be a subset of some Euclidean space. Define $\mathcal{H}^{\eta}(A,c)$ as
$$
\left\{
f:A\to\mathbb{R} \ \Bigg| \ 
\underset{\alpha:\norm{\alpha}_1\leq\tilde{\eta}}{\max} \ 
\underset{x\in A}{\sup} \ 
\left|D^\alpha f(x)\right|
+
\underset{\alpha:\norm{\alpha}_1=\tilde{\eta}}{\max} \ 
\underset{\substack{x,x'\in A\\x\neq x'}}{\sup} \ 
\frac{
    \left|D^\alpha f(x)-D^\alpha f(x')\right|
}{
    \norm{x-x'}_{2}^{\eta-\tilde{\eta}}
}
\leq c
\right\},
$$
where $\tilde{\eta}$ is the largest integer less than $\eta$.
For a vector-valued function
$f=(f_1,\ldots,f_k)^\top:A\to\mathbb{R}^k$, we write $f\in\left(\mathcal{H}^{\eta}(A,c)\right)^k$ if $f_j\in\mathcal{H}^{\eta}(A,c)$ for every $j=1,\ldots,k$.
\end{definition}

Next, we need a smoothness condition for $g^{\ast}(\cdot)$ from~\eqref{eqn:definition_of_m} analogous to Assumption 2 in \cite{xiu_rnn_ts}.

\begin{assumption}
\label{asmp:holder_smoothness}
Assume that there exist constants $b_{\mathrm{bnd}}\geq 4$ and $c_{\mathrm{H\ddot{o}l}}>0$ such that $\mathcal{X}\subseteq [-b_{\mathrm{bnd}},b_{\mathrm{bnd}}]^d$ and $g^{\ast}(\cdot)$ from~\eqref{eqn:definition_of_m} admits an extension $\tilde{g}^{\ast}(\cdot)$ to the domain $[-b_{\mathrm{bnd}},b_{\mathrm{bnd}}]^{2k+d(m+1)}$ that is $\eta$-H\"{o}lder smooth. That is, there exists some $\eta\in\mathbb{N}$, such that $$\tilde{g}^{\ast}(\cdot)\in
\left(
\mathcal{H}^{\eta}
\left([-b_{\mathrm{bnd}},b_{\mathrm{bnd}}]^{2k+d(m+1)},c_{\mathrm{H\ddot{o}l}}\right)
\right)^k.$$ 
\end{assumption}


The following invertibility assumption, analogous to Assumption 4 in \cite{xiu_rnn_ts}, is required.

\begin{assumption}
\label{eqn:invertibility}
For all $i,j\in [k]$, let $\tilde{g}_i^{\ast}(\cdot)$ be the $i$-th component
of $\tilde{g}^{\ast}(\cdot)$ and let
$$
\Delta_{i,j}
=\underset{(y,e,x)\in [-b_{\mathrm{bnd}},b_{\mathrm{bnd}}]^{2k+d(m+1)}}{\mathrm{ess}\ \mathrm{sup}}
\left|
\frac{\partial \tilde{g}_i^{\ast}(y,e,x)}{\partial e_j}
\right|,
$$ and assume that $\max_{i\in[k]} \sum_{j=1}^k\Delta_{i,j}<1$.
\end{assumption}

Next, we introduce the deep neural network (DNN) function class used in the RNN. Let $\sigma_b(x)
=
\left[\max(x_i-b_i,0)\right]_{i}$ be the shifted ReLU activation. For $a,b,\ell,w\in\mathbb{N}$ and $c_{\mathrm{bnd},1},c_{\mathrm{bnd},2}>0$, let
$\mathcal{M}_{n,a,b}^{\mathrm{DNN}}(\ell,w,c_{\mathrm{bnd},1},c_{\mathrm{bnd},2})$ denote the class of ReLU DNNs $f:\mathbb{R}^{a}\to\mathbb{R}^{b}$ of the form \begin{equation}\label{eqn:dnn_form}f(x)
=
A_{\ell}
\sigma_{b_{\ell}}
A_{\ell-1}
\sigma_{b_{\ell-1}}
\cdots
A_1
\sigma_{b_1}
A_0x,\end{equation} where the network has $\ell$ hidden layers, each hidden layer has
width at most $w$, and \begin{equation}\label{eqn:dnn_bounds}\max\left(
\max_{0\leq j\leq \ell}\norm{A_j}_{\infty},
\max_{1\leq j\leq \ell}\norm{b_j}_{\infty}
\right)
\leq c_{\mathrm{bnd},1},
\qquad
\norm{f}_{\infty}\leq c_{\mathrm{bnd},2}.\end{equation}

We now introduce the RNN function class. As in \cite{xiu_rnn_ts}, we use a single recurrent layer.

\begin{assumption}\label{asmpt:rnn_function_class}

Assume there exists $\epsilon_{\mathrm{RNN}}\in(0,1)$ such that, for every $n\in\mathbb{N}$, the RNN class is \begin{equation*}
\mathcal{M}^{\mathrm{RNN}}_n
=\left\{\begin{aligned}
&(\rho(\cdot),W_h,v_h,\lambda(\cdot)):
\rho(\cdot)
\in
\mathcal{M}_{n,d(m+1)+k,w_{h,n}}^{\mathrm{DNN}}
(\ell_n,w_n,c_{\mathrm{RNN},1},c_{\mathrm{RNN},2}),\\
&
\lambda(\cdot)
\in
\mathcal{M}_{n,w_{h,n},k}^{\mathrm{DNN}}
(\ell_n,w_n,c_{\mathrm{RNN},1},1),\quad 
W_h\in\mathbb{R}^{w_{h,n}\times w_{h,n}}
\text{ is upper triangular},\\
&
\norm{W_h}_{\infty}\leq c_{\mathrm{RNN},1},\quad
\norm{\mathrm{diag}(W_h)}_{\infty}\leq 1-\epsilon_{\mathrm{RNN}},\quad
\norm{v_h}_{\infty}\leq c_{\mathrm{RNN},1},
\end{aligned}\right\},
\end{equation*} where $c_{\mathrm{RNN},1}=c_{\mathrm{RNN},2}=n^{5\eta + 5}$ with $\eta$ from Assumption~\ref{asmp:holder_smoothness}. 
\end{assumption}

\subsection{Assumptions for Mixture Density Network Estimator}\label{subsection:assumptions_for_mixture_density_network_mdn_estimator}

To simplify the notation below, all suprema and infima are taken over the supports of the corresponding random variables, and the mixture weights satisfy $\sum_{g=1}^{M_j}\alpha_{g,j}(\tilde{z}_j)=1$ and $\alpha_{g,j}(\tilde{z}_j)\geq 0$.

The next assumption is analogous to Assumption 2 from \cite{Zhou_et_al_2023}. The first condition is that each conditional density function from~\eqref{eqn:factorization_conditional_density_Xt+1_given_Zt} can be approximated by a conditional Gaussian mixture model. The remaining conditions are standard boundedness and smoothness conditions.

\begin{assumption}\label{asmpt:density_regularity_conditions}
Assume that:
\begin{enumerate}
    \item There exist constants $\omega_1, c_{\mathrm{gmix}}>0$, such that, for every $j\in [d]$ and every $M_j\in\mathbb{N}$, there exist functions $\alpha_{g,j}=\alpha_{g,j}^{(M_j)}$, $\mu_{g,j}=\mu_{g,j}^{(M_j)}$, and $\sigma_{g,j}=\sigma_{g,j}^{(M_j)}$ satisfying 
    \begin{align*}
        \sup_{x_j,\tilde{z}_j}
        \left|
        f_{X_{t+1,j}|\tilde{Z}_{t,j}}(x_j|\tilde{z}_j)
        -
        \sum_{g=1}^{M_j}
        \frac{\alpha_{g,j}(\tilde{z}_j)}
        {\sqrt{2\pi}\sigma_{g,j}(\tilde{z}_j)}
        \exp\left(
        -\frac{(x_j-\mu_{g,j}(\tilde{z}_j))^2}
        {2\sigma_{g,j}^2(\tilde{z}_j)}
        \right)
        \right| \leq c_{\mathrm{gmix}} M_j^{-\omega_1}.
    \end{align*}

    \item There exists a constant $c_\mu>0$ such that for all $j\in [d]$,
    $$
        \max_{g\in[M_j]}
        \sup_{\tilde{z}_j}
        |\mu_{g,j}(\tilde{z}_j)|
        \leq c_\mu
        \quad\text{for every }M_j\in\mathbb{N},
    $$
    and there exist constants $c_{\sigma}^{+},c_{\sigma}^{-}>0$ and
    $\omega_2>0$ such that for all $j\in [d]$,
    $$
        c_{\sigma}^{-}M_j^{-\omega_2}
        \leq
        \min_{g\in[M_j]}\inf_{\tilde{z}_j}\sigma_{g,j}(\tilde{z}_j)
        \leq
        \max_{g\in[M_j]}\sup_{\tilde{z}_j}\sigma_{g,j}(\tilde{z}_j)
        \leq c_{\sigma}^{+} \quad\text{for every }M_j\in\mathbb{N}.
    $$

    \item There exist $\gamma\in\mathbb{N}$, $c_{\mathrm{Sob}}>0$ and $b_{\mathcal{X}}\geq 1$,
    such that $\mathcal{X} \subseteq [-b_{\mathcal{X}},b_{\mathcal{X}}]^d$ and, for all $j\in [d]$, $M_j\in\mathbb{N}$, and $g\in[M_j]$, the functions
    $\alpha_{g,j}(\cdot)$, $\mu_{g,j}(\cdot)$, and
    $\sigma_{g,j}(\cdot)$ admit continuous extensions $\tilde{\alpha}_{g,j}(\cdot)$, $\tilde{\mu}_{g,j}(\cdot)$, and $\tilde{\sigma}_{g,j}(\cdot)$, respectively, to the domain $B_j = [-b_{\mathcal{X}},b_{\mathcal{X}}]^{k+j-1}$ such that each extension lies in the Sobolev ball 
    $$
        \left\{
        h:
        \max_{\nu,\|\nu\|_1\leq\gamma}
        \underset{\tilde{z}_j\in B_j}{\mathrm{ess} \ \mathrm{sup}} \ |D^\nu h(\tilde{z}_j)|
        \leq c_{\mathrm{Sob}}
        \right\},
    $$
    where the maximum is taken over all $(k+j-1)$-dimensional
    $\mathbb{N}_0$-valued vectors $\nu$ whose entries sum to at most
    $\gamma$, and $D^\nu h$ denotes the weak derivative
    \citep{gine_nickl_2015}.

    \item There exists a constant $c_f^{-}>0$ so that, for all $j\in [d]$,
    $
        \inf_{x_j,\tilde{z}_j}
        f_{X_{t+1,j}|\tilde{Z}_{t,j}}(x_j|\tilde{z}_j)
        \geq c_f^{-}.
    $
\end{enumerate}
\end{assumption}

Note that, by parts 1 and 2 of Assumption~\ref{asmpt:density_regularity_conditions} with $M_j=1$, there exists a constant $c_f^{+}>0$ such that $$\sup_{x_j,\tilde{z}_j} f_{X_{t+1,j}|\tilde{Z}_{t,j}}(x_j|\tilde{z}_j)
 \leq c_f^{+}=c_{\mathrm{gmix}}+1/(\sqrt{2\pi}c_{\sigma}^{-})< \infty .$$

The next condition is analogous to Assumption 3 in \cite{Zhou_et_al_2023}. Note that we suppress the dependence of $\alpha_{g,j}(\cdot)$, $\mu_{g,j}(\cdot)$, and $\sigma_{g,j}(\cdot)$ on the neural network parameters to simplify the notation.

\begin{assumption}\label{asmpt:mdn_model}
Assume that there exist constants $c_{\mathrm{MDN}}>0$, $\kappa\in(0,1]$, and a sequence $(\Lambda_n)_{n\in\mathbb{N}}$, $\Lambda_n\geq1$, such that, for every $n\in\mathbb{N}$, $j\in[d]$, the MDN function class is
    \begin{align*}
     \mathcal{M}_{n,j}^{\mathrm{MDN}}
    =
    \left\{\begin{aligned}&
    f_j(x_j|\tilde{z}_j)
    =
    \sum_{g=1}^{M_j}
    \frac{\alpha_{g,j}(\tilde{z}_j)}
    {\sqrt{2\pi}\sigma_{g,j}(\tilde{z}_j)}
    \exp\left(
    -\frac{(x_j-\mu_{g,j}(\tilde{z}_j))^2}
    {2\sigma_{g,j}^2(\tilde{z}_j)}\right):\\
    &c_{\mathrm{MDN}}^{-1} \leq \inf_{x_j,\tilde{z}_j} f_j(x_j|\tilde{z}_j) \leq \sup_{x_j,\tilde{z}_j} f_j(x_j|\tilde{z}_j) \leq c_{\mathrm{MDN}},
    \qquad \\&
    \sup_{\tilde{z}_j,g}|\mu_{g,j}(\tilde{z}_j)|
    \leq c_{\mathrm{MDN}}, \ c_{\mathrm{MDN}}^{-1}M_j^{-\omega_2}
    \leq
    \inf_{\tilde{z}_j,g}\sigma_{g,j}(\tilde{z}_j)
    \leq
    \sup_{\tilde{z}_j,g}\sigma_{g,j}(\tilde{z}_j)
    \leq c_{\mathrm{MDN}},\\& \abs{\log f_j(x_j\mid z,u_j)-\log f_j(x_j\mid z',u_j)}
    \leq \Lambda_n\norm{z-z'}_{2}^\kappa\end{aligned}
    \right\}, 
    \end{align*}
    where the last condition holds for every $x_j\in\mathcal{X}_j$, $u_j\in\mathcal{X}_{1:j-1}$, and
    $z,z'\in [-1,1]^k$. The functions $\mu_{g,j}(\cdot)$ and
    $\sigma_{g,j}(\cdot)$ are parametrized by DNNs. Similarly, the mixture weights $\alpha_{g,j}(\cdot)$ are obtained by a DNN followed by a normalization $\alpha_{g,j}=a_{g,j}/\sum_{i=1}^{M_j}a_{i,j}$ to ensure $\sum_{g=1}^{M_j}\alpha_{g,j}(\tilde{z}_j)=1$, where the DNN outputs satisfy  $0\leq a_{i,j}(\tilde{z}_j)\leq c_a$ and $\inf_{\tilde{z}_j}\sum_{i=1}^{M_j}a_{i,j}(\tilde{z}_j) \geq c_{\alpha}$ for some $c_a, c_{\alpha} >0$. 
    \end{assumption}

The next assumption controls the rate of growth of the number of parameters in the MDN.
\begin{assumption}\label{asmpt:rate_of_growth_MDN_num_param}
    Recall $\gamma$ from Assumption~\ref{asmpt:density_regularity_conditions}. Assume the number of parameters in the $j$-th MDN is proportional to
    $$
        M_j^{(k+j-1+\gamma)/\gamma}
        n^{(k+j-1)/(2\gamma+k+j-1)}
        \log(M_jn),$$ and that the DNNs for the coefficients use ReLU activations where the number of hidden layers is $O\left(\log(\frac{M_j}{\epsilon_{n,j}})+1\right)$ and the numbers of weight parameters and hidden units are each $O\left(
M_j^{(k+j-1)/\gamma}
\epsilon_{n,j}^{-(k+j-1)/\gamma}
(
\log(\frac{M_j}{\epsilon_{n,j}})+1
)\right)$, where $\epsilon_{n,j}
=n^{-\gamma/(2\gamma+k+j-1)}$.
\end{assumption}

The next condition controls the average probability the Gaussian MDN assigns outside of the support $\mathcal{X}_j$. This is required because of the compactness from Assumption~\ref{asmpt:compact_X} and part (4) of Assumption~\ref{asmpt:density_regularity_conditions}.

\begin{assumption}\label{asmpt:control_MDN_prob_outside_support} Recall $\omega_1$ from Assumption~\ref{asmpt:density_regularity_conditions}. Assume there exists a constant $c_{\mathrm{tail}}>0$, such that, for all $j\in [d]$ and $M_j \in \mathbb{N}$, the functions $\alpha_{g,j}=\alpha_{g,j}^{(M_j)}$, $\mu_{g,j}=\mu_{g,j}^{(M_j)}$, and $\sigma_{g,j}=\sigma_{g,j}^{(M_j)}$ satisfy
\begin{equation*} \int_{\tilde{\mathcal{Z}}_j}\int_{\mathbb{R}\setminus\mathcal{X}_j} 
\sum_{g=1}^{M_j}
        \frac{\alpha_{g,j}(\tilde{z}_j)}
        {\sqrt{2\pi}\sigma_{g,j}(\tilde{z}_j)}
        \exp\left(
        -\frac{(x_j-\mu_{g,j}(\tilde{z}_j))^2}
        {2\sigma_{g,j}^2(\tilde{z}_j)}
        \right) dx_j
 dP_{\tilde{Z}_{t,j}}(\tilde{z}_j)
\leq c_{\mathrm{tail}}M_j^{-2\omega_1}.
\end{equation*}

\end{assumption}

The following high-level condition is used to establish the consistency of our two-stage estimator.

\begin{assumption}\label{asmpt:mdn_generated_covariate_stability}
Assume that there exists a constant $c_{\mathrm{aprx}}>0$, such that, for every $j\in [d]$, there exists an MDN   $g_{n,j}\in\mathcal{M}_{n,j}^{\mathrm{MDN}}$ that satisfies    \begin{align}
    \sup_{x_j,\tilde{z}_j}
    \abs{g_{n,j}(x_j\mid\tilde{z}_j)
    -f_{X_{t+1,j}\mid\tilde{Z}_{t,j}}(x_j\mid\tilde{z}_j)}
    &\leq c_{\mathrm{aprx}}a_{n,j},
    \label{eqn:stable_mdn_approximation}\\
    \int_{\tilde{\mathcal{Z}}_j}\int_{\mathbb{R}\setminus\mathcal{X}_j}
    g_{n,j}(x_j\mid\tilde{z}_j)dx_j
    dP_{\tilde{Z}_{t,j}}(\tilde{z}_j)
    &\leq c_{\mathrm{aprx}}a_{n,j}^2,
    \label{eqn:stable_mdn_tail}
    \end{align}
where $a_{n,j}=M_j^{-\omega_1}+M_j^{4\omega_2}\epsilon_{n,j}$ and $\epsilon_{n,j}=n^{-\gamma/(2\gamma+k+j-1)}$.
\end{assumption}

\subsection{Assumptions for Stationarity and Decay of Temporal Dependence}\label{subsection:stationarity_mixing_regularity_conditions}

The definition of a $\beta$-mixing process is required.

\begin{definition}\label{def:beta_mixing}
    Let $(\Omega, \mathcal{A}, \mathbb{P})$ be a probability space, and let $\mathcal{B}$, $\mathcal{C}$ be sub-$\sigma$-algebras of $\mathcal{A}$. Define       $$\beta(\mathcal{B},\mathcal{C}) = \mathbb{E}\left(\sup_{C\in\mathcal{C}}\abs{\mathbb{P}(C) - \mathbb{P}(C \mid \mathcal{B})}\right).$$ 
    For some process $V_t$, $t\in\mathbb{Z}$, let $\mathcal{F}_{t_1:t_2}=    \sigma(V_t: t_1 \leq t\leq t_2)$. Define the \underline{$\beta$-mixing coefficient} by
    \begin{equation*}  
        \beta(j) = \sup_{t\in\mathbb{Z}} \beta( \mathcal{F}_{-\infty:t},\mathcal{F}_{t+j:\infty}), \quad j\geq 1.
    \end{equation*} The process $V_t$, $t\in\mathbb{Z}$, is said to be $\beta$-mixing (or absolutely regular) if $\beta(j)\rightarrow 0$ as $j \rightarrow \infty.$
\end{definition} 

The mixing condition below is used to establish the consistency of the RNN and MDN estimators.

\begin{assumption}\label{asmpt:mixing} Assume that the process $(X_t,Y_t,Z_t,\xi_t)$, $t\in\mathbb{Z}$, is stationary and exponentially $\beta$-mixing, i.e., for some $c_{\mathrm{mix}}^1,c_{\mathrm{mix}}^2 >0$, we have $\beta(j)\leq c_{\mathrm{mix}}^1\mathrm{exp}(-c_{\mathrm{mix}}^2 j)$, $j\geq 1$. 
\end{assumption}

Sufficient conditions for the existence of a stationary distribution for the model from~\eqref{eqn:data_generating_process_conditional_distributions} have been studied by \cite{Doukhan_et_al_2023} with a coupling approach, and by \cite{chen_wu_AR_infinity_2016} using the physical dependence measure of \citet{wu_funct_dep_meas}. 


\subsection{Theoretical Results on Convergence Rates}\label{subsection:theoretical_results_on_convergence_rates}

We now present the convergence rate results for the estimator $\hat{f}$ from~\eqref{eqn:mdn_estimator}. 

\begin{theorem}\label{thm:rate_of_convergence} 
Suppose Assumptions~\ref{asmpt:identifiability_regularity_model}--\ref{asmpt:mixing}
hold and the number of random Fourier features satisfies $k\geq 2p+1$. Suppose the number of mixture components $M_j=M$ for
every $j\in[d]$. In the RNN estimator from~\eqref{eqn:rnn_estimates}, take the maximum width of the hidden layers in $\rho(\cdot)$ and $\lambda(\cdot)$ to satisfy
$$
w_n\asymp
n^{\max\left(
\frac{2k+d(m+1)}{4(2k+d(m+1))+2\eta},
\frac{2k+d(m+1)}{5(2k+d(m+1))+3}
\right)},
$$ the recurrent width $w_{h,n}=\dim(H_t)$ to satisfy $w_{h,n}\asymp w_n\log n$, and the feedforward depth to satisfy $2\leq\ell_n\leq C\log n$ for a constant $C>0$. Then, the estimator $\hat{f}$ from~\eqref{eqn:mdn_estimator} satisfies \begin{equation}\begin{aligned}\label{eqn:rate_of_our_estimator_L2}
\norm{\hat{f}-f_{X_{t+1}\mid Z_t}}_{L^2} &=
 \left(
\int_{\mathcal{Z}}
\int_{\mathcal{X}}
\left|
\hat{f}(x\mid z)
-
f_{X_{t+1}\mid Z_t}
(x\mid z)
\right|^2
dx
dP_{Z_t}(z)
\right)^{1/2} \\&=
 O_{\mathbb{P}}\Bigg(\begin{aligned}
& M^{-\omega_1}
+M^{\frac{\gamma+k+d-1}{2\gamma}+4\omega_2}
 n^{-\frac{\gamma}{2\gamma+k+d-1}}\log^3(nM)\\
& +
\Lambda_n^{1/2}
 n^{-\frac{\kappa}{4}\min\left(
 \frac{\eta}{\eta+4k+2d(m+1)},
 \frac{2k+d(m+1)+3}{10k+5d(m+1)+3}
 \right)}
 (\log n)^{11\kappa/4}
\end{aligned}\Bigg).
\end{aligned}\end{equation}
\end{theorem}

If $\Lambda_n$ from Assumption~\ref{asmpt:mdn_model} is $O(1)$, and the number of mixture components is an integer that satisfies
\begin{equation*}
M
\asymp
n^{
\frac{2\gamma^2}
{
(2\gamma+k+d-1)
\left(
(2\omega_1+1+8\omega_2)\gamma+k+d-1
\right)
}
},
\end{equation*}
then the estimator $\hat{f}$ from~\eqref{eqn:mdn_estimator} has the convergence rate 
\begin{equation*}
\norm{
\hat{f}-f_{X_{t+1}\mid Z_t}
}_{L^2}
=
O_{\mathbb{P}}\Bigg(\begin{aligned}&
n^{
-\frac{2\gamma^2\omega_1}
{
(2\gamma+k+d-1)
\left(
(2\omega_1+1+8\omega_2)\gamma+k+d-1
\right)
}
}
\log^3 n\\
&+
n^{-\frac{\kappa}{4}
\min\left(
\frac{\eta}{\eta+4k+2d(m+1)},
\frac{2k+d(m+1)+3}
{10k+5d(m+1)+3}
\right)}
(\log n)^{11\kappa/4}\end{aligned}
\Bigg).
\end{equation*}




Theorem~\ref{thm:rate_of_convergence} is a statement about $\hat{f}$ evaluated at the true predictive state representation, while in practice $\hat{f}$ is evaluated at the learned predictive state representation. Hence, we make a high-level RNN assumption to bound the out-of-sample error.

\begin{theorem}\label{thm:prediction_at_estimated_state}
Suppose the network choices and  assumptions from Theorem~\ref{thm:rate_of_convergence} hold, and that the rate from~\eqref{eqn:rate_of_our_estimator_L2} tends to zero. Fit the RNN and MDN estimators as in~\eqref{eqn:rnn_estimates}
and~\eqref{eqn:mdn_estimator}, respectively, using $X_{1:n}$.
For any fixed horizon $L\in\mathbb{N}$, let the prediction $\hat{Z}_{n+L-1}=\hat\lambda(\hat{H}_{n+L})$ be obtained by continuing the fitted recurrence with observed inputs through step $n+L-1$, without refitting the RNN.
Suppose, for some $C>0$ and all sufficiently large $n$, that the RNN satisfies
\begin{equation}\label{eqn:rnn_forecast_mse}
\mathbb{E}\norm{\hat{Z}_{n+L-1}-Z_{n+L-1}}_2^2
\leq
C n^{-\min\left(
\frac{\eta}{\eta+4k+2d(m+1)},
\frac{2k+d(m+1)+3}{10k+5d(m+1)+3}
\right)}(\log n)^{11}.
\end{equation}
Then, as the training sample size $n\to\infty$,
\begin{equation}\label{eqn:prediction_at_estimated_state_rate}
\begin{aligned}
&\left(
\int_{\mathcal{X}}
\left|
\hat{f}(x\mid\hat{Z}_{n+L-1})
-f_{X_{n+L}\mid Z_{n+L-1}}(x\mid Z_{n+L-1})
\right|^2dx
\right)^{1/2}\\
&=
O_{\mathbb{P}}\Bigg(
\Lambda_n^{\frac{k}{2\kappa+k}}
\left[
\begin{aligned}
&M^{-\omega_1}
+M^{\frac{\gamma+k+d-1}{2\gamma}+4\omega_2}
n^{-\frac{\gamma}{2\gamma+k+d-1}}\log^3(nM)\\
&+
\Lambda_n^{1/2}
n^{-\frac{\kappa}{4}\min\left(
\frac{\eta}{\eta+4k+2d(m+1)},
\frac{2k+d(m+1)+3}{10k+5d(m+1)+3}
\right)}
(\log n)^{11\kappa/4}
\end{aligned}
\right]^{\frac{2\kappa}{2\kappa+k}}
\Bigg).
\end{aligned}
\end{equation}
In particular, this error converges to zero in probability
if $\Lambda_n$ from Assumption~\ref{asmpt:mdn_model} is $O(1)$.
\end{theorem}

We suspect~\eqref{eqn:rnn_forecast_mse} can be proven by extending Lemma 11 in the supplement of \cite{xiu_rnn_ts} under more assumptions, but this is left for future work.

\section{Additional Theoretical Results}\label{section:additional_theoretical_results}

\subsection{Mixture Density Network}\label{subsection:mixture_density_network}

The following result extends the guarantees for the mixture density network (MDN) estimator from \cite{Zhou_et_al_2023} to the setting of covariates with lower-dimensional support. We must consider this extension because the predictive state representations $Z_t$ take values in $\Phi(\mathcal{S})\subset\mathbb{R}^k$ but the predictive state space $\mathcal{S}\subset \mathbb{R}^r$ has Hausdorff dimension $p$, where $k\geq 2p+1$. In particular, we do not assume $P_{\tilde{Z}_{t,j}}$ is absolutely continuous with respect to the Lebesgue measure. Originally, \cite{Zhou_et_al_2023} only needed to consider estimating the conditional distributions of $X_{t+1}$ given $X_t$ and $X_{t-1}$ given $X_t$, and they did not need to consider covariates with lower-dimensional support.

Recall the notation from Section~\ref{subsection:assumptions_for_mixture_density_network_mdn_estimator}. Also, for the following result, denote the supports of $\tilde{Z}_{t,j}$ and $X_{t+1,j}$ by $\tilde{\mathcal{Z}}_j$ and $\mathcal{X}_j$, respectively, the Lebesgue measure restricted to $\mathcal{X}_j$ by $\lambda_{\mathcal{X}_j}$, and the distribution of $\tilde{Z}_{t,j}$ by $P_{\tilde{Z}_{t,j}}$. Note that we use the superscript ``oracle'' to distinguish the estimators, $\hat{f}_{j}^{\mathrm{oracle}}$, $j\in [d]$, in Theorem~\ref{thm:mdn_rates} which use the true covariates $\tilde{Z}_{t,j}$, $j\in [d]$, from the feasible
estimators $\hat{f}_{j}$, $j\in [d]$, in Section~\ref{subsection:assumptions_for_mixture_density_network_mdn_estimator} which use the estimated covariates $\hat{\tilde{Z}}_{t,j}$, $j\in [d]$. When the covariates do not need to be estimated, the ``oracle'' in the superscript can be ignored. For the following result, denote the smallest and largest number of mixture components, respectively, by $$M_{-}=\min_{j\in [d]}(M_j),\quad M_{+}=\max_{j\in [d]}(M_j).$$

\begin{theorem}\label{thm:mdn_rates}
Suppose Assumptions~\ref{asmpt:compact_X},~\ref{asmpt:density_regularity_conditions},~\ref{asmpt:control_MDN_prob_outside_support},~\ref{asmpt:mixing} hold. Then, there exist MDN function classes
$\mathcal{M}_{n,j}^{\mathrm{MDN}}$, $j\in [d]$, satisfying
Assumptions~\ref{asmpt:mdn_model} and~\ref{asmpt:rate_of_growth_MDN_num_param}, such that the oracle conditional joint density estimator \begin{equation}\label{eqn:oracle_mdn_estimator}
\begin{aligned}
\hat{f}^{\mathrm{oracle}}=\prod_{j=1}^d \hat{f}_{j}^{\mathrm{oracle}},\quad
\hat{f}_{j}^{\mathrm{oracle}}
\in
\underset{f_j\in\mathcal{M}_{n,j}^{\mathrm{MDN}}}
{\arg\max}
\sum_{t=m+1}^{n-m-1}
\log
f_j
\left(
X_{t+1,j}\mid\tilde{Z}_{t,j}
\right), \quad j=1,\ldots,d,
\end{aligned}
\end{equation}

satisfies
\begin{equation}\label{eqn:mdn_rate_with_observed_covariates_joint}
\begin{aligned}
&
\left\|
\hat{f}^{\mathrm{oracle}}
-
f_{X_{t+1}\mid Z_t}
\right\|_{L^2}
\\
&=
\left(
\int_{\mathcal{Z}}
\int_{\mathcal{X}}
\left|
\hat{f}^{\mathrm{oracle}}
(x\mid z)
-
f_{X_{t+1}\mid Z_t}
(x\mid z)
\right|^2
dx
dP_{Z_t}(z)
\right)^{1/2}\\
&\leq
c
\left(
M_{-}^{-\omega_1}
+
M_{+}^{
\frac{\gamma+k+d-1}{2\gamma}
+
4\omega_2
}
n^{-\frac{\gamma}{2\gamma+k+d-1}}
\log^3(nM_{+})
\right),
\end{aligned}
\end{equation} for some constant $c>0$, with probability at least
$1-O(n^{-1})$.
\end{theorem}

\subsection{Predictive Sufficiency}\label{section:predictive_states}

The next result generalizes Corollary 2 of Theorem 1 in \cite{cosma_computational_mechanics} from discrete measurable spaces to standard Borel spaces. First, we introduce the general setting.

Let $(\mathcal{X},\mathfrak{X})$  and $(\mathcal{S},\mathfrak{S})$ be standard Borel spaces. Consider the model  \begin{equation}\label{eqn:data_generating_process_standard_Borel_conditional_distributions} X_{t+1} = G(\varepsilon_{t+1},S_t),\quad S_t = \tau(S_{t-1},X_t)=R(X_{-\infty:t}),\quad t\in\mathbb{Z},\end{equation} where $X_t$ is an $(\mathcal{X},\mathfrak{X})$-valued sequence, $\varepsilon_{t}$, $t\in\mathbb{Z}$, is an iid sequence of random elements satisfying $\varepsilon_{t+1} \indep \mathcal{F}_t$ for all $t\in \mathbb{Z}$, $S_t$, $t\in\mathbb{Z}$, is an $(\mathcal{S},\mathfrak{S})$-valued sequence, and the functions $G$, $\tau$, and $R$ are each Borel measurable. The conditional distribution of $X_{t+1}$ given $S_t$ is given by \begin{equation}\label{eqn:Borel_space_one_step_ahead_given_S_t} \mathcal{L}(X_{t+1} \mid S_t)=\mathcal{L}(X_{t+1} \mid \mathcal{F}_t),\quad t\in\mathbb{Z},\end{equation} because $S_t=R(X_{-\infty:t})$ is $\mathcal{F}_t$-measurable and because $\varepsilon_{t+1} \indep \mathcal{F}_t$. By Doob's conditional
independence property (Theorem~8.9 in~\cite{Kallenberg-mod-prob}),~\eqref{eqn:Borel_space_one_step_ahead_given_S_t} is equivalent to one-step-ahead predictive sufficiency, \begin{equation}\label{eqn:Borel_space_pred_suff_property_CI}X_{t+1} \indep X_{-\infty:t} \mid S_t,\quad t\in\mathbb{Z}.\end{equation}

\begin{theorem}\label{thm:one_step_implies_pred_suff_entire_future_standard_Borel_space} The predictive states from~\eqref{eqn:data_generating_process_standard_Borel_conditional_distributions} have the strong predictive sufficiency property,  
\begin{equation}
\label{eqn:entire_future_predictive_sufficiency_CI_standard_Borel_spaces}
X_{t+1:\infty}
\indep
X_{-\infty:t}
\mid S_t,\quad t\in\mathbb{Z}.
\end{equation}

\end{theorem}

\section{Preliminaries}\label{section:preliminaries}


We state several definitions and results from the literature that we use in our proofs. We mostly retain the notation from those papers, and we refer to them for proofs.

\begin{figure}[h!]
  \centering \includegraphics[width=1\linewidth]{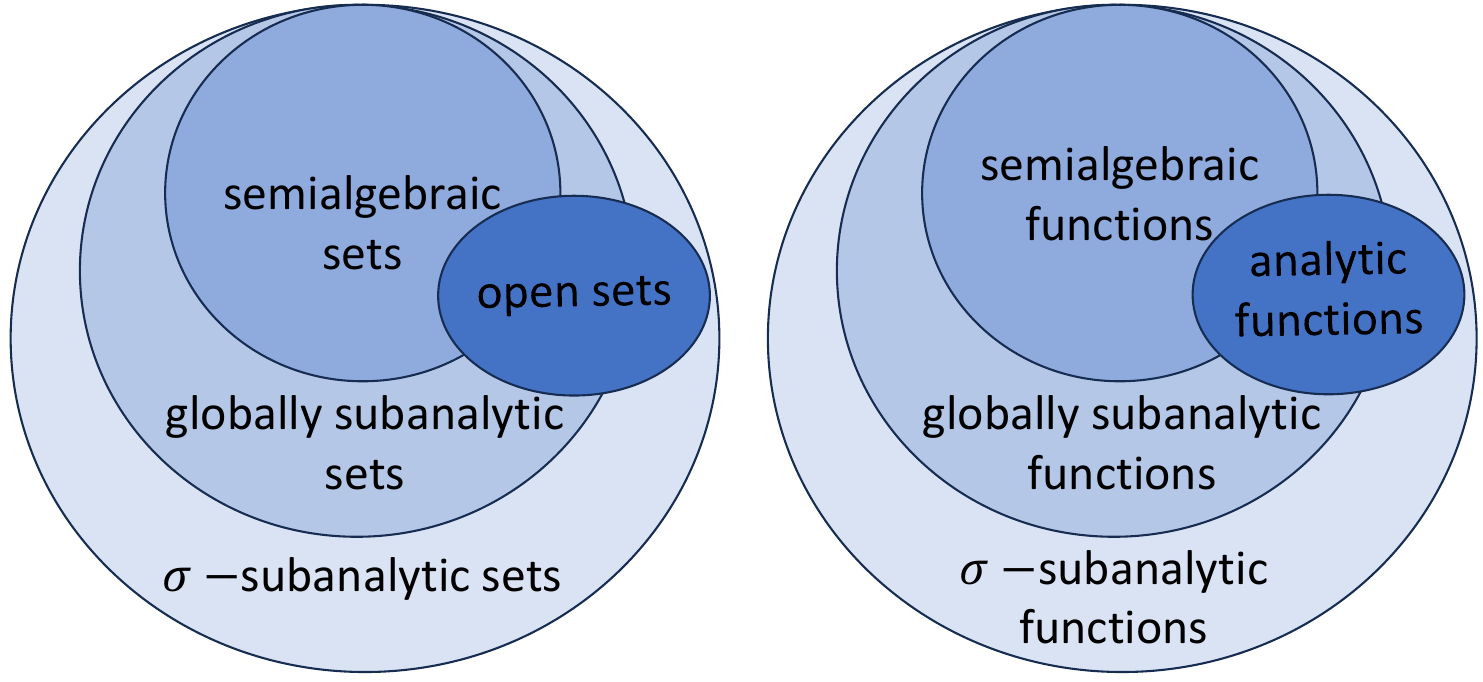}
  \caption{Relationships among classes of sets and functions (Figure 3 in \cite{amir_et_al_finite_witness_thm}).}\label{fig:diagram_of_subanalytic_sets_and_functions} 
\end{figure}

First, we present the definitions of $\sigma$-subanalytic sets and $\sigma$-subanalytic functions. These definitions require the definitions of globally subanalytic sets, globally semianalytic sets, and semianalytic sets. Figure~\ref{fig:diagram_of_subanalytic_sets_and_functions} (Figure 3 in \cite{amir_et_al_finite_witness_thm}) summarizes the relationships among the different classes of sets and functions; see the appendix of \cite{amir_et_al_finite_witness_thm} for more discussion.

\begin{definition}[Definition 1.1.1 in \cite{Valette_subanalytic}]
	A subset $E\subseteq \mathbb{R}^n$ is called \underline{semianalytic} if it is locally defined
	by finitely many real analytic equalities and inequalities. Namely, for each $a\in \mathbb{R}^n$,
	there is a neighborhood $U$ of $a$, and real analytic functions $f_{ij}, g_{ij}$ on $U$, where
	$i=1,\ldots,r$ and $ j= 1,\ldots s_i$, such that
	\begin{equation}\label{eq:semi}
	E\cap U=\bigcup_{i=1}^r \bigcap_{j=1}^{s_i}\{\mathbf{z}\in U|  g_{ij}(\mathbf{z})>0 \text{ and } f_{ij}(\mathbf{z})=0 \}. \end{equation}
\end{definition}

\begin{definition}[Definition 1.1.3 in \cite{Valette_subanalytic}]
	A subset $Z\subseteq \mathbb{R}^n$ is \underline{globally semianalytic} if $V_n(Z)$ is a semianalytic
	subset of $\mathbb{R}^n$, where $V_n : \mathbb{R}^n \xrightarrow[]{} (-1,1)^n$ is the homeomorphism defined by
	$$V_n\left(\mathbf{z}=(z_1,\ldots,z_n)\right)=\left(\frac{z_1}{\sqrt{1+|\mathbf{z}|^2}},\ldots , \frac{z_n}{\sqrt{1+|\mathbf{z}|^2}} \right). $$
\end{definition}

\begin{definition}[Definition 1.1.6 in \cite{Valette_subanalytic}]
	A subset $E\subseteq \mathbb{R}^n$ is \underline{globally subanalytic} if it can be presented
	as a linear projection of a globally semianalytic set; more precisely, if there
	exists a globally semianalytic set $Z \subseteq \mathbb{R}^{n+p}$  such that $E=\pi(Z) $, with   $\pi:\mathbb{R}^{n+p}\to \mathbb{R}^n$ being the projection operator that omits the last $p$ coordinates while leaving the remaining coordinates unchanged.
\end{definition}

\begin{definition}[Definition A.10 in \cite{amir_et_al_finite_witness_thm}]\label{def:sigma_subanalytic_set}
We say that a subset $A\subseteq\mathbb{R}^D$ is \underline{$\sigma$-subanalytic} if it is a countable union of globally subanalytic subsets of $\mathbb{R}^D$.
\end{definition}

\begin{definition}[Definition A.13 in \cite{amir_et_al_finite_witness_thm}]\label{def:sigma_subanalytic_function}
Let $\mathbb{M}\subseteq \mathbb{R}^D$ be a $\sigma$-subanalytic set. We say that $f:\mathbb{M}\xrightarrow[]{}\mathbb{R}^L$ is a \underline{$\sigma$-subanalytic function} if its graph is a $\sigma$-subanalytic subset of $\mathbb{R}^{D+L}$.
\end{definition}

Second, we state basic properties of $\sigma$-subanalytic sets.

\begin{lemma}[Proposition A.11 in \cite{amir_et_al_finite_witness_thm}]\label{lma:sigma_subanalytic_properties}
	Assume that $A,B \subseteq \mathbb{R}^D $ and $C \subseteq \mathbb{R}^M$ are $\sigma$-subanalytic sets. Then:
	\begin{enumerate}
		\item $A \cup B$ is $\sigma$-subanalytic. More generally, any countable union of $\sigma$-subanalytic sets is $\sigma$-subanalytic.
        \item $A\cap B$ is $\sigma$-subanalytic.
		\item $A\times C $ is $\sigma$-subanalytic.
		\item If $\pi:\mathbb{R}^D\to \mathbb{R}^L $ is a linear projection, then $\pi(A) $ is a $\sigma$-subanalytic set.
		\item $A $ is a \emph{countable} union of $C^\infty $ manifolds.
	\end{enumerate}
\end{lemma}

Third, we present a corollary of the finite witness theorem (Theorem A.2) from \cite{amir_et_al_finite_witness_thm}. The notion of dimension that is used is the Hausdorff dimension. The result holds for almost every $\left( \vartheta_{1},\ldots,\vartheta_{2D+1}\right)$ with respect to the Lebesgue measure when $\mathbb{W}=\mathbb{R}^q$ or is an open subset of $\mathbb{R}^q$.

\begin{lemma}[Corollary A.20 in \cite{amir_et_al_finite_witness_thm}]\label{lma:finite_witness}
Let $\mathbb{M} \subseteq \mathbb{R}^r$, $\mathbb{W} \subseteq \mathbb{R}^q$ be $\sigma$-subanalytic sets of dimension $D$ and $D_{\vartheta}$ respectively. Let $F:\mathbb{M} \times \mathbb{W} \to \mathbb{R}$ be a $\sigma$-subanalytic function. Define the set
$$\mathcal{N}=\{\left(\mathbf{x},\mathbf{y}\right) \in \mathbb{M} \times \mathbb{M} \mid F(\mathbf{x};\vartheta)=F(\mathbf{y};\vartheta),\ \forall \vartheta\in \mathbb{W} \}.$$
Suppose that for all $\left(\mathbf{x},\mathbf{y}\right) \in \mathbb{M} \times \mathbb{M} \setminus \mathcal{N}$
\begin{equation}\label{cond_dimension_defficiency_for_separation}
\dim\{\vartheta\in \mathbb{W}\mid F(\mathbf{x};\vartheta)=F(\mathbf{y};\vartheta) \}\leq D_{\vartheta}-1.\end{equation}
Then for generic $\left( \vartheta_{1},\ldots,\vartheta_{2D+1}\right) \in \mathbb{W}^{2D+1}$,
\begin{equation}\label{eqn:equality_of_functions_2D+1}
\mathcal{N}=\{\left(\mathbf{x},\mathbf{y}\right) \in \mathbb{M} \times \mathbb{M} \mid F(\mathbf{x};\vartheta_{i})=F(\mathbf{y};\vartheta_{i}),\ \forall i=1,\ldots 2D+1\} .
\end{equation}
Moreover, if $\mathbb{W}$ is an open and connected subset of $\mathbb{R}^q$, and $F(\mathbf{x};\vartheta)$ is analytic as a function of $\vartheta$ for all fixed $\mathbf{x} \in \mathbb{M}$, then~\eqref{cond_dimension_defficiency_for_separation} is not required, as it is automatically satisfied.
\end{lemma}

The next result is the uniqueness theorem for characteristic functions.

\begin{lemma}[Proposition 2.9 in \cite{multivariate_stats_book}]\label{lma:uniqueness_characteristic_function} The characteristic function $c:\mathbb{R}^n \to \mathbb{C}$ of $\mathbf{x}$, defined by
$c(\mathbf{t})
=
c_{\mathbf{x}}(\mathbf{t})
=
\mathbb{E}(e^{i\mathbf{t}'\mathbf{x}})$,
is unique: $$\mathbf{x} \overset{d}{=} \mathbf{y}
\quad\iff\quad
c_{\mathbf{x}}(\mathbf{t})
=
c_{\mathbf{y}}(\mathbf{t}),
\quad
\forall \mathbf{t}\in\mathbb{R}^n.$$
\end{lemma}

The following elementary result is stated as Theorem 4.17 in \cite{rudin_2021}.
\begin{lemma}\label{lma:thm_4_17_rudin}
Suppose $f$ is a continuous $1$-$1$ mapping of a compact metric space $X$ onto a metric space $Y$. Then the inverse mapping $f^{-1}$ defined on $Y$ by
$$
f^{-1}(f(x)) = x, \quad x \in X,
$$
is a continuous mapping of $Y$ onto $X$.
\end{lemma}

The following result is an approximation bound for deep neural networks (DNNs) stated as Lemma~2 in \cite{Zhou_et_al_2023}. For the following result, let $W^{\gamma,\infty}([-1,1]^d)$ denote the Sobolev space of functions whose weak derivatives up to order $\gamma$ are bounded, as in Assumption~\ref{asmpt:density_regularity_conditions}.
\begin{lemma} \label{lma:lemma2_zhou_approx_dnn}
	There exists a DNN class $\mathcal{M}^{\mathrm{DNN}}$ with the ReLU activation, such that, for any $0< \epsilon\leq 1$, 
	\begin{enumerate}
		\item $\mathcal{M}^{\mathrm{DNN}}$ approximates $W^{\gamma,\infty}([-1,1]^d)$, in the sense that, for any $g^* \in W^{\gamma,\infty}([-1,1]^d)$, there exists a $g_{\epsilon} \in \mathcal{M}^{\mathrm{DNN}}$ and $c_1>0$, such that $\|g_{\epsilon}-g^*\|_{\infty} \leq \epsilon$. 
		\item $H(\epsilon) \leq c_1 (\log(\frac{1}{\epsilon}) +1)$ and $W(\epsilon), U(\epsilon) \leq c_1 \cdot \epsilon^{-\frac{d}{\gamma}}(\log (\frac{1}{\epsilon})+1)$, where $H$ denotes the number of layers, $W$ the number of weights, and $U$ the total number of hidden units.
	\end{enumerate}
\end{lemma}

The next result is Lemma~3 in \cite{Zhou_et_al_2023}, which establishes the approximation error bound for the MDN. Note that our Assumptions
~\ref{asmpt:density_regularity_conditions},~\ref{asmpt:mdn_model}, and~\ref{asmpt:rate_of_growth_MDN_num_param} are analogous to Assumptions 2 and 3 from \cite{Zhou_et_al_2023}. In the following result, $f^*$ is the true density.

\begin{lemma} \label{lma:lemma3_in_zhou_et_al_mdn_approx}
	Suppose Assumptions 2 and 3 from \cite{Zhou_et_al_2023} hold. Then for any $0 < \epsilon \leq 1$ and integer $G\ge 1$, there exists a set of DNN functions $\{(g_{g1},g_{g2},g_{g3})_{g=1}^{G}\}$ whose network architectures depend on $\epsilon$ such that
	\begin{enumerate} 
		\item $ \|f^* - f_T\|_{\infty} \leq C'_2 G^{4\omega_2} \epsilon  +O(G^{-\omega_1}) \;\; \textrm{ where } \; f_T= \sum_{g=1}^G g_{g1}(x) \frac{1}{\sqrt{2 \pi }g_{g3}(x)}e^{-\frac{(y-g_{g2}(x))^2}{2 g^2_{g3}(x)}}, $
		for some constant $C'_2>0$ that is independent of $\epsilon$;
		\item the number of hidden layers $H_{gj}$, the number of parameters $W_{gj}$, and the total number of hidden units $U_{gj}$ satisfy  
		\begin{enumerate}
			\item $H_{gj} \leq c_1 \{ \log(G/\epsilon) +1 \}$
			\item $W_{gj} \leq c_1 G^{\frac{d}{\gamma}} \ \epsilon ^{-\frac{d}{\gamma}}\{ \log (G / \epsilon)+1 \}$
			\item $U_{gj} \leq c_1 G^{\frac{d}{\gamma}} \ \epsilon ^{-\frac{d}{\gamma}}\{ \log (G / \epsilon)+1 \}$
		\end{enumerate}
	\end{enumerate}
\end{lemma}

The next result is Lemma~4 from \cite{Zhou_et_al_2023}, which connects the metric entropy between the MDN function class and DNN function class. 
\begin{lemma}  \label{lma:lemma4_zhou}
	Suppose Assumptions 2 and 3 from \cite{Zhou_et_al_2023} hold. Then there exists a constant $c_3>0$, such that 
	\begin{align*} 
		\log \mathcal{T}(\delta,\mathcal{M},\|\cdot\|_n)  \leq    G \log  \mathcal{T}\left( \frac{ \delta}{c_3 G^{3/2 + 3\omega_2}} ,\mathcal{M}^{\mathrm{DNN}},\|\cdot\|_n \right),
	\end{align*} where $G$ is the number of mixture components. 
\end{lemma}

The following result is based on Lemma~6 in \cite{dnn_est_inf} (Theorem 6 in \cite{Bartlett_et_al_2019} in the ReLU case) which is used to bound the complexity of the function class of DNNs. Note that $\rm Pdim$ denotes the pseudo VC dimension of the function class.

\begin{lemma}
	\label{lem:lemma_6_farrell}
	Consider a ReLU network architecture $\mathcal{M} = \mathcal{M}^{\rm DNN}(W, L, U)$, then the pseudo VC dimension satisfies
	$$
		{\rm Pdim}(\mathcal{M}) \leq C \cdot WL \log W,
	$$
	for some universal constant $C>0$. 
\end{lemma}

The next result is Lemma~5 in \cite{Zhou_et_al_2023}.

\begin{lemma} \label{lma:beta_mixing_decoulping_lemma5_zhou}
	Suppose there is a sequence of stationary time series, $U = (X_1,X_2,\ldots,X_T)$, with a common marginal distribution, and each $X_t \in \mathbb{R}^d$, $t=1,\ldots, T$. Suppose the $\beta$-mixing coefficient of the sequence satisfies that $\beta(p) \leq c_1 e^{-c_2p}$ for some $c_1,c_2$. Then there exists a sequence, $U_i^0 = (X^0_{ip+1},X^0_{ip+2},\ldots,X^0_{ip+p})$, for $i \geq 0$, such that
	\begin{enumerate}
		\item $U_i^0$ has the same distribution as $U_i = (X_{ip+1},X_{ip+2},\ldots,X_{ip+p})$.
		\item The sequence $\{U^0_{2i}\}_{i \geq 0}$ is i.i.d., and so is  $\{U^0_{2i+1}\}_{i \geq 0}$.
		\item For any $i \geq 0$, $\mathbb{P}(U_i \neq U_i^0) \leq \beta(p)$
		\item $\mathbb{P}(|G_T(U)-G_T(U^0)| \neq 0) \leq p^{-1} T \beta(p)$, where $G_T(U) = \sqrt{T}\left\{ \frac{1}{T} \sum_{t=1}^T f(X_{t+1}|X_t) - \mathbb{E}f \right\}$, and $G_T(U^0) = \sqrt{T}\left\{ \frac{1}{T} \sum_{t=1}^T f(X^0_{t+1}|X^0_t) - \mathbb{E}f \right\}$.
	\end{enumerate}
\end{lemma}

\section{Proofs of Main Results}\label{section:main_results}

\subsection{Proof of Theorem~\ref{thm:one_step_implies_pred_suff_entire_future}}

The result follows directly from Theorem~\ref{thm:one_step_implies_pred_suff_entire_future_standard_Borel_space}, since Borel subsets of Euclidean spaces, equipped with their respective Borel $\sigma$-algebras, are standard Borel spaces.


\qed

\subsection{Proof of Theorem~\ref{thm:Phi_is_one_to_one}}

The proof is similar to that of Theorem 1 in \cite{random_features_estimation_paper} and Theorem 2 in \cite{random_features_identification_paper}. However, their motivations are very different from ours. They consider simulation-based parameter estimation for parametric models of time series, and they estimate \textit{unconditional expectations} of random features \textit{using time-averages} of the observed random features in order to injectively map the parameter space into feature space. In contrast, we estimate \textit{conditional expectations} of random features given the past observations \textit{using neural networks} in order to injectively map the predictive states of a stochastic process into feature space.

The key ingredient is the finite witness theorem (Lemma~\ref{lma:finite_witness}).  Lemma~\ref{lma:finite_witness} can be applied with any $k\geq 2p+1$, not necessarily $k=2p+1$; see the proof of Proposition 3.6 in \cite{amir_et_al_finite_witness_thm}.

We begin by showing that the assumptions of Lemma~\ref{lma:finite_witness} are satisfied.

Denote the collection of parameters for the $i$-th random Fourier feature, $\varphi_i$, by \begin{equation}\label{eqn:parameters_for_the_random_features}\vartheta_i=(\Omega_{i,1}^{\top},\ldots,\Omega_{i,m+1}^{\top},\alpha_i)^{\top},\end{equation} and let $\vartheta=(\vartheta_1,\ldots,\vartheta_k)$, which is an $\mathbb{R}^{(d(m+1)+1)\times k}$-valued random variable. Denote $$\mathbb{W}=\mathbb{R}^{d(m+1)}\times (-\pi,\pi).$$

By Assumption~\ref{asmpt:identifiability_regularity_model}, the predictive state space $\mathcal{S}$ is a compact subset of $\mathbb{R}^r$ with Hausdorff dimension $\mathrm{dim}_H(\mathcal{S})=p$. The predictive state space $\mathcal{S}$ is a semialgebraic subset of $\mathbb{R}^r$ by Assumption~\ref{asmpt:identifiability_regularity_model}, hence it is $\sigma$-subanalytic; see Figure~\ref{fig:diagram_of_subanalytic_sets_and_functions}.

By construction, the random parameters $\vartheta_i$, $i=1,\ldots,k$ from~\eqref{eqn:parameters_for_the_random_features} for the random Fourier features $\varphi_i$, $i=1,\ldots,k$ from~\eqref{eqn:random_Fourier_features} are each random variables taking values in $\mathbb{W}$, which is semialgebraic, so, as explained in Figure~\ref{fig:diagram_of_subanalytic_sets_and_functions}, it is $\sigma$-subanalytic.

By Assumption~\ref{asmpt:identifiability_regularity_model}, the real and imaginary parts of the map $(s,\omega)\mapsto \psi_{s}(\omega)$ admit jointly real analytic extensions to $O \times \mathbb{R}^{(m+1)\times d}$, and hence each $(s,\vartheta_i)\mapsto \int_{\mathcal{X}^{m+1}}\varphi_i(x)  dP_{s}(x)$, $i=1,\ldots,k$ admits a jointly real analytic extension to $O \times \mathbb{W}$ by using the definition of the characteristic function and each random Fourier feature. Note that the graph of this extension is $\sigma$-subanalytic, and that $\mathcal{S}$ is a $\sigma$-subanalytic set as discussed previously. Therefore, since $\sigma$-subanalytic sets are closed under finite intersections and Cartesian products by Lemma~\ref{lma:sigma_subanalytic_properties}, we have that the intersection of the graph of this extension with $\mathcal{S} \times \mathbb{W}\times \mathbb{R}$ is $\sigma$-subanalytic.

$\mathbb{W}$ is open and connected, and $(s,\vartheta_i)\mapsto \int_{\mathcal{X}^{m+1}}\varphi_i(x)  dP_{s}(x)$ is analytic as a function of $\vartheta_i$ for all fixed values of the predictive state $s\in\mathcal{S}$. Hence, the condition from~\eqref{cond_dimension_defficiency_for_separation} from the finite witness theorem (Lemma~\ref{lma:finite_witness}) is satisfied.

Let $x_1,\ldots,x_{m+1}$ be vectors in $\mathcal{X}$, and define $x=(x_1,\ldots,x_{m+1})^{\top}\in\mathcal{X}^{m+1}$. Letting $a=(a_1^{\top},\ldots,a_{m+1}^{\top})^{\top}\in\mathbb{R}^{d(m+1)}$, $b\in (-\pi,\pi)$, and $c=(a,b)\in\mathbb{W}$, define $$\phi_c(x)=\cos\left(\sum_{j=1}^{m+1} a_j \cdot x_{j} + b\right).$$

Define $$\mathcal{N}=\{\left(s_1,s_2\right) \in \mathcal{S}^2 \mid \int_{\mathcal{X}^{m+1}}\phi_c(x)  dP_{s_1}(x) =\int_{\mathcal{X}^{m+1}}\phi_c(x)  dP_{s_2}(x),\ \forall c\in\mathbb{W} \},$$ as the set of pairs of predictive state values that yield identical conditional expectations of the functions $\phi_c$ for all values of $c\in\mathbb{W}$. By the finite witness theorem (Lemma~\ref{lma:finite_witness}), for generic parameters $\left( \vartheta_1,\ldots,\vartheta_{k}\right) \in \mathbb{W}^k$ of the random Fourier features $\left( \varphi_1,\ldots,\varphi_{k}\right)$, we have the set equality $$\mathcal{N}=\{\left(s_1,s_2\right) \in \mathcal{S}^2 \mid \int_{\mathcal{X}^{m+1}}\varphi_i(x)  dP_{s_1}(x) =\int_{\mathcal{X}^{m+1}}\varphi_i(x)  dP_{s_2}(x),\ \forall i=1,\ldots,k \}.$$ By the discussion above Lemma~\ref{lma:finite_witness}, generic here means $\Pi$-almost every $\vartheta$, where $\Pi$ is the joint distribution from~\eqref{eqn:parameters_for_the_random_features} which is absolutely continuous with respect to the Lebesgue measure.

Denoting the $\mathbb{R}^{(m+1)\times d}$-valued random matrix by $\Omega_i=(\Omega_{i,1},\ldots,\Omega_{i,m+1})^{\top}$, we have  \begin{align*}&\int_{\mathcal{X}^{m+1}}\varphi_i(x)  dP_{s}(x)
\\& 
=\int_{\mathcal{X}^{m+1}}\cos\left( \sum_{j=1}^{m+1}\Omega_{i,j} \cdot x_j + \alpha_i\right) dP_{s}(x)
\\&
=
\cos(\alpha_i)
\int_{\mathcal{X}^{m+1}}\cos\left( \sum_{j=1}^{m+1}\Omega_{i,j} \cdot x_j\right) dP_{s}(x) 
\\&
- 
\sin(\alpha_i)
\int_{\mathcal{X}^{m+1}}\sin\left( \sum_{j=1}^{m+1}\Omega_{i,j} \cdot x_j\right) dP_{s}(x) 
\\&
= \cos(\alpha_i) \  \mathrm{Re}( \psi_{s}(\Omega_i)) - \sin(\alpha_i) \ \mathrm{Im}(  \psi_{s}(\Omega_i))
\\&
= M_{\vartheta_i}(s),
\end{align*}
where $\psi_{s}$ is the characteristic function. If the pair of predictive state values $(s_1,s_2)\in\mathcal{N}$, then $ M_{\vartheta_i}(s_1)=M_{\vartheta_i}(s_2)$ for every possible value of $\vartheta_i=(\mathrm{Vec}(\Omega_i),\alpha_i)\in\mathbb{W}$. In particular, the conditional expectation values of the random Fourier features correspond to the real and imaginary parts of the characteristic function by taking $\alpha_i=0$ and $\alpha_i=-\pi/2$, respectively, so by the uniqueness theorem (Lemma~\ref{lma:uniqueness_characteristic_function}) the corresponding conditional distributions must be equal, $P_{s_1}=P_{s_2}$, if $(s_1,s_2)\in\mathcal{N}$. 

However, by Assumption~\ref{asmpt:identifiability_regularity_model}, the map from the predictive state value to the conditional distribution, $s\mapsto P_{s}$, is one-to-one, i.e., if $s_1 \neq s_2$ then $P_{s_1}\neq P_{s_2}$, so we must have
$$(s_1,s_2)\in\mathcal{N}\implies (s_1,s_2)\in \{(s_1,s_2)\in \mathcal{S}^2 : s_1=s_2\}=\{(s,s):s\in\mathcal{S}\},$$ so $\mathcal{N}\subseteq \{(s,s):s\in\mathcal{S}\}$. Conversely, we have $\{(s,s):s\in\mathcal{S}\} \subseteq \mathcal{N}$ trivially. Therefore,
\begin{equation}\label{eqn:conclusion_mathcal_N_discrete_time}\mathcal{N}=\{(s,s):s\in\mathcal{S}\}.\end{equation}

Putting it all together, we have: for $\Pi$-almost any random parameters $(\vartheta_1,\ldots,\vartheta_k)$ for the random Fourier features $(\varphi_1,\ldots,\varphi_k)$, for every pair of predictive state values $s_1, s_2\in\mathcal{S}$, if $s_1\neq s_2$, then the conditional expectations of the random Fourier features are not equal, $\Phi(s_1)\neq \Phi(s_2)$. That is, the map $s\mapsto\Phi(s)$ is one-to-one for generic choices of $k \geq 2p+1$ random Fourier features.

\qed

\subsection{Proof of Theorem~\ref{thm:equality_of_cond_distr_St_Zt}}

First, we show that $\sigma(Z_t)\subseteq\sigma(S_t)$ by showing that $Z_t$ can be written as a measurable function of $S_t$. This holds because $Z_t=\Phi(S_t)$ by definition from~\eqref{eqn:Phi_S_t_conditional_expectation_random_features}, and $\Phi$ is measurable by the joint real analyticity of the real and imaginary parts of the characteristic function from Assumption~\ref{asmpt:identifiability_regularity_model}.

Second, we show that $\sigma(S_t)\subseteq\sigma(Z_t)$ by showing that $S_t$ can be written as a measurable function of $Z_t$. Observe that the inverse $\Phi^{-1}$ exists since $\Phi:\mathcal{S}\xrightarrow[]{}\Phi(\mathcal{S})$ is bijective by the injectivity of $\Phi$ from Assumption~\ref{asmpt:injective_Phi}. Also, observe that the inverse $\Phi^{-1}$ is measurable since it is continuous by Lemma~\ref{lma:thm_4_17_rudin}, which applies because (1) $\Phi$ is one-to-one by Assumption~\ref{asmpt:injective_Phi}, (2) $\Phi$ is continuous by the joint real analyticity of the real and imaginary parts of the characteristic function from Assumption~\ref{asmpt:identifiability_regularity_model}, and (3) $\mathcal{S}$ is compact by Assumption~\ref{asmpt:identifiability_regularity_model}. Therefore, $S_t=\Phi^{-1}(Z_t)$ and $\Phi^{-1}$ is measurable, hence $\sigma(S_t)\subseteq\sigma(Z_t)$.

Putting it all together, we have the desired result that $$
Q_{Z_t}
=
\mathcal{L}(X_{t+1}\mid Z_t)
= 
\mathcal{L}(X_{t+1}\mid S_t),
$$ almost surely, where the first equality is by definition and the second is because $\sigma(S_t)=\sigma(Z_t)$.

\qed

\section{Proofs of Additional Results}\label{section:proof_of_additional_results}

\subsection{Proof of Theorem~\ref{thm:rate_of_convergence}}
\label{subsection:proof_rate_of_convergence}

In this proof, we derive the convergence rates for the estimator $\hat{f}$ from~\eqref{eqn:mdn_estimator} based on the learned predictive state representations.

Denote the sample size by $$N=n-2m-1,$$ where $n$ is the number of observations of $X_t$, $t=1,\ldots,n$, and $m$ is from the random Fourier features from~\eqref{eqn:random_Fourier_features}.

We break the proof into three steps.

\paragraph*{Step 1.}
First, we obtain the average RNN estimation error.

For each fixed choice of RNN parameters, the initialization and recurrence from~\eqref{eqn:rnn_estimates} imply that the prediction $h_t=\lambda(H_{t+1})$ is $\mathcal{F}_t$-measurable, and that
$$Z_t \in \underset{h}{\arg\min} \ \mathbb{E}\norm{Y_{t+1}-h}_{2}^2,$$ where the argmin is over all $\mathcal{F}_t$-measurable random vectors. This follows by the definition of $\xi_{t+1}$ from~\eqref{eqn:xi_t+1} and expanding the square, \begin{align*}
\norm{Y_{t+1}-h_t}_{2}^2
&= \norm{Z_t-h_t+\xi_{t+1}}_{2}^2 \\
&= \norm{Z_t-h_t}_{2}^2
+ \norm{\xi_{t+1}}_{2}^2
+ 2\xi_{t+1}^\top(Z_t-h_t),
\end{align*}
so
\begin{align*}
\mathbb{E}\left[
\norm{Y_{t+1}-h_t}_{2}^2-\norm{\xi_{t+1}}_{2}^2
\right]
&=
\mathbb{E}\norm{Z_t-h_t}_{2}^2 
+2\mathbb{E}\left[
\xi_{t+1}^\top(Z_t-h_t)
\right],
\end{align*}
where the cross term is zero because
\begin{equation}\label{eqn:cross_term_zero}
\mathbb{E}\left[
\xi_{t+1}^\top(Z_t-h_t)
\right]
=
\mathbb{E}\left[
\mathbb{E}\left[
\xi_{t+1}^\top(Z_t-h_t)\mid\mathcal{F}_t
\right]
\right] 
=
\mathbb{E}\left[
(Z_t-h_t)^\top
\mathbb{E}[\xi_{t+1}\mid\mathcal{F}_t]
\right] =0.
\end{equation} Hence, we have \begin{align*}
\mathbb{E}\norm{Y_{t+1}-h_t}_{2}^2
&=\mathbb{E}\norm{\xi_{t+1}}_{2}^2 + \mathbb{E}\norm{h_t-Z_t}_{2}^2.
\end{align*} The first term on the right is independent of $h_t$, so $Z_t$ is the minimizer of $\mathbb{E}\norm{Y_{t+1}-h_t}_{2}^2$ over all $\mathcal{F}_t$-measurable random vectors $h_t$.

The target $Y_{t+1}=\varphi(X_{t+1:t+m+1})$ need not be $\mathcal{F}_{t+1}$-measurable when $m>0$. However, the RNN input $I_{t+1}=(X_{t-m:t},Y_{t-m})$ is
$\mathcal{F}_t$-measurable since
$Y_{t-m}=\varphi(X_{t-m:t})$ so the cross term from~\eqref{eqn:cross_term_zero} is zero for every RNN predictor $h_t=\lambda(H_{t+1})$, which establishes the conditional orthogonality condition used in the concentration arguments in the proofs of Theorems 1 and 2 in \cite{xiu_rnn_ts}.

The concentration and covering arguments used in the proofs of Theorems 1 and 2 in \citet{xiu_rnn_ts} require certain boundedness and mixing conditions. As we will show, these conditions are implied by our Assumptions~\ref{asmpt:compact_X} and~\ref{asmpt:mixing}, respectively.

Next, we check that these boundedness and mixing conditions hold, so that the Bernstein inequality (Lemma~6 in \cite{xiu_rnn_ts}) used in Lemma~8 in \cite{xiu_rnn_ts} is applicable. 

First, we check that the boundedness conditions hold. For fixed network parameters and an integer $\ell\geq0$, let $h_t^{(\ell)}$ be the prediction obtained by replacing the hidden state immediately before the most recent $\ell+1$ recurrent updates by zero, as in Definition 3 in the supplementary material of \cite{xiu_rnn_ts}. Since $h_t^{(\ell)}$ is $\mathcal{F}_t$-measurable, we have
$$
\mathbb{E}\left[
\xi_{t+1}^{\top}(Z_t-h_t^{(\ell)})
\right]
=0,
$$ by the same arguments as~\eqref{eqn:cross_term_zero}. Also, observe that $$
\mathbb{E}\left|
\xi_{t+1}^{\top}(Z_t-h_t^{(\ell)})
\right|^2
\leq
4k 
\mathbb{E}\norm{Z_t-h_t^{(\ell)}}_2^2,
$$ by Cauchy-Schwarz and since $\norm{\xi_{t+1}}_2^2\leq 4k$, and that $$\left|
\xi_{t+1}^{\top}(Z_t-h_t^{(\ell)})
\right| \leq 
\norm{\xi_{t+1}}_1
\norm{Z_t-h_t^{(\ell)}}_\infty
\leq 4k,$$ almost surely, by H\"{o}lder's inequality and since $\norm{\xi_{t+1}}_\infty\leq2$ and
$\norm{Z_t-h_t^{(\ell)}}_\infty\leq2$.

Second, we check that the mixing condition holds. Recall $V_t=(X_t^\top,Y_t^\top,Z_t^\top,\xi_t^\top)^\top$ from~\eqref{eqn:factor_process}. The random variable
$\xi_{t+1}^{\top}(Z_t-h_t^{(\ell)})$ is measurable with respect to $\sigma\left(
V_{t-\ell-m},\ldots,V_{t+1}
\right)$, so the $\beta$-mixing coefficient at lag $a>\ell+m+1$ of the process $(\xi_{t+1}^{\top}[Z_t-h_t^{(\ell)}])_{t\in\mathbb{Z}}$ is upper bounded by $\beta_V(a-\ell-m-1)$. Therefore, the argument involving the interlaced subsequences from Lemma~8 in the supplementary material of \citet{xiu_rnn_ts} applies with spacing $\ell+m+2$ and truncation $\ell\asymp w_{h,n}\log^2 n$. For the subsequences obtained by taking every
$(\ell+m+2)$-th term, the mixing coefficient at lag $a\geq 1$ is bounded by
\begin{align*}&
\beta_V(a(\ell+m+2)-\ell-m-1)
=
\beta_V((a-1)(\ell+m+2)+1)
\leq
c_{\mathrm{mix}}^1\exp(-c_{\mathrm{mix}}^2 a).
\end{align*} Note that we assume $\beta$-mixing, whereas \cite{xiu_rnn_ts} only assume $\alpha$-mixing. Since $\alpha$-mixing coefficients are upper bounded by $\beta$-mixing
coefficients (see, e.g., \cite{bradley_mixing_2005}), the concentration inequality used in Lemma~8 of the supplementary material of \citet{xiu_rnn_ts} applies.

Choose $\ell=\lceil C^{\ast} w_{h,n}\log^2 n \rceil$ for a sufficiently large
constant $C^{\ast}>0$. Lemma~7 from the supplementary material of
\citet{xiu_rnn_ts} controls the difference between the truncated and full RNN predictions. The covering and union bounding arguments from Lemma~10 in \cite{xiu_rnn_ts}, yields a concentration bound that holds uniformly over the RNN class, so in particular it applies to the fitted RNN.

Therefore, analogous arguments from the proof of Theorem 1 in \cite{xiu_rnn_ts} apply with the RNN input $I_{t+1}=(X_{t-m:t},Y_{t-m})$ and the filtration $\mathcal{F}_t$. Similarly, analogous RNN approximation arguments from the proof of Theorem 2 in \citet{xiu_rnn_ts} apply to the representation $$
Z_t=g^{\ast}(Y_{t-m},\xi_{t-m},X_{t-m:t}),
$$ from Lemma~\ref{lma:conditional_expectation_as_function_of_noise_process}, where the input dimension of $g^{\ast}(\cdot)$ is $2k+d(m+1)$. Since $m$ from~\eqref{eqn:random_Fourier_features} is fixed, using these lags only changes constants in the proof of Theorem 2 in \citet{xiu_rnn_ts} due to changes in the width of the recurrent layer.

By equation 18 in the proof of Theorem 1 in \citet{xiu_rnn_ts}, Theorem 2 of \citet{xiu_rnn_ts} with the widths and depths from the statement of Theorem~\ref{thm:rate_of_convergence}, and Markov's inequality, we have 
\begin{equation}\label{eqn:rnn_average_rate}
\frac{1}{N}\sum_{t=m+1}^{n-m-1}\norm{\hat{Z}_t-Z_t}_{2}^2
=O_{\mathbb{P}}\left(
 n^{-\min\left(
 \frac{\eta}{\eta+4k+2d(m+1)},
 \frac{2k+d(m+1)+3}{10k+5d(m+1)+3}
 \right)}(\log n)^{11}\right).
\end{equation}

\paragraph*{Step 2.}
Second, we obtain the MDN estimation error. 

Recall $P_j,P_{n,j},K_j$, and $\norm{\cdot}_{L^2,j}$ from the proof of
Theorem~\ref{thm:mdn_rates}. Since the estimator from~\eqref{eqn:mdn_estimator} is an argmax, by
Assumption~\ref{asmpt:mdn_model} and Jensen's inequality, we have
\begin{equation}\label{eqn:generated_covariate_likelihood_perturbation}
\begin{aligned}
P_{n,j}[\log g_{n,j}-\log\hat{f}_j]
&\leq\frac{2\Lambda_n}{N}\sum_{t=m+1}^{n-m-1}\norm{\hat{Z}_t-Z_t}_{2}^\kappa \\&
\leq2\Lambda_n\left(
\frac{1}{N}\sum_{t=m+1}^{n-m-1}\norm{\hat{Z}_t-Z_t}_{2}^2
\right)^{\kappa/2},
\end{aligned}
\end{equation}
where $P_{n,j}$ is from~\eqref{eqn:Pj_Pnj} and $g_{n,j}$ is the approximation from Assumption~\ref{asmpt:mdn_generated_covariate_stability}. Note that the factor of $2$ accounts
for replacing the covariates in both $g_{n,j}$ and $\hat{f}_j$. By~\eqref{eqn:Kj_upper_bound},
\eqref{eqn:stable_mdn_approximation}, and
\eqref{eqn:stable_mdn_tail}, we have
$$K_j(g_{n,j})\leq Ca_{n,j}^2.$$

Note that the mean value theorem applied to $\log$ and the density bounds give a log-likelihood ratio bounded by $Ca_{n,j}$.
Hence, the argument from~\eqref{eqn:dependent_MDN_approximation_bound},
with $\tilde{r}=3\log n$ and the coupling identities from~\eqref{eqn:mdn_coupled_approximation_identity} and~\eqref{eqn:mdn_coupled_estimation_identity}, imply that, with probability at least $1-O(n^{-1})$,
\begin{equation}\label{eqn:feasible_mdn_comparator_bound}
\begin{aligned}
P_{n,j}(
\log f_{X_{t+1,j}\mid\tilde{Z}_{t,j}}-\log g_{n,j}
)\leq C(
 M^{-\omega_1}
 +M^{\frac{\gamma+k+j-1}{2\gamma}+4\omega_2}
 n^{-\frac{\gamma}{2\gamma+k+j-1}}\log^3(nM)
)^2.
\end{aligned}
\end{equation}

Observe that
\begin{equation}\label{eqn:l2_upper_bound_3_terms}\begin{aligned}
\frac{c_f^-}{2c_+^2}
\norm{
\hat{f}_j-f_{X_{t+1,j}\mid\tilde{Z}_{t,j}}
}_{L^2,j}^2
&\leq
(P_j-P_{n,j})
\left[
\log f_{X_{t+1,j}\mid\tilde{Z}_{t,j}}
-\log\hat{f}_j
\right]\\
&\quad+
P_{n,j}
\left[
\log f_{X_{t+1,j}\mid\tilde{Z}_{t,j}}
-\log g_{n,j}
\right]\\
&\quad+
2\Lambda_n
\left(
\frac{1}{N}
\sum_{t=m+1}^{n-m-1}
\norm{\hat{Z}_t-Z_t}_2^2
\right)^{\kappa/2},
\end{aligned}\end{equation} by using the lower bound from~\eqref{eqn:Kj_lower_bound}, decomposing $
K_j(\hat{f}_j)$ from~\eqref{eqn:Kj} into
\begin{align*}
K_j(\hat{f}_j)
&=(P_j-P_{n,j})
 [\log f_{X_{t+1,j}\mid\tilde{Z}_{t,j}}-\log\hat{f}_j]\\
&\quad+P_{n,j}[\log f_{X_{t+1,j}\mid\tilde{Z}_{t,j}}-\log g_{n,j}] \\&\quad+P_{n,j}[\log g_{n,j}-\log\hat{f}_j],
\end{align*} and applying the inequality from~\eqref{eqn:generated_covariate_likelihood_perturbation} to the last term.

Applying the inequality from~\eqref{eqn:feasible_mdn_comparator_bound} to the second term in~\eqref{eqn:l2_upper_bound_3_terms}, and applying~\eqref{eqn:dependent_MDN_estimation_bound} and~\eqref{eqn:mdn_coefficient_pseudoVCdimension} to the first term in~\eqref{eqn:l2_upper_bound_3_terms} and then applying the same recursive improvement argument used in Step 2.4 of the proof of Theorem~\ref{thm:mdn_rates}, yields, with probability $1-O(n^{-1})$,
\begin{equation}\label{eqn:feasible_mdn_factor_squared_bound}
\begin{aligned}
\frac{c_f^-}{4c_+^2}
\norm{\hat{f}_j-f_{X_{t+1,j}\mid\tilde{Z}_{t,j}}}_{L^2,j}^2
&\leq C\left(
 M^{-\omega_1}
 +M^{\frac{\gamma+k+j-1}{2\gamma}+4\omega_2}
 n^{-\frac{\gamma}{2\gamma+k+j-1}}\log^3(nM)
\right)^2\\
&\quad+2\Lambda_n\left(
\frac{1}{N}\sum_{t=m+1}^{n-m-1}\norm{\hat{Z}_t-Z_t}_{2}^2
\right)^{\kappa/2}.
\end{aligned}
\end{equation}

\paragraph*{Step 3.} Third, we obtain the convergence rate.

The factorization from~\eqref{eqn:factorization_conditional_density_Xt+1_given_Zt} and the
argument from~\eqref{eqn:joint_density_coordinatewise_bound} give
$$
\norm{\hat{f}-f_{X_{t+1}\mid Z_t}}
 _{L^2}
\leq C\sum_{j=1}^d
\norm{\hat{f}_j-f_{X_{t+1,j}\mid\tilde{Z}_{t,j}}}_{L^2,j}.
$$ 
Taking square roots in
\eqref{eqn:feasible_mdn_factor_squared_bound}, using the inequality $\sqrt{a+b}\leq\sqrt a+\sqrt b$, substituting the RNN estimation error from~\eqref{eqn:rnn_average_rate}, and applying Markov's inequality gives
\begin{align*}
\norm{\hat{f}-f_{X_{t+1}\mid Z_t}}
 _{L^2}
&=O_{\mathbb{P}}\Bigg(\begin{aligned}&
 M^{-\omega_1}
 +M^{\frac{\gamma+k+d-1}{2\gamma}+4\omega_2}
 n^{-\frac{\gamma}{2\gamma+k+d-1}}\log^3(nM)\\
&+
 \Lambda_n^{1/2}n^{-\frac{\kappa}{4}\min\left(
 \frac{\eta}{\eta+4k+2d(m+1)},
 \frac{2k+d(m+1)+3}{10k+5d(m+1)+3}
 \right)}(\log n)^{11\kappa/4}\end{aligned}
\Bigg).\end{align*} 
Note that the MDN with $j=d$ has the slowest convergence rate due to the greater number of covariates, so the upper bound is largest at $j=d$.
\qed

\subsection{Proof of Theorem~\ref{thm:prediction_at_estimated_state}}
\label{subsection:proof_prediction_at_estimated_state}

We break the proof into three steps.

\paragraph*{Step 1.} 

Choose $g_{n,j}$ from
Assumption~\ref{asmpt:mdn_generated_covariate_stability}, and let
\begin{equation}\label{eqn:g_n_factorization}
g_n(x\mid z)=\prod_{j=1}^d g_{n,j}(x_j\mid\tilde{z}_j).\end{equation}
For our estimator $\hat{f}$ from~\eqref{eqn:mdn_estimator} and $g_n$ from~\eqref{eqn:g_n_factorization}, define \begin{equation}\label{eqn:q_n} q_n(z)=
\norm{\hat{f}(\cdot\mid z)-g_n(\cdot\mid z)}_{L^2(\mathcal{X})}.\end{equation}
By~\eqref{eqn:stable_mdn_approximation}, the factorization from~\eqref{eqn:factorization_conditional_density_Xt+1_given_Zt},
the density bounds in
Assumptions~\ref{asmpt:density_regularity_conditions}
and~\ref{asmpt:mdn_model}, and the compactness from
Assumption~\ref{asmpt:compact_X}, we have
\begin{equation}\label{eqn:prediction_mdn_approximation}
\sup_{z\in\mathcal{Z}}
\norm{g_n(\cdot\mid z)
-f_{X_{t+1}\mid Z_t}(\cdot\mid z)}_{L^2(\mathcal{X})}
\leq C\sum_{j=1}^d a_{n,j} \leq
C\left(M^{-\omega_1}
+M^{4\omega_2}n^{-\frac{\gamma}{2\gamma+k+d-1}}\right),
\end{equation}
 where the last inequality uses the facts that $M_j=M$, $j\leq d$, and $d$ is fixed.

By the triangle inequality,
\eqref{eqn:prediction_mdn_approximation}, and
\eqref{eqn:rate_of_our_estimator_L2}, we have
\begin{equation}\label{eqn:prediction_q_integrated_rate}
\begin{aligned}
\norm{q_n}_{L^2(P_{Z_t})}
&\leq
\norm{\hat{f}-f_{X_{t+1}\mid Z_t}}_{L^2}
+\norm{g_n-f_{X_{t+1}\mid Z_t}}_{L^2}\\
&=O_{\mathbb{P}}\Bigg(\begin{aligned}
&M^{-\omega_1}
+M^{\frac{\gamma+k+d-1}{2\gamma}+4\omega_2}
n^{-\frac{\gamma}{2\gamma+k+d-1}}\log^3(nM)\\
&+\Lambda_n^{1/2}
n^{-\frac{\kappa}{4}\min\left(
\frac{\eta}{\eta+4k+2d(m+1)},
\frac{2k+d(m+1)+3}{10k+5d(m+1)+3}
\right)}(\log n)^{11\kappa/4}
\end{aligned}\Bigg).
\end{aligned}
\end{equation}

For either $f=\hat{f}$ or $f=g_n$, the H\"older condition in Assumption~\ref{asmpt:mdn_model} implies that
\begin{equation}\label{eqn:prediction_log_holder}
\sup_{x\in\mathcal{X}}
|\log f(x\mid z)-\log f(x\mid z')|
\leq \Lambda_n d \norm{z-z'}_2^\kappa,
\quad z,z'\in[-1,1]^k.
\end{equation} Moreover, by the mean value theorem applied to $\exp(\cdot)$, i.e., $|e^a-e^b|\leq e^{\max(a,b)}|a-b|$, for all $x\in\mathcal{X}$ and $z,z'\in\mathcal{Z}$, we have
$$\begin{aligned}
|f(x\mid z)-f(x\mid z')| & = 
|\exp(\log f(x\mid z))- \exp( \log f(x\mid z'))|
\\&
\leq  \exp(\max(\log f(x\mid z),\log f(x\mid z')))|\log f(x\mid z)-  \log f(x\mid z')|
\\&
\leq  \exp(d \log c_{\mathrm{MDN}})|\log f(x\mid z)-  \log f(x\mid z')|
\\
&\leq c_{\mathrm{MDN}}^d
|\log f(x\mid z)-\log f(x\mid z')|\\
&\leq  c_{\mathrm{MDN}}^d\Lambda_n d \norm{z-z'}_2^\kappa,
\end{aligned}$$ because each of the $d$ MDN factors is upper bounded by $c_{\mathrm{MDN}}$ from Assumption~\ref{asmpt:mdn_model}. Define
\begin{equation}\label{eqn:C_0}
C_0=
\max\left(
1,
2d c_{\mathrm{MDN}}^d
\left(\int_{\mathcal{X}}1 dx\right)^{1/2}
\right),\end{equation} which is finite due to the compactness of $\mathcal{X}$ from
Assumption~\ref{asmpt:compact_X}. For $q_n(\cdot)$ from~\eqref{eqn:q_n}, for $z\in\mathcal{Z}$, the triangle inequality and Assumption~\ref{asmpt:mdn_model} yield
\begin{equation}\label{eqn:prediction_q_uniform_bound}
 q_n(z) \leq
\left(
\int_{\mathcal{X}}
\left(
|\hat{f}(x\mid z)|+|g_n(x\mid z)|
\right)^2dx
\right)^{1/2} 
\leq2c_{\mathrm{MDN}}^d
\left(\int_{\mathcal{X}}1 dx\right)^{1/2}
\leq C_0,
\end{equation} where the last inequality holds by~\eqref{eqn:C_0} and because $d \geq 1$. By the reverse triangle inequality, the triangle inequality, and the inequalities above, for $z,z'\in\mathcal{Z}$, we have \begin{equation}\label{eqn:prediction_q_holder}
\begin{aligned}
|q_n(z)-q_n(z')|
&\leq
\norm{\hat{f}(\cdot\mid z)-\hat{f}(\cdot\mid z')}_{L^2(\mathcal{X})}
+\norm{g_n(\cdot\mid z)-g_n(\cdot\mid z')}_{L^2(\mathcal{X})}\\
&\leq2d c_{\mathrm{MDN}}^d
\left(\int_{\mathcal{X}}1 dx\right)^{1/2}
\Lambda_n\norm{z-z'}_2^\kappa\\
&\leq C_0 \Lambda_n\norm{z-z'}_2^\kappa.
\end{aligned}
\end{equation}

\paragraph*{Step 2.} Second, we evaluate $q_n(\cdot)$ from~\eqref{eqn:q_n}
at the true predictive state representation.

Let $\delta>0$ and $0<v<C_0$. Divide each of the $k$ coordinate intervals
$[-1,1]$ from $[-1,1]^k$ into \begin{equation}\label{eqn:number_of_intervals}\left\lceil
2\sqrt{k}\left(\frac{2C_0\Lambda_n}{v}\right)^{1/\kappa}
\right\rceil,\end{equation} consecutive intervals of equal length. To make the intervals disjoint, we take each interval to include its
left endpoint but not its right endpoint, except for the final
interval, which also includes $1$. 

Denote the Cartesian products of these intervals across the $k$ coordinates by $B_1,\ldots,B_{J'}$ for some $J'\in\mathbb{N}$, each of which is a subset of $[-1,1]^k$. Define
$$
A_i=\mathcal Z\cap B_i,
$$
and denote the non-empty sets, i.e., such that $\mathcal Z\cap B_i\neq\varnothing$, as $A_1,\ldots,A_J$ for some $J\in\mathbb{N}$. Note that, by construction, the sets
$A_1,\ldots,A_J$ are disjoint, measurable subsets of $\mathcal Z$ that satisfy
$$
\mathcal Z=\bigcup_{i=1}^J A_i.
$$
Moreover, since each coordinate interval has length at most
$k^{-1/2}(v/(2C_0\Lambda_n))^{1/\kappa}$, we have
$$
\mathrm{diam}(A_i)=\sup_{a_1,a_2\in A_i}\norm{a_1 - a_2}_2
\leq
\left(\frac{v}{2C_0\Lambda_n}\right)^{1/\kappa}.
$$

By~\eqref{eqn:number_of_intervals}, the number of intervals in each of the $k$ coordinates is at most a constant
multiple of
$$
\left(\frac{\Lambda_n}{v}\right)^{1/\kappa},
$$
so, since $k$ is fixed, we have
\begin{equation}\label{eqn:J_ineq}
J\leq C\left(\frac{\Lambda_n}{v}\right)^{k/\kappa},
\end{equation} for some constant $C> 0$. Therefore, by~\eqref{eqn:prediction_q_holder} and the construction above, for any
$z,z'\in A_i$,
\begin{equation}\label{eqn:qz_ineq_diam}
|q_n(z)-q_n(z')|
\leq
C_0\Lambda_n
\mathrm{diam}(A_i)^\kappa
\leq \frac{v}{2}.\end{equation}

Suppose $q_n(Z_{n+L-1})>v$ and $\norm{q_n}_{L^2(P_{Z_t})}\leq\delta$. In this case, if $Z_{n+L-1}\in A_i$, then, 
$$q_n(z)\geq q_n(Z_{n+L-1})-v/2>v/2,$$ for every $z\in A_i$. Hence, we have \begin{equation}\label{eqn:v4_Pz_ineq}\frac{v^2}{4}P_{Z_t}(A_i)
\leq\int_{A_i}q_n(z)^2 dP_{Z_t}(z)
\leq\delta^2,\end{equation} where the second inequality is because $\norm{q_n}_{L^2(P_{Z_t})}\leq\delta$, and the first inequality follows from~\eqref{eqn:qz_ineq_diam} because when $q_n(Z_{n+L-1})>v$, if
$Z_{n+L-1}\in A_i$, then,
$$
q_n(z)
\geq
q_n(Z_{n+L-1})
-
|q_n(z)-q_n(Z_{n+L-1})|
>
v-\frac{v}{2}
=
\frac{v}{2},
$$ for every $z\in A_i$. Multiplying~\eqref{eqn:v4_Pz_ineq} by $4/v^2$ gives us
\begin{equation}\label{eqn:PZt_ineq} P_{Z_t}(A_i) \leq \frac{4}{v^2} \delta^2.\end{equation}

By Assumption~\ref{asmpt:mixing}, the process is stationary, so $\mathbb{P}(Z_{n+L-1}\in A_i)=P_{Z_t}(A_i)$.
Therefore, we have
\begin{equation}\label{eqn:prediction_partition_probability}
\begin{aligned}
\mathbb{P}(q_n(Z_{n+L-1})>v)&\leq \mathbb{P}\left(q_n(Z_{n+L-1})>v, \norm{q_n}_{L^2(P_{Z_t})}> \delta\right) \\& \quad + 
\mathbb{P}\left(q_n(Z_{n+L-1})>v,\norm{q_n}_{L^2(P_{Z_t})}\leq \delta\right) \\&\leq\mathbb{P}\left(\norm{q_n}_{L^2(P_{Z_t})}>\delta\right)
+\sum_{i:P_{Z_t}(A_i)\leq4\delta^2/v^2}
P_{Z_t}(A_i)\\
&\leq\mathbb{P}\left(\norm{q_n}_{L^2(P_{Z_t})}>\delta\right)
+C\delta^2\Lambda_n^{k/\kappa}v^{-2-k/\kappa},
\end{aligned}
\end{equation}
where the first inequality is by set inclusion, the last inequality is by~\eqref{eqn:J_ineq} and~\eqref{eqn:PZt_ineq}, and the second inequality is by set inclusion and the following argument: if the event in the second term, $\{q_n(Z_{n+L-1})>v,\norm{q_n}_{L^2(P_{Z_t})}\leq \delta\}$, occurs, then for some $i$ such that $P_{Z_t}(A_i) \leq \frac{4}{v^2} \delta^2$ we have $Z_{n+L-1}\in A_i$ by~\eqref{eqn:PZt_ineq}, so a union bound gives the inequality.

Recall $C_0$ from~\eqref{eqn:C_0}. For some $K_0>0$, substituting
$v=K_0\Lambda_n^{k/(2\kappa+k)}\delta^{2\kappa/(2\kappa+k)}$
into~\eqref{eqn:prediction_partition_probability} yields
\begin{equation}\label{eqn:prob_ineq_qnZ}\mathbb{P}\left(
q_n(Z_{n+L-1})>
K_0\Lambda_n^{\frac{k}{2\kappa+k}}
\delta^{\frac{2\kappa}{2\kappa+k}}
\right)
\leq
\mathbb{P}\left(\norm{q_n}_{L^2(P_{Z_t})}>\delta\right)
+CK_0^{-2-k/\kappa}.\end{equation} Note that, for $v\geq C_0$, $\mathbb{P}(q_n(Z_{n+L-1})>v)$ from~\eqref{eqn:prediction_partition_probability} is zero by~\eqref{eqn:prediction_q_uniform_bound}. 

For any given $\epsilon>0$, choose $\delta$ to be a sufficiently large constant multiple
of the expression inside the $O_{\mathbb{P}}(\cdot)$ from~\eqref{eqn:prediction_q_integrated_rate}, which makes the first term on the right side of~\eqref{eqn:prob_ineq_qnZ} at most $\epsilon/2$ for large $n$. Choosing the constant $K_0$ so that $CK_0^{-2-k/\kappa}$ from~\eqref{eqn:prob_ineq_qnZ} is less than or equal to $\epsilon/2$ implies that \begin{equation}\label{eqn:prediction_q_pointwise_rate}
\begin{aligned} &q_n(Z_{n+L-1})\\& =O_{\mathbb{P}}\Bigg(
\Lambda_n^{\frac{k}{2\kappa+k}}
\left[\begin{aligned}
&M^{-\omega_1}
+M^{\frac{\gamma+k+d-1}{2\gamma}+4\omega_2}
n^{-\frac{\gamma}{2\gamma+k+d-1}}\log^3(nM)\\
&+\Lambda_n^{1/2}
n^{-\frac{\kappa}{4}\min\left(
\frac{\eta}{\eta+4k+2d(m+1)},
\frac{2k+d(m+1)+3}{10k+5d(m+1)+3}
\right)}(\log n)^{11\kappa/4}
\end{aligned}\right]^{\frac{2\kappa}{2\kappa+k}}
\Bigg).
\end{aligned}
\end{equation}

\paragraph*{Step 3.} Third, we substitute the estimated predictive state.

By Assumption~\ref{asmpt:rnn_function_class}
and~\eqref{eqn:dnn_bounds}, we have
$$\hat{Z}_{n+L-1}\in[-1,1]^k.$$

Also, by~\eqref{eqn:rnn_forecast_mse} and Markov's inequality, we have
\begin{equation}\label{eqn:prediction_holder_argument}
\begin{aligned}
& \Lambda_n d \norm{\hat{Z}_{n+L-1}-Z_{n+L-1}}_2^\kappa\\
& =O_{\mathbb{P}}\left(
\Lambda_n n^{-\frac{\kappa}{2}\min\left(
\frac{\eta}{\eta+4k+2d(m+1)},
\frac{2k+d(m+1)+3}{10k+5d(m+1)+3}
\right)}(\log n)^{11\kappa/2}
\right)
\\&=o_{\mathbb P}(1),
\end{aligned}
\end{equation} which tends to zero by the assumption in the theorem statement because this rate from~\eqref{eqn:prediction_holder_argument} is the square of the last summand of the expression inside $O_{\mathbb{P}}(\cdot)$ from~\eqref{eqn:rate_of_our_estimator_L2}.

For $x\in\mathcal{X}$, we have
$$\begin{aligned}
&|\hat{f}(x\mid\hat{Z}_{n+L-1})-\hat{f}(x\mid Z_{n+L-1})|\\
&=
\hat f(x\mid Z_{n+L-1})
\left|
\exp\left(
\log\hat f(x\mid\hat Z_{n+L-1})
-\log\hat f(x\mid Z_{n+L-1})
\right)-1
\right|
\\&\leq\hat{f}(x\mid Z_{n+L-1})
\left[\exp\left(
\Lambda_n d \norm{\hat{Z}_{n+L-1}-Z_{n+L-1}}_2^\kappa
\right)-1\right]\\
& \leq c_{\mathrm{MDN}}^d
\left[\exp\left(
\Lambda_n d \norm{\hat{Z}_{n+L-1}-Z_{n+L-1}}_2^\kappa
\right)-1\right],
\end{aligned}$$ by properties of $\log(\cdot)$ and $\exp(\cdot)$, factoring out $\hat f(x\mid Z_{n+L-1})$,~\eqref{eqn:prediction_log_holder}, and Assumption~\ref{asmpt:mdn_model}.

By~\eqref{eqn:prediction_holder_argument}, $$\Lambda_n d
\norm{\hat Z_{n+L-1}-Z_{n+L-1}}_2^\kappa
=o_{\mathbb P}(1),$$ which implies that this term can be upper bounded by one with probability tending to one. By the mean value theorem, with probability tending to one, we have
\begin{align*}
&\norm{\hat{f}(\cdot\mid\hat{Z}_{n+L-1})
-\hat{f}(\cdot\mid Z_{n+L-1})}_{L^2(\mathcal{X})}\\
&\leq
3 c_{\mathrm{MDN}}^d
\left(\int_{\mathcal X}1 dx\right)^{1/2}
\Lambda_n d 
\norm{\hat Z_{n+L-1}-Z_{n+L-1}}_2^\kappa.
\end{align*}
Since $\mathcal{X}$ is compact by Assumption~\ref{asmpt:compact_X}, applying~\eqref{eqn:prediction_holder_argument} yields
\begin{equation}\label{eqn:prediction_substitution_rate}
\begin{aligned}
&\norm{\hat{f}(\cdot\mid\hat{Z}_{n+L-1})
-\hat{f}(\cdot\mid Z_{n+L-1})}_{L^2(\mathcal{X})}\\
&=
O_{\mathbb{P}}\left(
\Lambda_n n^{-\frac{\kappa}{2}\min\left(
\frac{\eta}{\eta+4k+2d(m+1)},
\frac{2k+d(m+1)+3}{10k+5d(m+1)+3}
\right)}
(\log n)^{11\kappa/2}
\right).
\end{aligned}
\end{equation}

Since the process is stationary by Assumption~\ref{asmpt:mixing}, $(X_{n+L},Z_{n+L-1})$ has the same distribution as
$(X_{t+1},Z_t)$, so we can use the same (version of) the conditional density. Observe that \begin{align*}
&\norm{\hat{f}(\cdot\mid\hat{Z}_{n+L-1})
-f_{X_{n+L}\mid Z_{n+L-1}}
(\cdot\mid Z_{n+L-1})}_{L^2(\mathcal X)}\\
&\leq
\norm{\hat{f}(\cdot\mid\hat{Z}_{n+L-1})
-\hat{f}(\cdot\mid Z_{n+L-1})}_{L^2(\mathcal X)}\\
& +
\norm{\hat{f}(\cdot\mid Z_{n+L-1})
-g_n(\cdot\mid Z_{n+L-1})}_{L^2(\mathcal X)}\\
&+
\norm{g_n(\cdot\mid Z_{n+L-1})
-f_{X_{t+1}\mid Z_t}
(\cdot\mid Z_{n+L-1})}_{L^2(\mathcal X)},
\end{align*} by adding and subtracting both $\hat f(\cdot\mid Z_{n+L-1})$ and
$g_n(\cdot\mid Z_{n+L-1})$, and by applying the triangle inequality. By the definition of $q_n(\cdot)$ from~\eqref{eqn:q_n}, $$q_n(Z_{n+L-1}) = \norm{\hat{f}(\cdot\mid Z_{n+L-1})
-g_n(\cdot\mid Z_{n+L-1})}_{L^2(\mathcal X)}.$$

Also, since $Z_{n+L-1}\in\mathcal Z$ almost surely,
\eqref{eqn:prediction_mdn_approximation} yields $$
\begin{aligned}&\norm{g_n(\cdot\mid Z_{n+L-1})
-f_{X_{t+1}\mid Z_t}
(\cdot\mid Z_{n+L-1})}_{L^2(\mathcal X)}\\
&\leq
\sup_{z\in\mathcal Z}
\norm{g_n(\cdot\mid z)
-f_{X_{t+1}\mid Z_t}(\cdot\mid z)}_{L^2(\mathcal X)}\\
&\leq
C\left(
M^{-\omega_1}
+
M^{4\omega_2}
n^{-\frac{\gamma}{2\gamma+k+d-1}}
\right).
\end{aligned}
$$

Combining the above inequalities, we have \begin{equation}\label{eqn:prediction_final_triangle}
\begin{aligned}
&\norm{\hat{f}(\cdot\mid\hat{Z}_{n+L-1})
-f_{X_{n+L}\mid Z_{n+L-1}}
(\cdot\mid Z_{n+L-1})}_{L^2(\mathcal X)}\\
&\leq
\norm{\hat{f}(\cdot\mid\hat{Z}_{n+L-1})
-\hat{f}(\cdot\mid Z_{n+L-1})}_{L^2(\mathcal X)}\\&
+q_n(Z_{n+L-1})\\
&+
C\left(
M^{-\omega_1}
+
M^{4\omega_2}
n^{-\frac{\gamma}{2\gamma+k+d-1}}
\right).
\end{aligned}
\end{equation} The first term has the rate from
\eqref{eqn:prediction_substitution_rate}, which is bounded by the
square of the expression in the square brackets from~\eqref{eqn:prediction_q_pointwise_rate}. The second term has the
rate from~\eqref{eqn:prediction_q_pointwise_rate}. The final term is
bounded by a constant times the same expression in the square brackets from~\eqref{eqn:prediction_q_pointwise_rate}. 

Hence, each term in~\eqref{eqn:prediction_final_triangle} is bounded by the
rate in~\eqref{eqn:prediction_at_estimated_state_rate} because the expression in the square brackets from~\eqref{eqn:prediction_q_pointwise_rate} tends to zero, because $\Lambda_n\geq1$ $2\kappa/(2\kappa+k)\in(0,1)$, and because $u^2\leq u\leq
\Lambda_n^{\frac{k}{2\kappa+k}}
u^{\frac{2\kappa}{2\kappa+k}}$ for $0<u\leq1$.  This establishes the desired rate from~\eqref{eqn:prediction_at_estimated_state_rate}.

\qed

\subsection{Proof of Theorem~\ref{thm:mdn_rates}}\label{subsection:proof_of_mdn_rates}

The proof is broken into three main steps, as in the proof of Theorem 3 in \cite{Zhou_et_al_2023}.

To begin, we introduce some notation. Recall that $$
\tilde{Z}_{t,j}=(Z_t^{\top},X_{t+1,1},\ldots,X_{t+1,j-1})^{\top}, \quad j \in [d].$$ Denote the sample size by $$N=n-2m-1,$$ where $n$ is the number of observations of $X_t$, $t=1,\ldots,n$, and $m$ is from the random Fourier features from~\eqref{eqn:random_Fourier_features}.

For every measurable function $h=h(X_{t+1,j}, \tilde{Z}_{t,j})$, define \begin{equation}\label{eqn:Pj_Pnj}P_j h=\mathbb{E}[h(X_{t+1,j},\tilde{Z}_{t,j})], \quad P_{n,j} h = \frac{1}{N} \sum_{t=m+1}^{n-m-1} h(X_{t+1,j}, \tilde{Z}_{t,j}),\end{equation} and let $$\norm{h}_{L^2,j}^2
=
\int_{\tilde{\mathcal{Z}}_j}
\int_{\mathcal{X}_j}
\abs{h(x_j, \tilde{z}_j)}^2
dx_j dP_{\tilde{Z}_{t,j}}(\tilde{z}_j),\quad \norm{h}_{n,j}^2
= \frac{1}{N} \sum_{t=m+1}^{n-m-1} \abs{h(X_{t+1,j}, \tilde{Z}_{t,j})}^2.$$  Similarly, denote $$f_j=f_j(X_{t+1,j}\mid \tilde{Z}_{t,j}),$$ for some conditional density $f_j(\cdot \mid \cdot)$, so that we may concisely write $$P_j f_j = \mathbb{E}[ f_j(X_{t+1,j}\mid \tilde{Z}_{t,j})],$$ and similarly $P_j \log f_j = \mathbb{E}[ \log f_j(X_{t+1,j}\mid \tilde{Z}_{t,j})]$.

\paragraph*{Step 1.} This step corresponds to Step 1 in the proof of Theorem 3 in \cite{Zhou_et_al_2023}, which itself uses ideas from \cite{dnn_est_inf}. In this step, the observations are temporarily
regarded as independent, and the extension to dependent observations is in Step 2.

\paragraph*{Step 1.1.} This step corresponds to Step 1.1 in the proof of Theorem 3 in \cite{Zhou_et_al_2023}, which starts in the iid setting.

We derive upper and lower bounds for the quantity $ K_j(\hat{f}_{j}^{\mathrm{oracle}})$, where for any measurable function $h$ that is $P_j$-almost surely positive and for which the following expectation is finite, we define \begin{align}\label{eqn:Kj}
K_j(h)&=P_j \left( \log f_{X_{t+1,j} \mid \tilde{Z}_{t,j}} - \log h\right) \\&= \int_{\tilde{\mathcal{Z}}_j} 
\int_{\mathcal{X}_j} f_{X_{t+1,j}\mid \tilde{Z}_{t,j}}(x_j\mid\tilde{z}_j) \left[ \log f_{X_{t+1,j} \mid \tilde{Z}_{t,j}}(x_j \mid \tilde{z}_j) - \log h(x_j \mid \tilde{z}_j)\right] dx_jdP_{\tilde{Z}_{t,j}}(\tilde{z}_j)\nonumber.\end{align} 
 
For $a \in [0,1]$, define $$f_{j,a}(x_j \mid \tilde{z}_j) = (1-a) f_{X_{t+1,j} \mid \tilde{Z}_{t,j}}(x_j \mid \tilde{z}_j) + a f_j(x_j \mid \tilde{z}_j).$$ Note that $f_{j,1}(x_j \mid \tilde{z}_j)=f_j(x_j\mid \tilde{z}_j)$ and $f_{j,0}(x_j \mid \tilde{z}_j)=f_{X_{t+1,j} \mid \tilde{Z}_{t,j}}(x_j \mid \tilde{z}_j)$, and that $f_{j,a}$ need not belong to the function class $\mathcal{M}_{n,j}^{\mathrm{MDN}}$.

By Taylor's theorem (with the integral form of the remainder) applied to $a \mapsto K_j(f_{j,a})$ about $a=0$ then evaluating at $a=1$, we have
\begin{equation}\label{eqn:taylor_theorem_integral_remainder}
K_j(f_j)
=K_j(f_{j,1})
=
K_j(f_{j,0})
+
\left.
\frac{d}{da}K_j(f_{j,a})
\right|_{a=0}
+
\int_0^1
(1-a)
\frac{d^2}{da^2}K_j(f_{j,a})
da,\end{equation} where $K_j(f_{j,0})=0$ by~\eqref{eqn:Kj}. Explicitly, we have
\begin{align*}
& K_j(f_j) \\
&=
-\int_{\tilde{\mathcal{Z}}_j}
\int_{\mathcal{X}_j}
\frac{
f_{X_{t+1,j}\mid \tilde{Z}_{t,j}}(x_j\mid \tilde{z}_j)
}{
f_{j,0}(x_j\mid \tilde{z}_j)
}
\left[
f_j(x_j\mid \tilde{z}_j)
-
f_{X_{t+1,j}\mid \tilde{Z}_{t,j}}(x_j\mid \tilde{z}_j)
\right]
dx_jdP_{\tilde{Z}_{t,j}}(\tilde{z}_j)
\\
& +
\int_0^1 (1-a)
\int_{\tilde{\mathcal{Z}}_j}
\int_{\mathcal{X}_j}
\frac{
f_{X_{t+1,j}\mid \tilde{Z}_{t,j}}(x_j\mid \tilde{z}_j)
}{
f_{j,a}^2(x_j\mid \tilde{z}_j)
}
\left|
f_j(x_j\mid \tilde{z}_j)
-
f_{X_{t+1,j}\mid \tilde{Z}_{t,j}}(x_j\mid \tilde{z}_j)
\right|^2
dx_jdP_{\tilde{Z}_{t,j}}(\tilde{z}_j)da\\
&=
-\int_{\tilde{\mathcal{Z}}_j}
\int_{\mathcal{X}_j}
\left[
f_j(x_j\mid \tilde{z}_j)
-
f_{X_{t+1,j}\mid \tilde{Z}_{t,j}}(x_j\mid \tilde{z}_j)
\right]
dx_jdP_{\tilde{Z}_{t,j}}(\tilde{z}_j) + \\
& 
\int_0^1 (1-a)
\int_{\tilde{\mathcal{Z}}_j}
\int_{\mathcal{X}_j}
\frac{
f_{X_{t+1,j}\mid \tilde{Z}_{t,j}}(x_j\mid \tilde{z}_j)
}{
f_{j,a}^2(x_j\mid \tilde{z}_j)
}
\left|
f_j(x_j\mid \tilde{z}_j)
-
f_{X_{t+1,j}\mid \tilde{Z}_{t,j}}(x_j\mid \tilde{z}_j)
\right|^2
dx_jdP_{\tilde{Z}_{t,j}}(\tilde{z}_j)da \\
&=
\int_{\tilde{\mathcal{Z}}_j}
\int_{\mathbb{R}\setminus\mathcal{X}_j}
f_j(x_j\mid \tilde{z}_j)
dx_jdP_{\tilde{Z}_{t,j}}(\tilde{z}_j) + \\
& 
\int_0^1 (1-a)
\int_{\tilde{\mathcal{Z}}_j}
\int_{\mathcal{X}_j}
\frac{
f_{X_{t+1,j}\mid \tilde{Z}_{t,j}}(x_j\mid \tilde{z}_j)
}{
f_{j,a}^2(x_j\mid \tilde{z}_j)
}
\left|
f_j(x_j\mid \tilde{z}_j)
-
f_{X_{t+1,j}\mid \tilde{Z}_{t,j}}(x_j\mid \tilde{z}_j)
\right|^2
dx_jdP_{\tilde{Z}_{t,j}}(\tilde{z}_j)da,
\end{align*} where the first equality is by differentiating~\eqref{eqn:Kj} along the path $f_{j,a}$ and applying~\eqref{eqn:taylor_theorem_integral_remainder}, the second equality is by cancellation because $f_{j,0}(x_j \mid \tilde{z}_j)=f_{X_{t+1,j} \mid \tilde{Z}_{t,j}}(x_j \mid \tilde{z}_j)$ as noted above, and the third equality is because \begin{align*}&-\int_{\tilde{\mathcal{Z}}_j}
\int_{\mathcal{X}_j}
\left[
f_j(x_j\mid \tilde{z}_j)
-
f_{X_{t+1,j}\mid \tilde{Z}_{t,j}}(x_j\mid \tilde{z}_j)
\right]
dx_jdP_{\tilde{Z}_{t,j}}(\tilde{z}_j) 
\\&
=  \int_{\tilde{\mathcal{Z}}_j}\left[\int_{\mathcal{X}_j}f_{X_{t+1,j}\mid \tilde{Z}_{t,j}}(x_j\mid \tilde{z}_j)dx_j - 
\int_{\mathcal{X}_j}
f_j(x_j\mid \tilde{z}_j)dx_j\right] dP_{\tilde{Z}_{t,j}}(\tilde{z}_j) 
\\&
= 
\int_{\tilde{\mathcal{Z}}_j} \left[ 1 -
\int_{\mathcal{X}_j}
f_j(x_j\mid \tilde{z}_j)
dx_j\right]dP_{\tilde{Z}_{t,j}}(\tilde{z}_j)
\\&
= 
\int_{\tilde{\mathcal{Z}}_j} \left[ \left(\int_{\mathcal{X}_j}
f_j(x_j\mid \tilde{z}_j)
dx_j + \int_{\mathbb{R} \setminus \mathcal{X}_j}
f_j(x_j\mid \tilde{z}_j)
dx_j\right) -
\int_{\mathcal{X}_j}
f_j(x_j\mid \tilde{z}_j)
dx_j\right]dP_{\tilde{Z}_{t,j}}(\tilde{z}_j)
\\&
= 
\int_{\tilde{\mathcal{Z}}_j}
\int_{\mathbb{R} \setminus \mathcal{X}_j}
f_j(x_j\mid \tilde{z}_j)
dx_jdP_{\tilde{Z}_{t,j}}(\tilde{z}_j),\end{align*}   
where the first equality follows by linearity, the second because $f_{X_{t+1,j} \mid \tilde{Z}_{t,j}}(\cdot \mid \tilde{z}_j)$ integrates to one over the support $\mathcal{X}_j$, and the third because $f_j(\cdot \mid \tilde{z}_j)$ integrates to one over $\mathbb{R}$, and the final equality follows by canceling terms.

Denote $c_{-}=\min(c_f^{-},c_{\mathrm{MDN}}^{-1})$ and $c_{+}=\max(c_f^{+},c_{\mathrm{MDN}})$, where the constants $c_f^{-},c_f^{+},c_{\mathrm{MDN}}$ are from Assumptions~\ref{asmpt:density_regularity_conditions} and~\ref{asmpt:mdn_model}. Note that $f_{j,a} \in [c_{-},c_{+}]$, so we may upper and lower bound $K_j(f_j)$ by
\begin{align}\label{eqn:Kj_upper_bound}&K_j(f_j) \leq \int_{\tilde{\mathcal{Z}}_j}
\int_{\mathbb{R}\setminus\mathcal{X}_j}
f_j(x_j\mid \tilde{z}_j)
dx_jdP_{\tilde{Z}_{t,j}}(\tilde{z}_j) + \frac{c_f^{+}}{2c_{-}^2}\norm{f_j-f_{X_{t+1,j}\mid \tilde{Z}_{t,j}}}_{L^2,j}^2,\\ \label{eqn:Kj_lower_bound}
&K_j(f_j) \geq \int_{\tilde{\mathcal{Z}}_j}
\int_{\mathbb{R}\setminus\mathcal{X}_j}
f_j(x_j\mid \tilde{z}_j)
dx_jdP_{\tilde{Z}_{t,j}}(\tilde{z}_j) + \frac{c_f^{-}}{2c_{+}^2} \norm{f_j-f_{X_{t+1,j}\mid \tilde{Z}_{t,j}}}_{L^2,j}^2.\end{align}

Therefore, for $\hat{f}_{j}^{\mathrm{oracle}}$ from the theorem statement and for any DNN approximation in the MDN function class, $f_{j}^{\mathrm{DNN}}\in\mathcal{M}_{n,j}^{\mathrm{MDN}}$, we can upper and lower bound $K_j(\hat{f}_{j}^{\mathrm{oracle}})$ by 
\begin{align*}
    & K_j(\hat{f}_{j}^{\mathrm{oracle}}) \leq (P_j - P_{n,j})\left[\log f_{X_{t+1,j}\mid \tilde{Z}_{t,j}} - \log \hat{f}_{j}^{\mathrm{oracle}}\right] +  P_{n,j}\left[\log f_{X_{t+1,j}\mid \tilde{Z}_{t,j}} - \log f_{j}^{\mathrm{DNN}} \right],    \\
    & K_j(\hat{f}_{j}^{\mathrm{oracle}})\geq \frac{c_f^{-}}{2c_{+}^2} \norm{\hat{f}_{j}^{\mathrm{oracle}}-f_{X_{t+1,j}\mid \tilde{Z}_{t,j}}}_{L^2,j}^2,
\end{align*} where the lower bound follows by dropping the nonnegative first term from~\eqref{eqn:Kj_lower_bound}, and the upper bound follows by adding and subtracting the same term $P_{n,j}[\log f_{X_{t+1,j}\mid \tilde{Z}_{t,j}} - \log \hat{f}_{j}^{\mathrm{oracle}}]$ to~\eqref{eqn:Kj}, where $P_{n,j}$ is from~\eqref{eqn:Pj_Pnj}, linearity, the triangle inequality, and noting that $\hat{f}_{j}^{\mathrm{oracle}}$ from~\eqref{eqn:oracle_mdn_estimator} is the argmax so we may upper bound by any $f_{j}^{\mathrm{DNN}}\in\mathcal{M}_{n,j}^{\mathrm{MDN}}$.

\paragraph*{Step 1.2.} This step corresponds to Step 1.2 in the proof of Theorem 3 in \cite{Zhou_et_al_2023}. We apply the approximation result from Lemma~3 in \cite{Zhou_et_al_2023}.

Fix the dimension $j\in[d]$, number of mixture components $M_j\in\mathbb{N}$, and the approximation level $\epsilon\in(0,1/4]$. By Assumption~\ref{asmpt:density_regularity_conditions}, the
conditional density
$f_{X_{t+1,j}\mid\tilde{Z}_{t,j}}$ is approximated by the mixture \begin{equation}
\label{eqn:ideal_Gaussian_mixture_error}
\sup_{x_j, \tilde{z}_j}
\left|
f_{X_{t+1,j}\mid\tilde{Z}_{t,j}}
(x_j\mid\tilde{z}_j)
- \sum_{g=1}^{M_j}
\frac{\alpha_{g,j}(\tilde{z}_j)}
{\sqrt{2\pi}\sigma_{g,j}(\tilde{z}_j)}
\exp\left(
-\frac{
(x_j-\mu_{g,j}(\tilde{z}_j))^2
}{
2\sigma_{g,j}^2(\tilde{z}_j)
}\right)
\right|
\leq
c_{\mathrm{gmix}}M_j^{-\omega_1}.
\end{equation}

\paragraph*{Step 1.2.1.} 
Next, we verify that the DNN approximation result used in Lemma~3
of \citet{Zhou_et_al_2023} applies to the functions in
\eqref{eqn:ideal_Gaussian_mixture_error}. 
By Assumption~\ref{asmpt:density_regularity_conditions}, for each
$g\in[M_j]$, the functions
$\alpha_{g,j}(\cdot)$, $\mu_{g,j}(\cdot)$, and
$\sigma_{g,j}(\cdot)$ admit $\gamma$-Sobolev-smooth extensions to
the domain
$$ B_j
=
[-b_{\mathcal{X}},b_{\mathcal{X}}]^{k+j-1}.$$  We may map $B_j$ onto $[-1,1]^{k+j-1}$, i.e., by normalizing $B_j / b_{\mathcal{X}}=[-1,1]^{k+j-1}$, so that the Sobolev norms of the functions will remain bounded by a constant that depends on $b_{\mathcal{X}}$, $\gamma$, and $c_{\mathrm{Sob}}$. Hence, Lemma~\ref{lma:lemma2_zhou_approx_dnn} (Lemma~2 in \cite{Zhou_et_al_2023}) applies with input dimension given by $k+j-1$.

Let $\ell=1,2,3$ correspond to
$\alpha_{g,j}$, $\mu_{g,j}$, $\sigma_{g,j}$, respectively. For each $g\in[M_j]$ and
$\ell\in\{1,2,3\}$, let
$$
H_{g,j,\ell}(\epsilon),\quad
W_{g,j,\ell}(\epsilon),\quad
U_{g,j,\ell}(\epsilon),
$$
denote the number of hidden layers, the number of
weights, and the total number of hidden units in the deep neural network (DNN) used to approximate the $g$th mixture
coefficient function.

Lemma~\ref{lma:lemma2_zhou_approx_dnn} (Lemma~2 in \cite{Zhou_et_al_2023}) with approximation level $\epsilon/M_j$ yields the approximation
\begin{align}
\label{eqn:coefficient_DNN_approximation}
&
\left\|
\alpha_{g,j}^{\mathrm{DNN}}
-
\alpha_{g,j}
\right\|_\infty
\leq
\frac{\epsilon}{M_j},
\quad
\left\|
\mu_{g,j}^{\mathrm{DNN}}
-
\mu_{g,j}
\right\|_\infty
\leq
\frac{\epsilon}{M_j},
\quad
\left\|
\sigma_{g,j}^{\mathrm{DNN}}
-
\sigma_{g,j}
\right\|_\infty
\leq
\frac{\epsilon}{M_j}.
\end{align} Note that the true mean and scale satisfy the bounds from Assumption~\ref{asmpt:density_regularity_conditions}, so we may clip the DNN approximations to the ranges from Assumption~\ref{asmpt:mdn_model} without increasing the approximation
errors in~\eqref{eqn:coefficient_DNN_approximation}.

For a
$(k+j-1)$-dimensional $\gamma$-Sobolev-smooth function
approximated with accuracy $\delta$, Lemma~\ref{lma:lemma2_zhou_approx_dnn} (Lemma~2 in \cite{Zhou_et_al_2023}) yields the following bounds on the number of hidden layers, weights, and hidden units, 
$$
H(\delta)
\leq
C(\log(1/\delta)+1),\quad 
W(\delta),U(\delta)
\leq
C
\delta^{-(k+j-1)/\gamma}
(\log(1/\delta)+1).$$

Letting $\delta=\epsilon/M_j$ yields $$
\delta^{-(k+j-1)/\gamma}
=
M_j^{(k+j-1)/\gamma}
\epsilon^{-(k+j-1)/\gamma},
$$ so there exists a constant $C>0$, independent of
$j$, $M_j$, and $\epsilon$, such that, for all $\ell\in \{1,2,3\}$,
\begin{equation}\begin{aligned}
\label{eqn:DNN_depth_parameter_bounds}
H_{g,j,\ell}(\epsilon)
&\leq
C
(
\log\left(\frac{M_j}{\epsilon}\right)+1
),\\
W_{g,j,\ell}(\epsilon)
&\leq
C
M_j^{(k+j-1)/\gamma}
\epsilon^{-(k+j-1)/\gamma}
(
\log\left(\frac{M_j}{\epsilon}\right)+1
),
\\
U_{g,j,\ell}(\epsilon)
&\leq
C
M_j^{(k+j-1)/\gamma}
\epsilon^{-(k+j-1)/\gamma}
(
\log\left(\frac{M_j}{\epsilon}\right)+1
).
\end{aligned}
\end{equation}

Note that Lemma~\ref{lma:lemma3_in_zhou_et_al_mdn_approx} (Lemma~3 in \citet{Zhou_et_al_2023}) applies because
Assumption~\ref{asmpt:density_regularity_conditions} gives the
same uniform bounds on the functions in the Gaussian mixture, and Assumption~\ref{asmpt:compact_X}
implies
$\sup_{x_j\in\mathcal{X}_j}|x_j|<\infty$. Also, note that an inspection of the proof of Lemma~\ref{lma:lemma3_in_zhou_et_al_mdn_approx} shows that replacing the bound $|x_j|\leq 1$ by $|x_j|\leq b_{\mathcal{X}}$ only changes the constants.

Next, we introduce the approximating MDN, $f_{j,\epsilon}^{\mathrm{app}}$, which is obtained by replacing the (infeasible) functions $\alpha_{g,j}(\cdot)$, $\mu_{g,j}(\cdot)$, and $\sigma_{g,j}(\cdot)$ from
Assumption~\ref{asmpt:density_regularity_conditions} by the DNN
approximations. Specifically,
\begin{equation}\label{eqn:fjepsilon_app}
f_{j,\epsilon}^{\mathrm{app}}
(x_j\mid\tilde{z}_j)
=
\sum_{g=1}^{M_j}
\alpha_{g,j}^{\mathrm{app}}(\tilde{z}_j)
\frac{1}
{\sqrt{2\pi}\sigma_{g,j}^{\mathrm{DNN}}(\tilde{z}_j)}
\exp\left(
-\frac{
\left(
x_j-\mu_{g,j}^{\mathrm{DNN}}(\tilde{z}_j)
\right)^2
}{
2\left(
\sigma_{g,j}^{\mathrm{DNN}}(\tilde{z}_j)
\right)^2
}
\right),\end{equation}
where $\alpha_{g,j}^{\mathrm{app}}(\cdot)$ are the normalized DNN
approximations of the mixture weights defined as \begin{equation}\label{eqn:aclip}\alpha_{g,j}^{\mathrm{app}}
=
\frac{a_{g,j}^{\mathrm{clip}}}
{\sum_{h=1}^{M_j}a_{h,j}^{\mathrm{clip}}},\quad a_{g,j}^{\mathrm{clip}}
=
\max(
\alpha_{g,j}^{\mathrm{DNN}}
-
\frac{\epsilon}{M_j}
,0).
\end{equation}

By \eqref{eqn:coefficient_DNN_approximation}, we have
\begin{equation}\label{eqn:mixture_weights_basic_inequalities} 0
\leq
a_{g,j}^{\mathrm{clip}}
\leq
\alpha_{g,j},
\quad
\alpha_{g,j}
-
a_{g,j}^{\mathrm{clip}}
\leq
\frac{2\epsilon}{M_j}.\end{equation}
Since the mixture weights sum to one, we have, by~\eqref{eqn:mixture_weights_basic_inequalities}, summing over the mixture components, and subtraction, that
\begin{equation}\label{eqn:sum_of_mixture_in_approx_a} 1-2\epsilon \leq
\sum_{g=1}^{M_j}a_{g,j}^{\mathrm{clip}}
\leq 1.\end{equation}

Next, we will bound the quantity \begin{equation}\label{eqn:sum_mixture_absolute_difference_to_bound}
\sum_{g=1}^{M_j}
\left|
\alpha_{g,j}^{\mathrm{app}}(\tilde{z}_j)
-
\alpha_{g,j}^{\mathrm{DNN}}(\tilde{z}_j)
\right|. 
\end{equation} 

By the triangle inequality, we have
\begin{align}
\sum_{g=1}^{M_j}
\left|
\alpha_{g,j}^{\mathrm{app}}(\tilde{z}_j)
-
\alpha_{g,j}^{\mathrm{DNN}}(\tilde{z}_j)
\right| \leq
\sum_{g=1}^{M_j}
\left|
\alpha_{g,j}^{\mathrm{app}}(\tilde{z}_j)
-
a_{g,j}^{\mathrm{clip}}(\tilde{z}_j)
\right|
+
\sum_{g=1}^{M_j}
\left|
a_{g,j}^{\mathrm{clip}}(\tilde{z}_j)
-
\alpha_{g,j}^{\mathrm{DNN}}(\tilde{z}_j)
\right|,
\label{eqn:mixture_weight_triangle}
\end{align} so we will consider each term separately.

\paragraph*{Step 1.2.1.1.} For the first term in~\eqref{eqn:mixture_weight_triangle}, since
$$
\alpha_{g,j}^{\mathrm{app}}(\tilde{z}_j)
=
\frac{
a_{g,j}^{\mathrm{clip}}(\tilde{z}_j)
}{
\sum_{h=1}^{M_j} a_{h,j}^{\mathrm{clip}}(\tilde{z}_j)
},
$$
and $$\sum_{h=1}^{M_j} a_{h,j}^{\mathrm{clip}}(\tilde{z}_j)\leq 1,$$ we have
\begin{align}\label{eqn:mixture_weight_normalization_error}
\sum_{g=1}^{M_j}
\left|
\alpha_{g,j}^{\mathrm{app}}(\tilde{z}_j)
-
a_{g,j}^{\mathrm{clip}}(\tilde{z}_j)
\right| \leq
2\epsilon,
\end{align} because, by distributing terms and rearranging the terms from~\eqref{eqn:sum_of_mixture_in_approx_a}, we have $$\sum_{g=1}^{M_j}
\left|
\alpha_{g,j}^{\mathrm{app}}(\tilde{z}_j)
-
a_{g,j}^{\mathrm{clip}}(\tilde{z}_j)
\right|
=
\sum_{g=1}^{M_j}
a_{g,j}^{\mathrm{clip}}(\tilde{z}_j)
\left(
\frac{1}{\sum_{h=1}^{M_j} a_{h,j}^{\mathrm{clip}}(\tilde{z}_j)}-1
\right)
=
1-\sum_{h=1}^{M_j} a_{h,j}^{\mathrm{clip}}(\tilde{z}_j)
\leq
2\epsilon.$$

\paragraph*{Step 1.2.1.2.} For the second term in~\eqref{eqn:mixture_weight_triangle}, we must consider two cases. 

First, if
$\alpha_{g,j}^{\mathrm{DNN}}(\tilde{z}_j)
\geq \epsilon/M_j$, then
$$
\left|
a_{g,j}^{\mathrm{clip}}(\tilde{z}_j)
-
\alpha_{g,j}^{\mathrm{DNN}}(\tilde{z}_j)
\right|
=
\frac{\epsilon}{M_j},$$ since 
$a_{g,j}^{\mathrm{clip}}(\tilde{z}_j)
=
\max\left(
\alpha_{g,j}^{\mathrm{DNN}}(\tilde{z}_j)
-
\frac{\epsilon}{M_j},
0
\right)
$.

Second, if $\alpha_{g,j}^{\mathrm{DNN}}(\tilde{z}_j)
<
\epsilon/M_j$, then  \begin{equation}\label{eqn:a_zero}a_{g,j}^{\mathrm{clip}}(\tilde{z}_j)=0.\end{equation} Since
$\alpha_{g,j}(\tilde{z}_j)\geq0$ and, by
\eqref{eqn:coefficient_DNN_approximation}, we have
$$
\left|
\alpha_{g,j}^{\mathrm{DNN}}(\tilde{z}_j)
-
\alpha_{g,j}(\tilde{z}_j)
\right|
\leq
\frac{\epsilon}{M_j},
$$ then we must have
\begin{equation}\label{eqn:alpha_geq_neg}
\alpha_{g,j}^{\mathrm{DNN}}(\tilde{z}_j)
\geq
-\frac{\epsilon}{M_j}.\end{equation}
Therefore, by~\eqref{eqn:a_zero} and~\eqref{eqn:alpha_geq_neg}, we have
$$\left|
a_{g,j}^{\mathrm{clip}}(\tilde{z}_j)
-
\alpha_{g,j}^{\mathrm{DNN}}(\tilde{z}_j)
\right| \leq \left|0+\frac{\epsilon}{M_j}\right|
=
\frac{\epsilon}{M_j}.$$

Hence, in both cases, we have
$$
\left|
a_{g,j}^{\mathrm{clip}}(\tilde{z}_j)
-
\alpha_{g,j}^{\mathrm{DNN}}(\tilde{z}_j)
\right| \leq 
\frac{\epsilon}{M_j},
$$
so summing over the mixture components yields 
\begin{equation}
\label{eqn:mixture_weight_clipping_error}
\sum_{g=1}^{M_j}
\left|
a_{g,j}^{\mathrm{clip}}(\tilde{z}_j)
-
\alpha_{g,j}^{\mathrm{DNN}}(\tilde{z}_j)
\right|
\leq
\epsilon.
\end{equation}

Combining
\eqref{eqn:mixture_weight_triangle},
\eqref{eqn:mixture_weight_normalization_error}, and
\eqref{eqn:mixture_weight_clipping_error}, we obtain
\begin{equation}
\label{eqn:sum_mixture_absolute_difference}
\sum_{g=1}^{M_j}
\left|
\alpha_{g,j}^{\mathrm{app}}(\tilde{z}_j)
-
\alpha_{g,j}^{\mathrm{DNN}}(\tilde{z}_j)
\right|
\leq
3\epsilon.
\end{equation}

\paragraph*{Step 1.2.1.3.} By \eqref{eqn:coefficient_DNN_approximation}, we have the DNN approximation  \begin{equation}\label{eqn:alpha_dnn_approx_true}
\sum_{g=1}^{M_j}
\left|
\alpha_{g,j}^{\mathrm{DNN}}(\tilde{z}_j)
-
\alpha_{g,j}(\tilde{z}_j)
\right|
\leq
M_j\frac{\epsilon}{M_j}
=
\epsilon.
\end{equation}
Therefore, by the triangle inequality,~\eqref{eqn:sum_mixture_absolute_difference}, and~\eqref{eqn:alpha_dnn_approx_true}, we have
$$
\sum_{g=1}^{M_j}
\left|
\alpha_{g,j}^{\mathrm{app}}(\tilde{z}_j)
-
\alpha_{g,j}(\tilde{z}_j)
\right|
\leq 3\epsilon + \epsilon =
4\epsilon.
$$

Note that the quantity below is bounded,
$$
\sup_{x_j,\tilde{z}_j,g}
\frac{1}
{\sqrt{2\pi}\sigma_{g,j}^{\mathrm{DNN}}(\tilde{z}_j)}
\exp\left(
-\frac{(x_j-\mu_{g,j}^{\mathrm{DNN}}(\tilde{z}_j))^2}
{2(\sigma_{g,j}^{\mathrm{DNN}}(\tilde{z}_j))^2}
\right)
\leq
C M_j^{\omega_2}.
$$ By~\eqref{eqn:sum_mixture_absolute_difference} and the previous bound on the components of the Gaussian mixture, replacing the DNN
mixture weights $\alpha_{g,j}^{\mathrm{DNN}}$ by the normalized
weights $\alpha_{g,j}^{\mathrm{app}}$ introduces an additional error
of at most \begin{equation}\label{eqn:error_approx_dnn_bound}
C M_j^{\omega_2}\epsilon,
\end{equation}
for some constant $C>0$.

Therefore, by the triangle inequality, Lemma~\ref{lma:lemma3_in_zhou_et_al_mdn_approx} (with input dimension $k+j-1$), and the approximation bound from~\eqref{eqn:error_approx_dnn_bound}, there exists an MDN
$f_{j,\epsilon}^{\mathrm{app}}$ from~\eqref{eqn:fjepsilon_app} satisfying
\begin{equation}
\label{eqn:MDN_DNN_approximation}
\sup_{x_j\in\mathcal{X}_j,
      \tilde{z}_j\in\tilde{\mathcal{Z}}_j}
\left|
f_{j,\epsilon}^{\mathrm{app}}
(x_j\mid\tilde{z}_j)
-
f_{X_{t+1,j}\mid\tilde{Z}_{t,j}}
(x_j\mid\tilde{z}_j)
\right|
\leq
C
(
M_j^{4\omega_2}\epsilon
+
M_j^{-\omega_1}
).
\end{equation}

Note that, for sufficiently small $\epsilon$ and sufficiently large $c_{\mathrm{MDN}}$, the construction of $f_{j,\epsilon}^{\mathrm{app}}$ from~\eqref{eqn:fjepsilon_app} belongs to the function class $\mathcal{M}_{n,j}^{\mathrm{MDN}}$ from Assumption~\ref{asmpt:mdn_model}. 

\paragraph*{Step 1.2.2.} Next, we check that the architecture above yields the parameter count in Assumption~\ref{asmpt:rate_of_growth_MDN_num_param}.
There are $3M_j$ networks in the $j$th MDN, i.e., one network for each of
$\alpha_{g,j}$, $\mu_{g,j}$, and $\sigma_{g,j}$,
$g\in[M_j]$. Therefore, denoting the total number of parameters in the $j$th
MDN by $N_{j,\epsilon}^{\mathrm{par}}$, we have \begin{align}\label{eqn:total_MDN_parameter_bound}
N_{j,\epsilon}^{\mathrm{par}}
&
\leq
C M_j
\max_{\substack{g\in[M_j]\\ \ell\in\{1,2,3\}}}
(W_{g,j,\ell}(\epsilon) + U_{g,j,\ell}(\epsilon))
\\&
\leq
C
M_j^{(k+j-1+\gamma)/\gamma}
\epsilon^{-(k+j-1)/\gamma}
(
\log\left(\frac{M_j}{\epsilon}\right)+1
).
\end{align}
Letting
$$\epsilon
=
n^{-\gamma/(2\gamma+k+j-1)},$$ so that $$
\epsilon^{-(k+j-1)/\gamma}
=
n^{(k+j-1)/(2\gamma+k+j-1)},\quad
\log(M_j/\epsilon)
\leq
C\log(nM_j),$$
we have
\begin{equation}
\label{eqn:total_MDN_parameter_bound_n}
N_{j,\epsilon}^{\mathrm{par}}
\leq
C
M_j^{(k+j-1+\gamma)/\gamma}
n^{(k+j-1)/(2\gamma+k+j-1)}
\log(nM_j),
\end{equation}
as required by
Assumption~\ref{asmpt:rate_of_growth_MDN_num_param}.

\paragraph*{Step 1.2.3.} Next, we derive two upper bounds that will be used later.

First, we upper bound the quantity
$$\int_{\tilde{\mathcal{Z}}_j}
\int_{\mathbb{R}\setminus\mathcal{X}_j}
f_{j,\epsilon}^{\mathrm{app}}
(x_j\mid\tilde{z}_j)
dx_j
dP_{\tilde{Z}_{t,j}}(\tilde{z}_j),$$ where the approximating MDN $f_{j,\epsilon}^{\mathrm{app}}$ is from~\eqref{eqn:fjepsilon_app}.

For each \(g\in[M_j]\) and \(\tilde{z}_j\in\tilde{\mathcal{Z}}_j\), let $$
P_{g,j,\tilde{z}_j}
=
N\left(
\mu_{g,j}(\tilde{z}_j),
\sigma_{g,j}^2(\tilde{z}_j)
\right),$$
denote the Gaussian distribution corresponding to the mixture from
Assumption~\ref{asmpt:density_regularity_conditions}, and let
$$P_{g,j,\tilde{z}_j}^{\mathrm{app}}=N\left(
\mu_{g,j}^{\mathrm{DNN}}(\tilde{z}_j),
(\sigma_{g,j}^{\mathrm{DNN}})^2(\tilde{z}_j)
\right),$$ denote the corresponding Gaussian
distribution obtained by replacing $\mu_{g,j}$ and $\sigma_{g,j}$
with their DNN approximations from
\eqref{eqn:coefficient_DNN_approximation}.

Denote the squared Hellinger
distance between two distributions \(P\) and \(Q\), with densities
\(p\) and \(q\), by
$$
H^2(P,Q)
=
\int_{\mathbb{R}}
\left(\sqrt{p(x)}-\sqrt{q(x)}\right)^2dx.
$$

By the reverse triangle inequality, we have $$
\sqrt{P_{g,j,\tilde{z}_j}^{\mathrm{app}}(\mathbb{R}\setminus\mathcal{X}_j)}
\leq
\sqrt{P_{g,j,\tilde{z}_j}(\mathbb{R}\setminus\mathcal{X}_j)}
+
H(P_{g,j,\tilde{z}_j}^{\mathrm{app}},
  P_{g,j,\tilde{z}_j}).$$

Squaring the previous inequality and using the inequality
$(a+b)^2\leq 2a^2+2b^2$ yields
$$
P_{g,j,\tilde{z}_j}^{\mathrm{app}}
(\mathbb{R}\setminus\mathcal{X}_j)
\leq
2P_{g,j,\tilde{z}_j}
(\mathbb{R}\setminus\mathcal{X}_j)
+
2H^2\left(
P_{g,j,\tilde{z}_j}^{\mathrm{app}},
P_{g,j,\tilde{z}_j}
\right).
$$

The squared Hellinger distance between Gaussians (which is known in closed form), and the lower bounds
$$\sigma_{g,j}(\tilde{z}_j)
\geq c_{\sigma}^-M_j^{-\omega_2},
\quad
\sigma_{g,j}^{\mathrm{DNN}}(\tilde{z}_j)
\geq c_{\mathrm{MDN}}^{-1}M_j^{-\omega_2},$$
from Assumptions~\ref{asmpt:density_regularity_conditions}
and~\ref{asmpt:mdn_model}, and
\eqref{eqn:coefficient_DNN_approximation} yield $$H^2(P_{g,j,\tilde{z}_j}^{\mathrm{app}},
    P_{g,j,\tilde{z}_j})
\leq
C M_j^{2\omega_2}
\left(\frac{\epsilon}{M_j}\right)^2.$$

By~\eqref{eqn:aclip} and~\eqref{eqn:sum_of_mixture_in_approx_a}, we have
$$
\alpha_{g,j}^{\mathrm{app}}
=
\frac{a_{g,j}^{\mathrm{clip}}}
{\sum_{h=1}^{M_j}a_{h,j}^{\mathrm{clip}}} \leq \frac{a_{g,j}^{\mathrm{clip}}}{1-2\epsilon} \leq \frac{\alpha_{g,j}}{1-2\epsilon}.
$$

Using the corresponding DNN approximation of the mixture weights,
which satisfy \begin{equation}\label{eqn:bound_alpha}\alpha_{g,j}^{\mathrm{app}}\leq  2 \alpha_{g,j},\end{equation} because $1/(1-2\epsilon)\in (1,2]$ since $\epsilon \in (0,1/4]$, summing over
\(g\in[M_j]\), integrating over \(\tilde{z}_j\), and applying the upper bounds from Assumption~\ref{asmpt:control_MDN_prob_outside_support} yields
$$
\int_{\tilde{\mathcal{Z}}_j}
\int_{\mathbb{R}\setminus\mathcal{X}_j}
f_{j,\epsilon}^{\mathrm{app}}
(x_j\mid\tilde{z}_j)
dx_jdP_{\tilde{Z}_{t,j}}(\tilde{z}_j)
\leq
C M_j^{-2\omega_1}
+
C M_j^{2\omega_2-2}\epsilon^2,$$ where $f_{j,\epsilon}^{\mathrm{app}}$ is from~\eqref{eqn:fjepsilon_app}. Since $M_j\geq 1$ and $\omega_2>0$, we have
$$
M_j^{2\omega_2-2}\epsilon^2
\leq
M_j^{8\omega_2}\epsilon^2.$$
Therefore, we have the desired upper bound \begin{equation}
\label{eqn:DNN_MDN_tail_bound}
\int_{\tilde{\mathcal{Z}}_j}
\int_{\mathbb{R}\setminus\mathcal{X}_j}
f_{j,\epsilon}^{\mathrm{app}}
(x_j\mid\tilde{z}_j)
dx_j
dP_{\tilde{Z}_{t,j}}(\tilde{z}_j)
\leq
C(M_j^{4\omega_2}\epsilon + M_j^{-\omega_1})^2.
\end{equation}

Second, we upper bound the quantity from \eqref{eqn:Kj_upper_bound} in Step 1.1. By \eqref{eqn:MDN_DNN_approximation} and the compactness of
$\mathcal{X}_j$, we have
\begin{equation}
\label{eqn:L2_MDN_approximation}
\left\|
f_{j,\epsilon}^{\mathrm{app}}
-
f_{X_{t+1,j}\mid\tilde{Z}_{t,j}}
\right\|_{L^2,j}^2
\leq
C(M_j^{4\omega_2}\epsilon + M_j^{-\omega_1})^2,
\end{equation}  where $f_{j,\epsilon}^{\mathrm{app}}$ is from~\eqref{eqn:fjepsilon_app}. By~\eqref{eqn:DNN_MDN_tail_bound} and~\eqref{eqn:L2_MDN_approximation}, the quantity from~\eqref{eqn:Kj_upper_bound} in Step 1.1 can be bounded by
\begin{equation}
\label{eqn:population_approximation_KL}
K_j(f_{j,\epsilon}^{\mathrm{app}})
\leq
C(M_j^{4\omega_2}\epsilon + M_j^{-\omega_1})^2.
\end{equation}

\paragraph*{Step 1.2.4.} 
Next, we obtain the final approximation bound as in Step 1.2 of \citet{Zhou_et_al_2023}. By Assumptions~\ref{asmpt:density_regularity_conditions} and
\ref{asmpt:mdn_model}, both densities are uniformly bounded away
from zero on
$\mathcal{X}_j\times\tilde{\mathcal{Z}}_j$, so the mean value theorem applied to the logarithm and~\eqref{eqn:MDN_DNN_approximation} imply that
$$|\log
f_{X_{t+1,j}\mid\tilde{Z}_{t,j}}
(X_{t+1,j}\mid\tilde{Z}_{t,j})
-
\log
f_{j,\epsilon}^{\mathrm{app}}
(X_{t+1,j}\mid\tilde{Z}_{t,j})|
\leq
C(M_j^{4\omega_2}\epsilon + M_j^{-\omega_1}),$$ almost surely, and that
$$
\mathrm{Var}(\log
f_{X_{t+1,j}\mid\tilde{Z}_{t,j}}
(X_{t+1,j}\mid\tilde{Z}_{t,j})
-
\log
f_{j,\epsilon}^{\mathrm{app}}
(X_{t+1,j}\mid\tilde{Z}_{t,j}))
\leq
C(M_j^{4\omega_2}\epsilon + M_j^{-\omega_1})^2.$$

Recall $P_{n,j}$ from~\eqref{eqn:Pj_Pnj}. Under the setting of independent observations from Step 1, Bernstein's inequality gives, for every
$\tilde{r}>0$, with probability at least
$1-\exp(-\tilde{r})$,
\begin{align}
&
P_{n,j}
\left[
\log f_{X_{t+1,j}\mid\tilde{Z}_{t,j}}
-
\log f_{j,\epsilon}^{\mathrm{app}}
\right]
\nonumber\\
&\leq
C(M_j^{4\omega_2}\epsilon + M_j^{-\omega_1})^2
+
C(M_j^{4\omega_2}\epsilon + M_j^{-\omega_1})
\sqrt{\frac{\tilde{r}}{N}}
+
C(M_j^{4\omega_2}\epsilon + M_j^{-\omega_1})
\frac{\tilde{r}}{N}.
\label{eqn:empirical_MDN_approximation_bound}
\end{align}

\paragraph*{Step 1.3.}  This step corresponds to Step 1.3 in the proof of Theorem 3 in \cite{Zhou_et_al_2023}.

We bound the quantity $$
(P_j-P_{n,j})
\left[
\log f_{X_{t+1,j}\mid\tilde{Z}_{t,j}}
-
\log\hat{f}_j^{\mathrm{oracle}}
\right],
$$
from Step 1.1 using arguments analogous to those from \citet{Zhou_et_al_2023} who themselves use the ideas from \citet{dnn_est_inf}.

\paragraph*{Step 1.3.1.}
For a constant $r_{0,j}>0$, define $$
\mathcal{M}_{n,j}^{\mathrm{MDN}}(r_{0,j})
=
\left\{
f_j\in\mathcal{M}_{n,j}^{\mathrm{MDN}}:
\left\|f_j-f_{X_{t+1,j}\mid\tilde{Z}_{t,j}}\right\|_{L^2,j}
\leq r_{0,j} M_j^{\omega_2}
\right\}.
$$
Note that, since $\mathcal{X}_j$ is compact and both densities are uniformly bounded by $c_+$, we can select $r_{0,j}>0$ sufficiently large so that $$ \left\| f_j-f_{X_{t+1,j}\mid\tilde{Z}_{t,j}} \right\|_{L^2,j} \leq r_{0,j}M_j^{\omega_2},$$ for every $f_j\in\mathcal{M}_{n,j}^{\mathrm{MDN}}$, since $M_j^{\omega_2}\geq 1$. Define the corresponding classes of differences by
$$
\begin{aligned}
\mathcal{H}_j(r_{0,j})
&=
\left\{
f_j-f_{X_{t+1,j}\mid\tilde{Z}_{t,j}}:
f_j\in\mathcal{M}_{n,j}^{\mathrm{MDN}}(r_{0,j})
\right\}\cup\{0\},
\\
\mathcal{G}_j(r_{0,j})
&=
\left\{
\log f_{X_{t+1,j}\mid\tilde{Z}_{t,j}}-\log f_j:
f_j\in\mathcal{M}_{n,j}^{\mathrm{MDN}}(r_{0,j})
\right\}\cup\{0\}.
\end{aligned}
$$

Recall $c_-$ and $c_+$ from Step 1.1. By Assumption~\ref{asmpt:density_regularity_conditions}, we have 
\begin{equation}
\label{eqn:mdn_population_norm_comparison}
c_f^-\|h\|_{L^2,j}^2
\leq P_jh^2
\leq c_f^+\|h\|_{L^2,j}^2,
\end{equation}
for any measurable function $h$, because $$
P_jh^2
=
\int_{\tilde{\mathcal{Z}}_j}
\int_{\mathcal{X}_j}
|h(x_j,\tilde{z}_j)|^2
f_{X_{t+1,j}\mid\tilde{Z}_{t,j}}(x_j\mid\tilde{z}_j)
dx_j dP_{\tilde{Z}_{t,j}}(\tilde{z}_j),
$$

Thus, by the mean value theorem for the logarithm, we have $$
\begin{aligned}
&\left|
\log f_{X_{t+1,j}\mid\tilde{Z}_{t,j}}(x_j\mid\tilde{z}_j)
-\log f_j(x_j\mid\tilde{z}_j)
\right|
\leq
c_-^{-1}
\left|
f_j(x_j\mid\tilde{z}_j)
-f_{X_{t+1,j}\mid\tilde{Z}_{t,j}}(x_j\mid\tilde{z}_j)
\right|,
\end{aligned}
$$ so $$
\sup_{g_j\in\mathcal{G}_j(r_{0,j})}P_jg_j^2
\leq \frac{c_f^+}{c_-^2}r_{0,j}^2 M_j^{2\omega_2},
\quad
\sup_{g_j\in\mathcal{G}_j(r_{0,j})}\|g_j\|_\infty
\leq\log(c_+/c_-).
$$

\paragraph*{Step 1.3.2.}
First, we upper bound the quantity $$\sup_{g_j\in\mathcal{G}_j(r_{0,j})}|(P_j-P_{n,j})g_j|.$$
For some function class $\mathcal{A}$ and iid Rademacher random variables $\zeta_t$, independent of the data, denote
$$
\mathbb{E}_\zeta R_{n,j}\mathcal{A}
=
\mathbb{E}_\zeta\left[
\sup_{h\in\mathcal{A}}
\left|\frac{1}{N}\sum_{t=m+1}^{n-m-1}
\zeta_t h(X_{t+1,j},\tilde{Z}_{t,j})\right|
\right].
$$

By Theorem 2.1 of \citet{Bartlett_et_al_2005}, for every $\tilde{r}>0$, with probability at least $1-2e^{-\tilde{r}}$, we have
\begin{equation}
\label{eqn:mdn_local_concentration}
\sup_{g_j\in\mathcal{G}_j(r_{0,j})}|(P_j-P_{n,j})g_j|
\leq C\left(
\mathbb{E}_\zeta R_{n,j}\mathcal{G}_j(r_{0,j})
+r_{0,j} M_j^{\omega_2}\sqrt{\frac{\tilde{r}}{N}}
+M_j^{\omega_2} \frac{ \tilde{r}}{N}\right),
\end{equation} for some constant $C>0$.

\paragraph*{Step 1.3.3.} We split this step into two parts.

\paragraph*{Step 1.3.3.1.}

Let $\mathcal{M}_{n,j}^{\mathrm{DNN}}$ be the ReLU DNN
function class with the architecture from~\eqref{eqn:DNN_depth_parameter_bounds}. Let $\log \mathcal{T}(u,\mathcal{A},\|\cdot\|_n)$ denote the metric entropy, and let $\mathrm{Pdim}(\mathcal{M}_{n,j}^{\mathrm{DNN}})$ denote the pseudo VC dimension of the function class. Observe that, for $u\in (0,2c_+]$, we have \begin{equation}
\label{eqn:mdn_empirical_covering_bound}
\begin{aligned}
\log\mathcal{T}(u,\mathcal{M}_{n,j}^{\mathrm{MDN}},\|\cdot\|_{n,j})
&\leq 3M_j\log\mathcal{T}\left(
\frac{u}{CM_j^{3/2+3\omega_2}},
\mathcal{M}_{n,j}^{\mathrm{DNN}},\|\cdot\|_{n,j}\right)\\
&\leq CM_j\mathrm{Pdim}(\mathcal{M}_{n,j}^{\mathrm{DNN}})
\log\left(\frac{CnM_j^{3/2+3\omega_2}}{u}\right),
\end{aligned}
\end{equation} where the first inequality holds by Lemma~\ref{lma:lemma4_zhou} (Lemma~4 of \citet{Zhou_et_al_2023}), and the second inequality follows by Theorems 12.2 and 14.1 in \cite{Anthony_Bartlett_1999}. Note that the normalization in
Assumption~\ref{asmpt:mdn_model} is chosen so that the normalization map is Lipschitz, so the same arguments in the proof of Lemma~\ref{lma:lemma4_zhou} apply with a change in
constants.


By Lemma~\ref{lem:lemma_6_farrell}, applied to the number of parameters $W_{g,j,\ell}(\epsilon)+U_{g,j,\ell}(\epsilon)$ and the number of hidden layers $H_{g,j,\ell}(\epsilon)$ from~\eqref{eqn:DNN_depth_parameter_bounds}, the pseudo VC dimension of the DNN function class is upper bounded by
\begin{equation}
\label{eqn:mdn_coefficient_pseudoVCdimension}
\mathrm{Pdim}(\mathcal{M}_{n,j}^{\mathrm{DNN}})
\leq
CM_j^{(k+j-1)/\gamma}\epsilon^{-(k+j-1)/\gamma}
(\log(M_j/\epsilon)+1)^3.
\end{equation}

\paragraph*{Step 1.3.3.2.}
We justify the comparison of the population norm and empirical norm (which we use in Step 1.3.4) by following the arguments from Appendix A.2.1 of
\citet{dnn_est_inf}. For every $r_{0,j}>0$ and $\tilde{r}>0$ satisfying
\begin{equation}\label{eqn:mdn_localization_condition}
r_{0,j}M_j^{\omega_2}\geq c_{\mathrm{loc}}\sqrt{ \frac{\max\left(1,M_j\mathrm{Pdim}(\mathcal{M}_{n,j}^{\mathrm{DNN}})\right)\log(nM_j)+\tilde{r}}{N}},
\end{equation} where $c_{\mathrm{loc}}>0$ is a sufficiently large constant, there exists a constant $c_{\mathrm{cmp}}>0$, independent of
$n,M_j,r_{0,j},\tilde{r}$, such that, with probability at least
$1-2e^{-\tilde{r}}$, we have  
\begin{equation}\label{eqn:mdn_population_empirical_inclusion}
\mathcal{H}_j(r_{0,j})
\subseteq
\left\{h:\norm{h}_{n,j}
\leq c_{\mathrm{cmp}}r_{0,j}M_j^{\omega_2}\right\}.
\end{equation}

This class inclusion holds for the following reasons. By~\eqref{eqn:mdn_population_norm_comparison},
$\sup_h P_jh^2\leq c_f^+(r_{0,j}M_j^{\omega_2})^2$; also $\norm{h}_\infty\leq2c_+$
by the density bounds from Step 1.1.
The empirical-Rademacher version of Theorem 2.1 of
\citet{Bartlett_et_al_2005}, applied to the class of squared
functions, and contraction give, with probability at least
$1-2e^{-\tilde{r}}$,
$$
\sup_{h\in\mathcal{H}_j(r_{0,j})}P_{n,j}h^2\leq c_f^+(r_{0,j}M_j^{\omega_2})^2+C\left(
\mathbb{E}_\zeta R_{n,j}\mathcal{H}_j(r_{0,j})
+(r_{0,j}M_j^{\omega_2})\sqrt{\frac{\tilde{r}}{N}}+\frac{\tilde{r}}{N}\right).
$$
Also $\mathrm{Var}(h^2)\leq4c_+^2P_jh^2$ and the square
function is $4c_+$-Lipschitz on $[-2c_+,2c_+]$.
Dudley's inequality and~\eqref{eqn:mdn_empirical_covering_bound} yield
$$
\mathbb{E}_\zeta R_{n,j}\mathcal{H}_j(r_{0,j})
\leq\frac{C}{N}
+C\sqrt{\sup_{h\in\mathcal{H}_j(r_{0,j})}P_{n,j}h^2}\sqrt{\frac{\max\left(1,M_j\mathrm{Pdim}(\mathcal{M}_{n,j}^{\mathrm{DNN}})\right)\log(nM_j)}{N}}.
$$
Combining the previous inequalities and using the inequality $ab\leq\delta a^2+b^2/(4\delta)$ yields $$
\sup_{h\in\mathcal{H}_j(r_{0,j})}P_{n,j}h^2\leq C(r_{0,j}M_j^{\omega_2})^2+
C\frac{\max\left(1,M_j\mathrm{Pdim}(\mathcal{M}_{n,j}^{\mathrm{DNN}})\right)\log(nM_j)+\tilde{r}}{N},
$$ so~\eqref{eqn:mdn_localization_condition} implies~\eqref{eqn:mdn_population_empirical_inclusion}.

\paragraph*{Step 1.3.4.}

Suppose~\eqref{eqn:mdn_localization_condition} holds.
If~\eqref{eqn:mdn_population_empirical_inclusion} holds,
which it does with probability at least $1-2e^{-\tilde{r}}$, the contraction and Dudley chaining arguments from Step 1.3 of \citet{Zhou_et_al_2023} give
the following inequalities: \begin{align*}
&\mathbb{E}_\zeta R_{n,j}\mathcal{G}_j(r_{0,j})
\\
&=
\mathbb{E}_\zeta
\left[
R_{n,j}
\left\{
g_j:
g_j=
\log f_{X_{t+1,j}\mid\tilde{Z}_{t,j}}
-\log f_j,
\right.\right.
\\
&\left.\left.
f_j\in\mathcal{M}_{n,j}^{\mathrm{MDN}},
\left\|
f_j-f_{X_{t+1,j}\mid\tilde{Z}_{t,j}}
\right\|_{L^2,j}
\leq
r_{0,j}M_j^{\omega_2}
\right\}
\right]
\\
&\leq
\frac{2}{c_-}
\mathbb{E}_\zeta
\left[
R_{n,j}
\left\{
f_j-f_{X_{t+1,j}\mid\tilde{Z}_{t,j}}:
f_j\in\mathcal{M}_{n,j}^{\mathrm{MDN}},
\right.\right.
\\
&\left.\left.
\left\|
f_j-f_{X_{t+1,j}\mid\tilde{Z}_{t,j}}
\right\|_{L^2,j}
\leq
r_{0,j}M_j^{\omega_2}
\right\}
\right]
\\
&\leq
\frac{2}{c_-}
\mathbb{E}_\zeta
R_{n,j}
\left\{
f_j-f_{X_{t+1,j}\mid\tilde{Z}_{t,j}}:
f_j\in\mathcal{M}_{n,j}^{\mathrm{MDN}},
\right.
\\
&\left.
\left\|
f_j-f_{X_{t+1,j}\mid\tilde{Z}_{t,j}}
\right\|_{n,j}
\leq
c_{\mathrm{cmp}}r_{0,j}M_j^{\omega_2}
\right\}
\\
&\leq
\frac{2}{c_-}
\inf_{0<\alpha<c_{\mathrm{cmp}}r_{0,j}M_j^{\omega_2}}
\left(
4\alpha
+
\frac{12}{\sqrt{N}}
\int_\alpha^{c_{\mathrm{cmp}}r_{0,j}M_j^{\omega_2}}
\sqrt{1+
\log
\mathcal{T}
\left(
\delta,
\mathcal{M}_{n,j}^{\mathrm{MDN}},
\|\cdot\|_{n,j}
\right)
}
d\delta
\right)
\\
&\leq
\frac{2}{c_-}
\inf_{0<\alpha<c_{\mathrm{cmp}}r_{0,j}M_j^{\omega_2}}
\\& \left(
4\alpha + 
\frac{12}{\sqrt{N}}
\int_\alpha^{c_{\mathrm{cmp}}r_{0,j}M_j^{\omega_2}}
\sqrt{1+
3M_j
\log
\mathcal{T}
\left(
\frac{\delta}
{CM_j^{3/2+3\omega_2}},
\mathcal{M}_{n,j}^{\mathrm{DNN}},
\|\cdot\|_{n,j}
\right)
}
d\delta
\right),
\end{align*}
 where the first inequality follows by contraction since the logarithm
is $c_-^{-1}$-Lipschitz, the second is by the localization
argument used in Step 1.3 of \citet{Zhou_et_al_2023} (following Appendix A.2 of \citet{dnn_est_inf}), i.e., the class inclusion from~\eqref{eqn:mdn_population_empirical_inclusion}, the third is from
Dudley's chaining \citep{Mendelson_2003}, and the fourth is from
\eqref{eqn:mdn_empirical_covering_bound}.

Applying Theorems~12.2 and~14.1 of
\citet{Anthony_Bartlett_1999} as in
\citet{Zhou_et_al_2023} yields
\begin{align*}
&\mathbb{E}_\zeta R_{n,j}\mathcal{G}_j(r_{0,j})
\\ 
&\leq
\frac{2}{c_-}
\inf_{0<\alpha<c_{\mathrm{cmp}}r_{0,j}M_j^{\omega_2}}
\left(
4\alpha
+
\frac{C}{\sqrt{N}}
\int_\alpha^{c_{\mathrm{cmp}}r_{0,j}M_j^{\omega_2}} 
\sqrt{
M_j
\mathrm{Pdim}(\mathcal{M}_{n,j}^{\mathrm{DNN}})
\log\left(
\frac{
CnM_j^{3/2+3\omega_2}
}{
\delta
}
\right)
}
d\delta
\right).
\end{align*} Similar to \cite{Zhou_et_al_2023}, let $$
\alpha=
c_{\mathrm{cmp}}r_{0,j}M_j^{\omega_2}
\min\left(
\frac{1}{2},
\sqrt{
\frac{
M_j\mathrm{Pdim}(\mathcal{M}_{n,j}^{\mathrm{DNN}})
}{
N
}
}
\right),$$ so that, whenever~\eqref{eqn:mdn_localization_condition} holds, there exists a constant $C_4>0$, such that, with probability at least $1-2e^{-\tilde{r}}$, \begin{equation}
\label{eqn:mdn_local_rademacher_bound}
\mathbb{E}_\zeta R_{n,j}\mathcal{G}_j(r_{0,j})
\leq
C_4 r_{0,j}M_j^{\omega_2}
\sqrt{
\frac{
M_j\mathrm{Pdim}(\mathcal{M}_{n,j}^{\mathrm{DNN}})
}{
N
}
\log(nM_j)
}.
\end{equation}

Plugging~\eqref{eqn:mdn_local_rademacher_bound} into~\eqref{eqn:mdn_local_concentration}, we obtain, with
probability at least $1-6e^{-\tilde{r}}$,
\begin{equation}\label{eqn:mdn_local_estimation_bound_iid}
\begin{aligned}
&
\sup_{f_j\in\mathcal{M}_{n,j}^{\mathrm{MDN}}(r_{0,j})}
\left|
(P_j-P_{n,j})
\left[
\log f_{X_{t+1,j}\mid\tilde{Z}_{t,j}}
-\log f_j
\right]
\right|
\\
&\leq C \left(r_{0,j}M_j^{\omega_2}
\sqrt{
\frac{
M_j\mathrm{Pdim}(\mathcal{M}_{n,j}^{\mathrm{DNN}})
}{
N
}
\log(nM_j)
}
+
 r_{0,j}M_j^{\omega_2}
\sqrt{\frac{\tilde{r}}{N}}
+
 M_j^{\omega_2}
\frac{\tilde{r}}{N}\right),
\end{aligned}\end{equation} for some constant $C>0$.

\paragraph*{Step 2.} This step corresponds to Step 2 in the proof of Theorem 3 in \cite{Zhou_et_al_2023}.

\paragraph*{Step 2.1.}  We establish that, for each $j\in [d]$, the process $$
(\tilde{Z}_{t,j}^\top,X_{t+1,j})^\top,\quad t\in\mathbb{Z},$$ is stationary and $\beta$-mixing.

Denote the stationary process from Assumption~\ref{asmpt:mixing} by \begin{equation}
\label{eqn:factor_process} V_t=(X_t^\top,Y_t^\top,Z_t^\top,\xi_t^\top)^\top,\quad t\in\mathbb{Z}.
\end{equation} Each process $(\tilde{Z}_{t,j}^\top,X_{t+1,j})^\top$, $t\in\mathbb{Z}$, is stationary because there exists a Borel map $\Psi_j$ such that 
$$(\tilde{Z}_{t,j}^\top,X_{t+1,j})^\top=\Psi_j(V_t,V_{t+1}).$$

For each $j\in [d]$, denote the $\beta$-mixing coefficient of $(\tilde{Z}_{t,j}^\top,X_{t+1,j})^\top$, $t\in\mathbb{Z}$, by $\beta_j(i)$, and that of $V_t$, $t\in\mathbb{Z}$, by $\beta_{V}(i)$, with a slight abuse of notation. Note that, for every $t'\in\mathbb{Z}$ and $i > 1$,
$$
\sigma((\tilde{Z}_{t,j}^\top,X_{t+1,j})^\top:t\leq t')
\subseteq
\sigma(V_t:t\leq t'+1),
$$
and
$$\sigma((\tilde{Z}_{t,j}^\top,X_{t+1,j})^\top:t\geq t'+i)
\subseteq
\sigma(V_t:t\geq t'+i),$$
hence $$\beta_j(i)
\leq
\beta_{V}(i-1)
\leq
c_{\mathrm{mix}}^1\exp(-c_{\mathrm{mix}}^2(i-1)), \quad i \geq 2,$$ by the definition of the $\beta$-mixing coefficient from Definition~\ref{def:beta_mixing} because the $\sigma$-algebras are nested, where $c_{\mathrm{mix}}^1$ and $c_{\mathrm{mix}}^2$ are from Assumption~\ref{asmpt:mixing}. Therefore, there exist constants $c_{\mathrm{mix}}^{1'}, c_{\mathrm{mix}}^{2'} >0$, such that  
$$\max_{j\in [d]}\beta_{j}(i)
\leq c_{\mathrm{mix}}^{1'} \exp(-c_{\mathrm{mix}}^{2'} i), \quad i \geq 1.$$

\paragraph*{Step 2.2.} We use \cite{berbee_1979} coupling following Step 2 in the proof of Theorem 3 from \cite{Zhou_et_al_2023}. For some block length $b_n\in \mathbb{N}$, partition the process into blocks of the form
$$
U_{i,j}
=
\left(
(\tilde{Z}_{m+t,j}^{\top},X_{m+t+1,j})^{\top}
\right)_{t=ib_n+1}^{(i+1)b_n},
\quad i\geq0.
$$
By the decoupling lemma, i.e., Lemma~\ref{lma:beta_mixing_decoulping_lemma5_zhou} (Lemma~5 in \citet{Zhou_et_al_2023}), there exist copies
$$
U_{i,j}^{0}
=
\left(
((\tilde{Z}_{m+t,j}^{0})^{\top},X_{m+t+1,j}^{0})^{\top}
\right)_{t=ib_n+1}^{(i+1)b_n},
$$
such that each $U_{i,j}^{0}$ has the same distribution as
$U_{i,j}$, the sequences $(U_{2i,j}^{0})_{i\geq0}$ and
$(U_{2i+1,j}^{0})_{i\geq0}$ are each iid, and
\begin{equation}\label{eqn:probability_coupling_error}
\mathbb{P}(U_{i,j}^{0}\neq U_{i,j})
\leq\beta_j(b_n).
\end{equation}

Using a union bound over the (at most)
$\lceil N/b_n\rceil$ blocks that intersect with the first $N$
observations, all of these blocks satisfy $U_{i,j}^0=U_{i,j}$
simultaneously with probability at least
\begin{equation}
\label{eqn:mdn_berbee_coupling_probability}
1-
\left\lceil\frac{N}{b_n}\right\rceil
\beta_j(b_n).
\end{equation}

For the couplings $(X_{t+1,j}^{0},\tilde{Z}_{t,j}^{0})_t$ introduced previously, define 
$$
P_{n,j}^{0}h
=
\frac{1}{N}
\sum_{t=m+1}^{n-m-1}
h(X_{t+1,j}^{0},\tilde{Z}_{t,j}^{0}).
$$

As in
equations (27) and (30) of \citet{Zhou_et_al_2023}, with probability at least that of~\eqref{eqn:mdn_berbee_coupling_probability},  \begin{align}
P_{n,j}
\left[
\log f_{X_{t+1,j}\mid\tilde{Z}_{t,j}}
-\log f_{j,\epsilon}^{\mathrm{app}}
\right]
&=
P_{n,j}^{0}
\left[
\log f_{X_{t+1,j}\mid\tilde{Z}_{t,j}}
-\log f_{j,\epsilon}^{\mathrm{app}}
\right],
\label{eqn:mdn_coupled_approximation_identity}
\\
(P_j-P_{n,j})
\left[
\log f_{X_{t+1,j}\mid\tilde{Z}_{t,j}}
-\log\hat{f}_j^{\mathrm{oracle}}
\right]
&=
(P_j-P_{n,j}^{0})
\left[
\log f_{X_{t+1,j}\mid\tilde{Z}_{t,j}}
-\log\hat{f}_j^{\mathrm{oracle}}
\right].
\label{eqn:mdn_coupled_estimation_identity}
\end{align}

Next, as in equation (28) of \citet{Zhou_et_al_2023}, we
separate the even and odd blocks. For notational simplicity, suppose that
$2b_n$ divides $N$. Then
$$
\begin{aligned}
P_{n,j}^{0}h
&=
\frac{1}{2}
\frac{2b_n}{N}
\sum_{i=0}^{N/(2b_n)-1}
\frac{1}{b_n}
\sum_{\ell=1}^{b_n}
h\left(
X_{m+2ib_n+\ell+1,j}^{0},
\tilde{Z}_{m+2ib_n+\ell,j}^{0}
\right)
\\&
+
\frac{1}{2}
\frac{2b_n}{N}
\sum_{i=0}^{N/(2b_n)-1}
\frac{1}{b_n}
\sum_{\ell=1}^{b_n}
h\left(
X_{m+(2i+1)b_n+\ell+1,j}^{0},
\tilde{Z}_{m+(2i+1)b_n+\ell,j}^{0}
\right).
\end{aligned}
$$
Note that, for a fixed measurable function $h$, each quantity  
$$\frac{1}{b_n}
\sum_{\ell=1}^{b_n}
h\left(
X_{m+2ib_n+\ell+1,j}^{0},
\tilde{Z}_{m+2ib_n+\ell,j}^{0}
\right),\quad \frac{1}{b_n}
\sum_{\ell=1}^{b_n}
h\left(
X_{m+(2i+1)b_n+\ell+1,j}^{0},
\tilde{Z}_{m+(2i+1)b_n+\ell,j}^{0}
\right),$$ is an iid block
average with expectation $P_jh$, so the average consists of two quantities involving iid block averages, each with $N/(2b_n)$
blocks. When $2b_n$ does not divide $N$, the remaining observations are fewer than
$2b_n$, so they contribute $O(b_n/n)=O(\log n/n)$ to the average and thus they do not affect the final rate.

Finally, choose
$$
b_n
=
\left\lceil
\frac{2}{c_{\mathrm{mix}}^{2'}}
\log n
\right\rceil,
$$
so that
$$
\left\lceil\frac{N}{b_n}\right\rceil\beta_j(b_n)
\leq
c_{\mathrm{mix}}^{1'}
\left\lceil\frac{N}{b_n}\right\rceil n^{-2}
\leq
c_{\mathrm{mix}}^{1'}n^{-1}.
$$

Therefore, with probability at least $1-O(n^{-1})$, every block that
intersects with the first $N$ observations is equal to its coupled
copy. Hence, the coupling identities
\eqref{eqn:mdn_coupled_approximation_identity} and
\eqref{eqn:mdn_coupled_estimation_identity} hold simultaneously.
Moreover, since $b_n\asymp\log n$, the even-indexed block subsequence, $(U_{2i,j}^{0})_{i\geq0}$, and the odd-indexed
block subsequence, $(U_{2i+1,j}^{0})_{i\geq0}$, each contain $
N/(2b_n)
\asymp n/\log(n)$ complete blocks in the first $N$ observations.

\paragraph*{Step 2.3.}

Next, as in \cite{Zhou_et_al_2023}, we use the iid sequences of blocks from Step 2.2 to extend the
bounds from Steps 1.2.4 and 1.3.4 to the setting of dependent sequences.

In each of the even-indexed and odd-indexed block subsequences, there are
$N/(2b_n)$ iid blocks. Therefore, the iid concentration and empirical process arguments may be applied to the even-indexed or odd-indexed block subsequence with sample size $N/(2b_n)$. Also, by Jensen's inequality averaging over observations within each block does not increase the bounds from Step 1, so the covering
number bounds remain valid for the corresponding block averages.

For the comparison of the population norm and empirical norm, we apply Step 1.3.3.2 to the iid coupled blocks. By Jensen's inequality, the block averages of $h^2$ have mean $P_jh^2$ and variance bounded by $4c_+^2P_jh^2$, so the covering and Dudley bounds continue to hold for the block averages. Since the number of iid blocks is proportional to $N/b_n$, the corresponding condition is
\begin{equation}\label{eqn:mdn_block_localization_condition}
r_{0,j}M_j^{\omega_2}
\geq c_{\mathrm{loc}}\sqrt{
\frac{b_n}{N}\left(
\max\left(1,M_j\mathrm{Pdim}(\mathcal{M}_{n,j}^{\mathrm{DNN}})\right)
\log(nM_j)+\tilde{r}\right)}.
\end{equation}

The following inequalities are analogous to equations (29) and (31) in \citet{Zhou_et_al_2023}. Applying the argument from Step 1.2.4 to the even-indexed and
odd-indexed iid sequences of blocks and combining the two bounds yields,
for every $\tilde{r}>0$, with probability at least
$1-K e^{-\tilde{r}}$ for some constant $K>0$,
\begin{equation}\label{eqn:dependent_MDN_approximation_bound} \begin{aligned}
&
P_{n,j}^{0}
\left[
\log f_{X_{t+1,j}\mid\tilde{Z}_{t,j}}
-
\log f_{j,\epsilon}^{\mathrm{app}}
\right]
\\
&\leq
C
\left(
M_j^{4\omega_2}\epsilon
+
M_j^{-\omega_1}
\right)^2
\\
&\quad+
C
\left(
M_j^{4\omega_2}\epsilon
+
M_j^{-\omega_1}
\right)
\sqrt{
\frac{
b_n\tilde{r}
}{
N
}
}
+
C
\left(
M_j^{4\omega_2}\epsilon
+
M_j^{-\omega_1}
\right)
\frac{
b_n\tilde{r}
}{
N
}.
\end{aligned}
\end{equation}

Similarly, applying the argument from Step 1.3 to each iid block
subsequence, provided~\eqref{eqn:mdn_block_localization_condition} holds, yields, with probability
at least $1-K' e^{-\tilde{r}}$ for some constant $K' >0$,
\begin{equation}
\label{eqn:dependent_MDN_estimation_bound}
\begin{aligned}
&
\sup_{f_j\in\mathcal{M}_{n,j}^{\mathrm{MDN}}(r_{0,j})}
\left|
(P_j-P_{n,j}^{0})
\left[
\log f_{X_{t+1,j}\mid\tilde{Z}_{t,j}}
-
\log f_j
\right]
\right|
\\
&\leq
C
\Bigg(
r_{0,j}M_j^{\omega_2}
\sqrt{
\frac{
b_n
M_j
\mathrm{Pdim}(\mathcal{M}_{n,j}^{\mathrm{DNN}})
}{
N
}
\log(nM_j)
}+
r_{0,j}M_j^{\omega_2}
\sqrt{
\frac{
b_n\tilde{r}
}{
N
}
}
+
M_j^{\omega_2}
\frac{
b_n\tilde{r}
}{
N
}
\Bigg).
\end{aligned}
\end{equation}

\paragraph*{Step 2.4.}

Next, we obtain the convergence rate for each conditional density estimator by combining the previous bounds. Using the bounds on $K_j(\hat{f}_j^{\mathrm{oracle}})$ from Step 1.1, taking $f_j^{\mathrm{DNN}}=f_{j,\epsilon}^{\mathrm{app}}$, the coupling identities from Step 2.2,~\eqref{eqn:dependent_MDN_approximation_bound} and~\eqref{eqn:dependent_MDN_estimation_bound}, we have that, for every $\tilde{r}>0$ satisfying~\eqref{eqn:mdn_block_localization_condition}, with probability at least $1-K''e^{-\tilde{r}}-O(n^{-1})$ for some constant $K''>0$,
\begin{align*}
\left\|
\hat{f}_j^{\mathrm{oracle}}
-
f_{X_{t+1,j}\mid\tilde{Z}_{t,j}}
\right\|_{L^2,j}^2
&\leq
C\Bigg(
r_{0,j}M_j^{\omega_2}
\sqrt{
\frac{
b_nM_j
\mathrm{Pdim}(\mathcal{M}_{n,j}^{\mathrm{DNN}})
}{
N
}
\log(nM_j)
}
\\
&\quad+
r_{0,j}M_j^{\omega_2}
\sqrt{
\frac{b_n\tilde{r}}{N}
}
+
M_j^{\omega_2}
\frac{b_n\tilde{r}}{N}
\\
&\quad+
\left(
M_j^{4\omega_2}\epsilon
+
M_j^{-\omega_1}
\right)^2
+
\left(
M_j^{4\omega_2}\epsilon
+
M_j^{-\omega_1}
\right)
\sqrt{
\frac{b_n\tilde{r}}{N}
}\\
& \quad +
\left(
M_j^{4\omega_2}\epsilon
+
M_j^{-\omega_1}
\right)
\frac{b_n\tilde{r}}{N}
\Bigg),
\end{align*}
for some constant $C>0$.

As in Step 2 of \citet{Zhou_et_al_2023}, we apply the arguments from Appendices A.2.1--A.2.4 of
\citet{dnn_est_inf} to recursively improve the upper bound $r_{0,j}$. Since $b_n\asymp\log n$, this yields, for every
$\tilde{r}>0$, with probability at least
$1-K'''e^{-\tilde{r}}-O(n^{-1})$ for some constant $K'''>0$,
\begin{align*}
\left\|
\hat{f}_j^{\mathrm{oracle}}
-
f_{X_{t+1,j}\mid\tilde{Z}_{t,j}}
\right\|_{L^2,j}
&\leq
C
\Bigg(
M_j^{4\omega_2}
\Bigg[
\sqrt{
\frac{
M_j\mathrm{Pdim}(\mathcal{M}_{n,j}^{\mathrm{DNN}})
}{
N
}
}
\log(nM_j)
\nonumber\\
&\qquad\quad+
\log n
\sqrt{
\frac{
\log\log n+\tilde{r}
}{
N
}
}
+
\epsilon\log n
\Bigg]
+
M_j^{-\omega_1}
\Bigg).
\end{align*}

By~\eqref{eqn:mdn_coefficient_pseudoVCdimension},
\begin{align*}
\mathrm{Pdim}(\mathcal{M}_{n,j}^{\mathrm{DNN}})
\leq
C
M_j^{(k+j-1)/\gamma}
\epsilon^{-(k+j-1)/\gamma}
\left(
\log(M_j/\epsilon)+1
\right)^3.
\end{align*}
Recall from Step 1.2.2 that
$$
\epsilon
=
n^{-\gamma/(2\gamma+k+j-1)}.
$$
Therefore, we have 
\begin{align*}
\sqrt{
\frac{
M_j\mathrm{Pdim}(\mathcal{M}_{n,j}^{\mathrm{DNN}})
}{
N
}
}
\log(nM_j)\leq
C
M_j^{\frac{\gamma+k+j-1}{2\gamma}}
n^{-\frac{\gamma}{2\gamma+k+j-1}}
\log^3(nM_j).
\end{align*}
Also, setting $\tilde{r}=\log n$, we have
\begin{align*}
\log n
\sqrt{
\frac{
\log\log n+\tilde{r}
}{
N
}
}
\leq
C
n^{-1/2}\log^{3/2}n, \quad \epsilon\log n
=
n^{-\frac{\gamma}{2\gamma+k+j-1}}
\log n.
\end{align*}
Since $M_j \geq 1$ and $\gamma/(2\gamma+k+j-1)<1/2$, both terms are bounded by
$$
C
M_j^{\frac{\gamma+k+j-1}{2\gamma}}
n^{-\frac{\gamma}{2\gamma+k+j-1}}
\log^3(nM_j).
$$

Putting it all together, for each $j\in[d]$, with probability at
least $1-O(n^{-1})$, we have
\begin{equation}
\label{eqn:mdn_rate_with_observed_covariates}
\begin{aligned}
\left\|
\hat{f}_j^{\mathrm{oracle}}
-
f_{X_{t+1,j}\mid\tilde{Z}_{t,j}}
\right\|_{L^2,j}
\leq
C
\left(
M_j^{-\omega_1}
+
M_j^{
\frac{\gamma+k+j-1}{2\gamma}
+
4\omega_2
}
n^{-\frac{\gamma}{2\gamma+k+j-1}}
\log^3(nM_j)
\right),
\end{aligned}
\end{equation}
for some constant $C>0$.

\paragraph*{Step 3.}
Lastly, we combine the conditional density bounds for each $j\in [d]$ from
Step 2.4 to obtain the bound for the conditional joint density. By the factorization of $f_{X_{t+1}\mid Z_t}$ from~\eqref{eqn:factorization_conditional_density_Xt+1_given_Zt}, the definition of $\hat{f}^{\mathrm{oracle}}$ in
\eqref{eqn:oracle_mdn_estimator}, and a telescoping identity for the difference of two products, 
\begin{align*}
&
\hat{f}^{\mathrm{oracle}}(x\mid z)
-
f_{X_{t+1}\mid Z_t}(x\mid z)\\
&=\prod_{j=1}^d \hat{f}_{j}^{\mathrm{oracle}}(x_j\mid \tilde{z}_j)-
\prod_{j=1}^d
    f_{X_{t+1,j}\mid \tilde{Z}_{t,j}}(x_j\mid \tilde{z}_j)
\\
&=
\sum_{j=1}^d
\left(
\prod_{i=1}^{j-1}
f_{X_{t+1,i}\mid\tilde{Z}_{t,i}}
(x_i\mid\tilde{z}_i)
\right)
\left(
\hat{f}_j^{\mathrm{oracle}}
(x_j\mid\tilde{z}_j)
-
f_{X_{t+1,j}\mid\tilde{Z}_{t,j}}
(x_j\mid\tilde{z}_j)
\right)
\\
&
\left(\prod_{i=j+1}^{d}
\hat{f}_i^{\mathrm{oracle}}
(x_i\mid\tilde{z}_i)
\right).
\end{align*}

For every
$j\in[d]$,
\begin{align*}
&
\Bigg\|
\left(
\prod_{i=1}^{j-1}
f_{X_{t+1,i}\mid\tilde{Z}_{t,i}}
\right)
\left(
\hat{f}_j^{\mathrm{oracle}}
-
f_{X_{t+1,j}\mid\tilde{Z}_{t,j}}
\right)
\left(
\prod_{i=j+1}^{d}
\hat{f}_i^{\mathrm{oracle}}
\right)
\Bigg\|_{L^2}^2
\\
&\leq
(c_f^+)^{j-1}
c_{\mathrm{MDN}}^{2(d-j)}
\left(
\prod_{i=j+1}^{d}
\lambda_{\mathcal{X}_i}(\mathcal{X}_i)
\right)
\left\|
\hat{f}_j^{\mathrm{oracle}}
-
f_{X_{t+1,j}\mid\tilde{Z}_{t,j}}
\right\|_{L^2,j}^2
\\
&\leq
C
\left\|
\hat{f}_j^{\mathrm{oracle}}
-
f_{X_{t+1,j}\mid\tilde{Z}_{t,j}}
\right\|_{L^2,j}^2,
\end{align*} 
for some constant $C>0$, where the first inequality holds because the true conditional density factors are upper bounded by $c_f^+$ and the estimated MDN factors are upper bounded by $c_{\mathrm{MDN}}$ by Assumptions~\ref{asmpt:density_regularity_conditions}
and~\ref{asmpt:mdn_model}, and the second inequality follows because
$d$ is fixed and $\mathcal{X}$ is compact by Assumption~\ref{asmpt:compact_X}. Hence, by the triangle inequality, we have
\begin{align}
\left\|
\hat{f}^{\mathrm{oracle}}
-
f_{X_{t+1}\mid Z_t}
\right\|_{L^2}\leq
C
\sum_{j=1}^d
\left\|
\hat{f}_j^{\mathrm{oracle}}
-
f_{X_{t+1,j}\mid\tilde{Z}_{t,j}}
\right\|_{L^2,j}.
\label{eqn:joint_density_coordinatewise_bound}
\end{align}

Since $d$ is fixed, a union bound over~\eqref{eqn:mdn_rate_with_observed_covariates}
implies that, with probability at least $1-O(n^{-1})$, the bounds hold simultaneously for each $j\in[d]$. 

Recall $M_-=\min_{j\in [d]}(M_j)$ and $M_+=\max_{j\in [d]}(M_j)$. Observe that $M_j^{-\omega_1}
\leq
M_-^{-\omega_1},
$ and $1 \leq M_j\leq M_+$, so we have
\begin{align*}
M_j^{
\frac{\gamma+k+j-1}{2\gamma}
+
4\omega_2
}
n^{-\frac{\gamma}{2\gamma+k+j-1}}
\log^3(nM_j)
\leq
M_+^{
\frac{\gamma+k+d-1}{2\gamma}
+
4\omega_2
}
n^{-\frac{\gamma}{2\gamma+k+d-1}}
\log^3(nM_+).
\end{align*}

Combining these inequalities with
\eqref{eqn:joint_density_coordinatewise_bound} yields, with probability
at least $1-O(n^{-1})$,
\begin{align*}
\left\|
\hat{f}^{\mathrm{oracle}}
-
f_{X_{t+1}\mid Z_t}
\right\|_{L^2} \leq
c
\left(
M_-^{-\omega_1}
+
M_+^{
\frac{\gamma+k+d-1}{2\gamma}
+
4\omega_2
}
n^{-\frac{\gamma}{2\gamma+k+d-1}}
\log^3(nM_+)
\right),
\end{align*}
for some constant $c>0$, which proves~\eqref{eqn:mdn_rate_with_observed_covariates_joint}.
\qed

\subsection{Proof of Theorem~\ref{thm:one_step_implies_pred_suff_entire_future_standard_Borel_space}}

The proof is broken into two main steps.

\paragraph*{Step 1.} First, we prove via induction that, for every $L\in\mathbb{N}$, we have $$X_{t+1:t+L}
\indep
X_{-\infty:t}
\mid S_t.$$ 

\paragraph*{Step 1.1.} The base case $L=1$, that \begin{equation}\label{eqn:base_case} X_{t+1} \indep X_{-\infty:t} \mid S_t,\end{equation} holds by the one-step-ahead predictive sufficiency property from~\eqref{eqn:Borel_space_pred_suff_property_CI}. 

\paragraph*{Step 1.2.} Suppose that, for some $L\in\mathbb{N}$, we have \begin{equation}\label{eqn:inductive_hypothesis}X_{t+1:t+L}
\indep
X_{-\infty:t}
\mid S_t.\end{equation} 
Equivalently, by Doob's conditional
independence property (Theorem~8.9 in~\cite{Kallenberg-mod-prob}),
\begin{equation}\label{eqn:inductive_hypothesis_equivalence}
\mathcal{L}(X_{t+1:t+L}\mid\mathcal{F}_t)
=
\mathcal{L}(X_{t+1:t+L}\mid S_t),\end{equation} almost surely.

Observe that, for each $L\in\mathbb{N}$, there exists a measurable function $\tau^{(L)}$ that is defined as the recursive application of $\tau$ to $S_t$ and $X_{t+1:t+L}$, such that
\begin{equation}\label{eqn:tau_L}
S_{t+L}
=
\tau^{(L)}(S_t,X_{t+1:t+L})=
\tau(\cdots
\tau(
\tau(S_t, X_{t+1}),
X_{t+2}
)
\cdots,
X_{t+L}
).
\end{equation} For example, for $L=1$ and $L=2$, we have \begin{align*}
S_{t+1}
&=
\tau^{(1)}(S_t,X_{t+1}) = \tau(S_t,X_{t+1}), \\ S_{t+2}
&=
\tau^{(2)}(S_t,X_{t+1:t+2}) = \tau(\tau(S_t,X_{t+1}),X_{t+2})=\tau(S_{t+1},X_{t+2}).
\end{align*}

For any sets $A\in\mathfrak{X}^{\otimes L}$ and
$B\in\mathfrak{X}$, where $\mathfrak{X}^{\otimes L}$ is the product $\sigma$-algebra, we have, almost surely,  \begin{equation*}
\mathbb{P}
\left(
X_{t+1:t+L}\in A,
X_{t+L+1}\in B
\mid
\mathcal{F}_t
\right)
=
\mathbb{E}
\left[
\mathbf{1}\{X_{t+1:t+L}\in A\}
\mathbb{P}
\left(
X_{t+L+1}\in B
\mid
\mathcal{F}_{t+L}
\right)
\mid
\mathcal{F}_t
\right],
\end{equation*} by the tower property of conditional expectation, and
$$
\mathbb{P}
\left(
X_{t+L+1}\in B
\mid
\mathcal{F}_{t+L}
\right)
=
\mathbb{P}
\left(
X_{t+L+1}\in B
\mid
S_{t+L}
\right),
$$ by the one-step-ahead predictive sufficiency property from~\eqref{eqn:Borel_space_pred_suff_property_CI}. Therefore, we have \begin{align*}
\mathbb{P}
\left(
X_{t+1:t+L}\in A,
X_{t+L+1}\in B
\mid
\mathcal{F}_t
\right)
&
=
\mathbb{E}
\left[
\mathbf{1}\{X_{t+1:t+L}\in A\}
\mathbb{P}
\left(
X_{t+L+1}\in B
\mid
S_{t+L}
\right)
\mid
\mathcal{F}_t
\right],
\end{align*} almost surely, by applying the previous equalities.

By~\eqref{eqn:tau_L}, the quantity $\mathbf{1}\{X_{t+1:t+L}\in A\}
\mathbb{P}
\left(
X_{t+L+1}\in B
\mid
S_{t+L}
\right)$ is a measurable function of $(S_t, X_{t+1:t+L})$, so by~\eqref{eqn:inductive_hypothesis_equivalence}, we have
\begin{align*}
\mathbb{P}
\left(
X_{t+1:t+L}\in A,
X_{t+L+1}\in B
\mid
\mathcal{F}_t
\right)
&
=
\mathbb{E}
\left[
\mathbf{1}\{X_{t+1:t+L}\in A\}
\mathbb{P}
\left(
X_{t+L+1}\in B
\mid
S_{t+L}
\right)
\mid
S_t
\right],
\end{align*} almost surely. On the other hand, by the tower property of conditional expectation and the one-step-ahead predictive sufficiency property from~\eqref{eqn:Borel_space_pred_suff_property_CI}, we have \begin{align*}
\mathbb{P}
\left(
X_{t+1:t+L}\in A,
X_{t+L+1}\in B
\mid
S_t
\right)
&=
\mathbb{E}
\left[
\mathbf{1}\{X_{t+1:t+L}\in A\}
\mathbb{P}
\left(
X_{t+L+1}\in B
\mid
\mathcal{F}_{t+L}
\right)
\mid
S_t
\right]
\\
& =
\mathbb{E}
\left[
\mathbf{1}\{X_{t+1:t+L}\in A\}
\mathbb{P}
\left(
X_{t+L+1}\in B
\mid
S_{t+L}
\right)
\mid
S_t
\right],
\end{align*} almost surely. Therefore, we have, almost surely, that
\begin{equation}\label{eqn:equality_A_B_St_Ft}
\mathbb{P}
\left(
X_{t+1:t+L}\in A,
X_{t+L+1}\in B
\mid
\mathcal{F}_t
\right)
=
\mathbb{P}
\left(
X_{t+1:t+L}\in A,
X_{t+L+1}\in B
\mid
S_t
\right).\end{equation} 

The Cartesian products $A \times B$, with
$A \in \mathfrak{X}^{\otimes L}$, $B \in \mathfrak{X}$, form a $\pi$-system, i.e., it is closed under finite intersections, that generates $\mathfrak{X}^{\otimes (L+1)}$. Moreover, the collection of sets $C \in \mathfrak{X}^{\otimes (L+1)}$ for which the almost sure equality from~\eqref{eqn:equality_A_B_St_Ft} holds is a $\lambda$-system, i.e., it is closed under proper differences and increasing limits, and it contains $\mathcal{X}^{L+1}$. Therefore, by the monotone class theorem
(Theorem 1.1 in \cite{Kallenberg-mod-prob}), the almost sure equality from~\eqref{eqn:equality_A_B_St_Ft} holds for every set $C \in \mathfrak{X}^{\otimes (L+1)}$. Since $\sigma(S_t)\subseteq \mathcal{F}_t$, Doob's conditional
independence property (Theorem~8.9 in~\cite{Kallenberg-mod-prob}) implies that 
$$
X_{t+1:t+L+1}
\!\perp\!\!\!\perp
X_{-\infty:t}
\mid S_t.
$$ Hence, if the inductive hypothesis from~\eqref{eqn:inductive_hypothesis} holds for $L$, then it holds for $L+1$.

\paragraph*{Step 1.3.} By~\eqref{eqn:Borel_space_pred_suff_property_CI}, the inductive hypothesis holds for $L=1$, so for every $L\in\mathbb{N}$ we have,  \begin{equation}\label{eqn:finite_L_cond_indep}
X_{t+1:t+L}
\!\perp\!\!\!\perp
X_{-\infty:t}
\mid S_t.\end{equation} Since $S_t=R(X_{-\infty:t})$ by~\eqref{eqn:data_generating_process_standard_Borel_conditional_distributions}, we have
$\sigma(S_t)\subseteq\mathcal{F}_t$, so by Doob's conditional
independence property (Theorem~8.9 in~\cite{Kallenberg-mod-prob}),~\eqref{eqn:finite_L_cond_indep} is equivalent to
\begin{equation}\label{eqn:finite_L_equality}\mathcal{L}(X_{t+1:t+L}\mid\mathcal{F}_t)
=
\mathcal{L}(X_{t+1:t+L}\mid S_t),
\end{equation} almost surely.

\paragraph*{Step 2.} Second, we extend this statement to the infinite future, $$
X_{t+1:\infty}
\!\perp\!\!\!\perp
X_{-\infty:t}
\mid S_t,
$$ via the monotone class theorem (Theorem 1.1 in \cite{Kallenberg-mod-prob}). Define \begin{equation}\label{defn:D_lambda_class}\mathcal{D}=\{A\in\sigma(X_{t+1:\infty}):
\mathbb{P}(A\mid\mathcal{F}_t)
=
\mathbb{P}(A\mid S_t) \text{ almost surely}\}.\end{equation} We claim that $\mathcal{D}$ is a $\lambda$-system, i.e., it is closed under proper differences and increasing limits, and it contains $\Omega$. First, if $A,B\in\mathcal{D}$ with
$B\subseteq A$, then by the linearity property of conditional expectation,
\begin{equation*}
\mathbb{P}(A\setminus B\mid\mathcal{F}_t) =
\mathbb{P}(A\mid\mathcal{F}_t)
-
\mathbb{P}(B\mid\mathcal{F}_t)
=
\mathbb{P}(A\mid S_t)
-
\mathbb{P}(B\mid S_t)
=
\mathbb{P}(A\setminus B\mid S_t),
\end{equation*}
almost surely, so $A\setminus B\in\mathcal{D}$. Second, if
$A_n\in\mathcal{D}$ with $A_n\uparrow A$, then, almost surely,
$$
\mathbb{P}(A\mid\mathcal{F}_t)
=
\underset{n\xrightarrow[]{}\infty}{\lim}\mathbb{P}(A_n\mid\mathcal{F}_t)
= \underset{n\xrightarrow[]{}\infty}{\lim} \mathbb{P}(A_n\mid S_t)
=
\mathbb{P}(A\mid S_t),
$$
by the monotone
convergence property of conditional expectation. Third, observe that $$\mathbb{P}(\Omega\mid\mathcal{F}_t)
=
\mathbb{P}(\Omega\mid S_t)=1,$$ almost surely, hence $\Omega\in\mathcal{D}$. Therefore, $\mathcal{D}$ is a $\lambda$-system.

Next, we claim that the class
$$\mathcal{C}=
\bigcup_{L\in\mathbb{N}}\sigma(X_{t+1:t+L}),
$$
is a $\pi$-system, i.e., it is closed under finite intersections. If $A,B \in \mathcal{C}$ then there must exist $L_1,L_2\in\mathbb{N}$ such that $A \in \sigma(X_{t+1:t+L_1})$ and $B \in \sigma(X_{t+1:t+L_2})$, therefore $A\cap B \in \sigma(X_{t+1:t+\max(L_1,L_2)})$ because the sequence of $\sigma$-algebras $\sigma(X_{t+1:t+L})$, $L\in\mathbb{N}$, is increasing.

 By~\eqref{eqn:finite_L_equality}, we have that
$$
\mathcal{C}
\subseteq
\mathcal{D}.
$$ Thus, by the monotone class theorem
(Theorem 1.1 in~\cite{Kallenberg-mod-prob}), we have that $$
\sigma(
\mathcal{C})
\subseteq \mathcal{D}.
$$ 
Moreover, observe that
$$
\sigma(
\mathcal{C})
=
\sigma(X_{t+1:\infty}),
$$ so $\sigma(X_{t+1:\infty})\subseteq \mathcal{D}$. By the definition of $\mathcal{D}$ from~\eqref{defn:D_lambda_class}, we have $\mathcal{D}\subseteq \sigma(X_{t+1:\infty})$. Therefore, $$\mathcal{D} = \sigma(X_{t+1:\infty}).$$
Hence, for every $A\in\sigma(X_{t+1:\infty})$, we have, almost surely, that
\begin{equation}\label{eqn:equality_prob_all_future_events}
\mathbb{P}(A\mid\mathcal{F}_t)
=
\mathbb{P}(A\mid S_t).\end{equation} Since $S_t=R(X_{-\infty:t})$ by~\eqref{eqn:data_generating_process_standard_Borel_conditional_distributions}, we have
$\sigma(S_t)\subseteq\mathcal{F}_t$. By Doob's conditional
independence property (Theorem~8.9 in~\cite{Kallenberg-mod-prob}),
the equality from~\eqref{eqn:equality_prob_all_future_events} is equivalent to
$$
X_{t+1:\infty}
\!\perp\!\!\!\perp
X_{-\infty:t}
\mid S_t,
$$
which is the desired result.
\qed

\subsection{Proof of Lemma~\ref{lma:conditional_expectation_as_function_of_noise_process}}
We may always decompose $Y_{t+1}$ from~\eqref{eqn:target_Yt+1_varphi_Xt+1} as \begin{equation}\label{eqn:infinite_past_conditional_expectation_decomposition}
    Y_{t+1} = Z_t +\xi_{t+1},
\end{equation} where $Z_t=\mathbb{E}[Y_{t+1}\mid \mathcal{F}_t]$ is from~\eqref{eqn:Phi_S_t_conditional_expectation_random_features}, and the noise process $\xi_{t+1}$ satisfies $$\mathbb{E}[\xi_{t+1}\mid\mathcal{F}_t]=\mathbb{E}[Y_{t+1}-Z_t\mid\mathcal{F}_t]=\mathbb{E}[Y_{t+1}\mid \mathcal{F}_t]-\mathbb{E}[Y_{t+1}\mid \mathcal{F}_t]=0,$$ almost surely, where the first equality is by the definition of $\xi_{t+1}=Y_{t+1} - Z_t$ by rearranging~\eqref{eqn:infinite_past_conditional_expectation_decomposition}, and the second equality is by the definition of $Z_t=\mathbb{E}[Y_{t+1}\mid \mathcal{F}_t]$, the linearity property of conditional expectation, and because $\mathbb{E}[Y_{t+1}\mid \mathcal{F}_t]$ is $\mathcal{F}_t$-measurable. Next, observe that \begin{equation}
    \begin{aligned}
        Y_{t+1}  
        &\overset{(1)}{=} Z_t+\xi_{t+1} \\ 
        &\overset{(2)}{=} \Phi(S_t)+\xi_{t+1}\\
        &\overset{(3)}{=}  \Phi(\tau^{(m+1)}(S_{t-m-1},X_{t-m:t}))+\xi_{t+1}\\ 
        &\overset{(4)}{=}  \Phi(\tau^{(m+1)}(\Phi^{-1}(Z_{t-m-1}),X_{t-m:t}))+\xi_{t+1}\\ 
        &\overset{(5)}{=}  \Phi(\tau^{(m+1)}(\Phi^{-1}(Y_{t-m}-\xi_{t-m}),X_{t-m:t}))+\xi_{t+1} \\ 
        &\overset{(6)}{=} g^{\ast}(Y_{t-m},\xi_{t-m}, X_{t-m:t}) +\xi_{t+1},
    \end{aligned}
\end{equation} where (1) is from~\eqref{eqn:infinite_past_conditional_expectation_decomposition}, (2) is by~\eqref{eqn:Phi_S_t_conditional_expectation_random_features}, (3) is by defining the function $\tau^{(m+1)}$ as the recursive application of $\tau$ from~\eqref{eqn:data_generating_process_conditional_distributions} a total of $m+1$ times, (4) is because the inverse $\Phi^{-1}$ exists since $\Phi:\mathcal{S}\xrightarrow[]{}\Phi(\mathcal{S})$ is bijective by the injectivity of $\Phi$ from Assumption~\ref{asmpt:injective_Phi}, (5) is by~\eqref{eqn:infinite_past_conditional_expectation_decomposition} with $t$ replaced by $t-m-1$, and (6) is by the definition of $g^{\ast}(\cdot)$. Specifically, for domain $$\mathcal{D}=\{(y,e,x)\in [-1,1]^k \times [-2,2]^k \times \mathcal{X}^{m+1}:y-e\in \Phi(\mathcal{S}),\tau^{(m+1)}(\Phi^{-1}(y-e),x)\in\mathcal{S}\},$$ where we define the function $g^{\ast}:\mathcal{D}\xrightarrow[]{}\Phi(\mathcal{S})$ as \begin{equation*}g^{\ast}(y,e,x)=\Phi(\tau^{(m+1)}(\Phi^{-1}(y-e),x)). 
\end{equation*} 
\qed

\end{document}